\documentclass[accepted]{uai2024} % after acceptance, for a revised version; 
\usepackage{natbib} % has a nice set of citation styles and commands
    \renewcommand{\bibsection}{\subsubsection*{References}}
\usepackage{mathtools} % amsmath with fixes and additions
\usepackage{booktabs} % commands to create good-looking tables
\usepackage{tikz} % nice language for creating drawings and diagrams

\newcommand{\swap}[3][-]{#3#1#2} % just an example

\usepackage{algorithm}
\usepackage{algorithmic}
\usepackage{amsmath,bm}
\usepackage{comment}
\usepackage {diagbox} 
\usepackage{comment}
\usepackage{mathrsfs}  
\usepackage{amsmath,amsthm}
\newtheorem{theorem}{Theorem} [section]
\newtheorem{lemma}{Lemma}[section]

\newtheorem{assumption}{Assumption}[section]
\newtheorem{definition}{Definition}

\usepackage{amssymb}
\usepackage{natbib}
\usepackage{amsfonts}

\newcommand{\1}{$\mathrm{I}$}
\newcommand{\2}{$\mathrm{I}\hspace{-1.2pt}\mathrm{I}$}
\newcommand{\3}{$\mathrm{i}$}
\newcommand{\4}{$\mathrm{i}\hspace{-0.8pt}\mathrm{i}$}
\newcommand{\5}{$\mathrm{i}\hspace{-0.8pt}\mathrm{i}\hspace{-0.8pt}\mathrm{i}$}
\newcommand{\6}{$\mathrm{i}\hspace{-0.8pt}\mathrm{v}$}
\newcommand{\indep}{\perp \!\!\! \perp}
\usepackage{amsmath}             
\usepackage{lscape}
\usepackage{algorithm}

\usepackage{color}
\usepackage{tikz}
\usepackage{amsmath,amsthm}

\usepackage{comment}
\usepackage{here}
\allowdisplaybreaks[4]
\usepackage{bbm}
\usepackage{caption}
\usepackage{bbding}
\usepackage{arydshln}
\usepackage{afterpage}

\usepackage{mathtools}

\usepackage{wrapfig,lipsum,booktabs}

\usepackage{mathrsfs}
\DeclareMathOperator*{\plim}{p-lim}

\newcommand{\jin}[1]{\textcolor{blue}{#1}}
\newcommand{\yuta}[1]{\textcolor{red}{#1}}
\usepackage{newfloat}
\usepackage{listings}
\floatstyle{ruled}
\newfloat{listing}{tb}{lst}{}
\floatname{listing}{Listing}
\usepackage{xcolor}
\usepackage{color}
\usepackage{tikz}
\tikzset{%
mynode/.style={circle,minimum width=.5ex, fill=none,draw}, % no filling
myfillnode/.style={circle,minimum width=.5ex, fill=lightgray,draw}, % fill with black
}
\usepackage{amsmath}
\usepackage{amssymb}

\usepackage{wrapfig}
\usepackage[utf8]{inputenc} % allow utf-8 input
\usepackage[T1]{fontenc}    % use 8-bit T1 fonts
\usepackage{hyperref}       % hyperlinks
\usepackage{url}            % simple URL typesetting
\usepackage{booktabs}       % professional-quality tables
\usepackage{amsfonts}       % blackboard math symbols
\usepackage{nicefrac}       % compact symbols for 1/2, etc.
\usepackage{microtype}      % microtypography
\usepackage{xcolor}         % colors

\usepackage[american]{babel}
\usepackage{booktabs} % commands to create good-looking tables
\usepackage{tikz} % nice language for creating drawings and diagrams

\title{Identification and Estimation of Conditional Average Partial Causal Effects\\ via Instrumental Variable}

\author[1,2]{Yuta Kawakami}
\author[1]{Manabu Kuroki}
\author[2]{Jin Tian}

\affil[1]{%
Department of Mathematics, Physics, Electrical Engineering and Computer Science, Yokohama National University, Yokohama, Kanagawa, JAPAN
}
\affil[2]{%
Department of Computer Science, Iowa State University, Ames, Iowa, USA
}

\begin{document}
\maketitle

\begin{abstract}
%Instrumental variable (IV) analysis is a powerful tool used to elucidate causal relationships.
There has been considerable recent interest in estimating heterogeneous causal effects. In this paper, we study conditional average partial causal effects (CAPCE) to reveal the heterogeneity of causal effects with continuous treatment. We provide conditions for identifying CAPCE in an instrumental variable setting. Notably, CAPCE is identifiable under a weaker assumption than required by a commonly used measure for estimating heterogeneous causal effects of continuous treatment. We develop three families of CAPCE estimators: sieve, parametric, and reproducing kernel Hilbert space (RKHS)-based, and analyze their statistical properties. We illustrate the proposed CAPCE estimators on synthetic and real-world data.
\end{abstract}

%Keyword: Heterogeneous causal effects, Instrumental variable, Continuous treatment
%TL;DR "Too Long; Didn't Read": a short sentence describing your paper
% We provide identification conditions and estimation methods for conditional average partial causal effects in an instrumental variable setting.

\section{Introduction}

Instrumental variable (IV) analysis is a powerful tool used to elucidate causal relationships between treatment ($X$) and outcome ($Y$)  when a controlled experiment is not feasible % or when a randomized experiment is not able to successfully treat each unit
\citep{Imbens2014,Angrist2001}. 
Traditionally, there are a large number of works focusing on binary or categorical treatment variables \citep{Imbens1994,Balke1997}; recently, there has been a growing interest in continuous treatment variables \citep{Imbens2004,Kennedy2017,Bahadori2022}.
%, which is the focus of this paper.
%To identify the structural function $\mathbb{E}[Y_x]$ of a continuous treatment using IV, \cite{Whitney2003} introduced an integral equation, which is widely used in the field of machine learning \citep{Hartford2017,Singh2019,Muandet2020}.
%\yuta{Recently, \citep{Wong2022} introduced another integral equation for identifying average partial causal effect (APCE) $\mathbb{E}[\partial_xY_x]$, and \citep{Kawakami2023} developed two estimation methods of APCE, named P-APCE and N-APCE estimator.}
There is also considerable recent interest in estimating  heterogeneous causal effects %of a binary treatment 
across subsets of the population \citep{Athey2016,Ding2016,Athey2019,Kunzel2019,Wager2018,Zhang2022,Singh2023},  %or continuous treatment \citep{Zhang2022} %\jin{Explain what "heterogeneous" causal effect means.} \yuta{Heterogeneity of causal effect is the nonrandom and explainable variability of causal effects   
%\citep{Varadhan2013}. 
including IV-based methods %is also helpful for estimating CACE 
\citep{Angrist2004,Syrgkanis2019,Klein2020,Bargagli2022}. 
Most of the works focus on \emph{conditional average causal effect (CACE)} $\mathbb{E}[Y_1-Y_0|{\boldsymbol w}]$, also known as conditional average treatment effect (CATE), for evaluating heterogeneous causal effects of a binary $X$, where $Y_x$ denotes the potential outcome under treatment $X=x$, and ${\boldsymbol W}$ are covariates (e.g. gender, age, and race).

In this work, we study estimating  heterogeneous causal effects of a \emph{continuous} treatment via the IV method. Existing work in this direction has focused on  estimating $\mathbb{E}[Y_{x}|{\boldsymbol w}]$.  %and relies on a separability assumption for identifying $\mathbb{E}[Y_{x}|{\boldsymbol w}]$ \citep{Whitney2003}.  
The most widely used methods include parametric two-stage least squared (PTSLS) \citep{Wright1928,Angrist2009,Wooldridge2010}, sieve nonparametric two-stage least squared (sieve NTSLS)  \citep{Whitney2003,Chen2018}, and Kernel IV \citep{Singh2019}. The line of works in \citep{Syrgkanis2019,Dikkala2020,Muandet2020,Bennett2023} focus on the efficiency of estimators assuming simple additive errors. % functions. 
All these methods rely on a \emph{separability} assumption for identifying $\mathbb{E}[Y_{x}|{\boldsymbol w}]$ \citep{Whitney2003}. 

Another quantity for evaluating the causal effects of a continuous treatment is average partial causal effect (APCE) $\mathbb{E}[\partial_xY_x]$ \citep{Chamberlain1984,Wooldridge2005,Graham2012}. \citet{Wong2022} provided a condition for identifying $\mathbb{E}[\partial_xY_x]$ and \citet{Kawakami2023} presented APCE estimators. 

In this paper, 
we consider $\mathbb{E}[\partial_x Y_{x}|{\boldsymbol w}]$, termed \emph{conditional average partial causal effect (CAPCE)},   to capture the heterogeneous causal effects of a continuous treatment. CAPCE extends APCE and is a natural generalization of the CACE of a binary treatment. 
{The quantity represented by CAPCE has been implicitly studied in the literature (e.g. \citep{Galagate2016}). % but is never formally defined to the best of our knowledge. %and it is not tied to the IV analysis. 
Still existing works have focused on $\mathbb{E}[Y_x|{\boldsymbol w}]$. 
One contribution of this work is to show that under the IV model, CAPCE is identifiable under a \emph{weaker} separability assumption than required by  the previous work (sieve NTSLS, PTSLS, Kernel IV) for identifying $\mathbb{E}[Y_x|{\boldsymbol w}]$. Thus, computing CAPCE allows scientists to estimate causal effects in a larger class of models.
%We present theoretical and empirical results to show the usefulness of formally defining and investigating CAPCE and the merits of estimating CAPCE on behalf of $\mathbb{E}[Y_x|{\boldsymbol w}]$. 
Granted, given an estimated $\mathbb{E}[Y_x|{\boldsymbol w}]$, one can compute its derivative to obtain CAPCE, but not the other way around. However, in practice,  the causal effect from a reference point (e.g., CACE) is often the main interest, and CAPCE is enough to compute causal effects from a reference point: $\displaystyle\mathbb{E}[Y_{x''}-Y_{x'}|{\boldsymbol w}]=\int_{x'}^{x''} \mathbb{E}[\partial_x Y_x|{\boldsymbol w}]dx$.
} 

We then develop three families of methods for estimating CAPCE: sieve, parametric, and reproducing 
kernel Hilbert space (RKHS)-based, and analyze their statistical properties. %, extending existing estimators for $\mathbb{E}[Y_x|{\boldsymbol w}]$ \citep{Whitney2003,Singh2019} and APCE \citep{Kawakami2023}. 
%\yuta{Our parametric estimator is the generalization of P-APCE estimator, and the other estimators are not in \citep{Kawakami2023}. Unfortunately, we can not apply N-APCE estimator for estimating CAPCE because the integral kernel of the integral equation for identifying CACPE is the conditional density function.}
Finally, we illustrate the proposed  estimators on synthetic data, showing superior performance to existing methods. %sieve NTSLS and PTSLS. 
We also evaluate CAPCE in   a real-world dataset. % to elicit causal relation between years of education and wages, which is of  interest in economics. % \citep{Card1999,Angrist1991}.

\section{Notation and Background}
%\begin{wrapfigure}{r}[1pt]{0.47\textwidth}
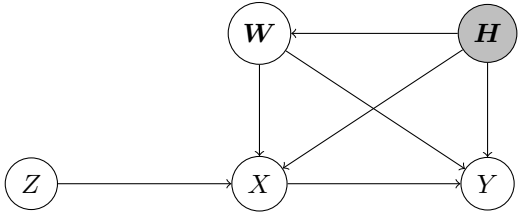
\begin{figure}
    \centering
    %\vspace{-1cm}
    \scalebox{1}{
\begin{tikzpicture}
    % x node set with absolute coordinates
    \node[mynode] (x) at (0,0) {$X$};
    \node[mynode] (y) at (3,0) {$Y$};
    \node[mynode] (z) at (-3,0) {$Z$};
    \node[myfillnode] (u) at (3,2) {${\boldsymbol H}$};
    \node[mynode] (w) at (0,2) {${\boldsymbol W}$};
    %\node[mynode] (d) at (-1.5,2) {${\boldsymbol u}_X$};
    %\node[mynode] (e) at (4.5,2) {${\boldsymbol u}_Y$};

    % Directed edge
    \path (x) edge[->] (y);
    \path (z) edge[->]  (x);

    \path (u) edge[->] (y);
    \path (u) edge[->]  (x);
    \path (u) edge[->]  (w);

    \path (w) edge[->] (y);
    \path (w) edge[->]  (x);
    
    %\path (e) edge[->] (y);
    %\path (d) edge[->]  (x);

\end{tikzpicture}
}
\vspace{0.25cm}
    \caption{A causal graph representing the IV model.}% setting with covariates.}% Causal graph and two types of non-separability in the IV setting, ${\cal M}_{IV}$.}
    \label{DAG1}    
\vspace{-0cm}
\end{figure}
%\end{wrapfigure} 

%In this section, we introduce the basic notations and definitions in this paper and explain the research background.
We represent each variable with a capital letter $(X)$ and its realized value with a small letter $(x)$.
%Let $\mathbbm{1}_{\Omega}(x)$ be an indicator function, which is $1$ if $x \in \Omega$; and $0$ if $x \notin \Omega$.
Let $\Omega_X$ be the domain of $X$, $\mathbb{E}[Y]$ be the expectation of $Y$, $\mathbb{P}(X\leq x)$ be the cumulative distribution function (CDF) of $X$, and $\mathfrak{p}(X=x)$ be the probability density function (PDF) of $X$.
%In addition, $\mathbb{E}[Y|X=x]$, $\mathbb{P}(X\leq x|Z=z)$, and $\mathfrak{p}(X = x|Z=z)$ denote the conditional expectation of $Y$ given $X=x$, the conditional CDF and PDF of $X$ given $Z=z$.
%, and $\mathbb{V}(Y|X=x)$ be the conditional variance of $Y$ given $X=x$.
%We write $g(x)={\cal O}(h(x))$ as $x \rightarrow \infty$ if there exists a positive real number $M$ and a real number $\delta$ such that $|g(x)| \leq Mh(x)$ for all $x \geq \delta$.
%In contrast, we write $g(x)={\cal O}(h(x))$ as $x \rightarrow 0$ if there exists a positive real number $M$ and a real number $\delta$ such that $|g(x)| \leq Mh(x)$ for all $0 \leq |x| \leq \delta$.
{
A metric space $\left<\Omega,d\right>$, where distance function $d(x,y)$ is defined by a given norm $\|x-y\|$ for $x,y \in \Omega$,  is compact if every sequence in $\Omega$ has a convergent sub-sequence whose limit is in $\Omega$. }
If every Cauchy sequence of points in $\Omega$ has a limit in $\Omega$, $\Omega$ is called complete.

%{\bf Sobolev Space.} We define 
{\bf Sobolev norm} \citep{Gallant1987,Leoni2009}. 
Let ${\boldsymbol \lambda}$ be a $d+1$ dimensional vector of non-negative integer, $\displaystyle|{\boldsymbol \lambda}|=\sum_{l=1}^{d+1}{\lambda}_l$, and $\displaystyle D^{{\boldsymbol \lambda}}f(x,{\boldsymbol w})=\partial^{|{\boldsymbol \lambda}|}f(x,{\boldsymbol w})/\partial x^{{\lambda}_1}\partial {w_1}^{{\lambda}_2}\cdots\partial {w_d}^{{\lambda}_{d+1}}$.
Sobolev norm is defined as follows:
%\begin{eqnarray}
%\label{RES1}
    $\displaystyle\|f\|_{W^{l,p}}=\left\{\sum_{|{\boldsymbol \lambda}|\leq l} \int \{D^{{\boldsymbol \lambda}}f(x,{\boldsymbol w})\}^p dxd{\boldsymbol w} \right\}^{1/p}$
%\end{eqnarray}
for $1\leq p < \infty$, and
%\begin{eqnarray}
%\label{RES1}
    $\|f\|_{W^{l,\infty}}=\max_{|{\boldsymbol \lambda}|\leq l}\sup_{(x,{\boldsymbol w})} D^{{\boldsymbol \lambda}}f(x,{\boldsymbol w})$.
%\end{eqnarray}
Note that $W^{0,p}$ norm coincides with $L_p$ norm for $1\leq p \leq \infty$.

{\bf Structural Causal Models (SCM).} %We use the language of Structural Causal Models (SCM) as our basic semantic and inferential framework \citep{Pearl09}.
We use SCM as our framework \citep{Pearl09}.
An SCM ${\cal M}$ is a tuple $\left<{\boldsymbol V},{\boldsymbol U}, {\cal F}, \mathbb{P}_{\boldsymbol U} \right>$, where ${\boldsymbol U}$ is a set of exogenous (unobserved) variables following a joint distribution $\mathbb{P}_{\boldsymbol U}$, and ${\boldsymbol V}$ is a set of endogenous (observable) variables whose values are determined by structural functions ${\cal F}=\{f_{V_i}\}_{V_i \in {\boldsymbol V}}$ such that $v_i:= f_{V_i}({\mathbf{pa}}_{V_i},{\boldsymbol u}_{V_i})$ where ${\mathbf{PA}}_{V_i} \subseteq {\boldsymbol V}$ and $U_{V_i} \subseteq {\boldsymbol U}$. 
%The values of ${\boldsymbol H}$ are drawn from the distribution $\mathbb{P}_{\boldsymbol U,\epsilon_Y}$.
Each SCM ${\cal M}$ induces an observational distribution $\mathbb{P}_{\boldsymbol V}$ over ${\boldsymbol V}$, and a causal graph $G({\cal M})$ over ${\boldsymbol V}$ in which there exists a directed edge from every variable in ${\mathbf{PA}}_{V_i}$ to $V_i$.
%An intervention of setting a set of endogenous variables ${\boldsymbol X}$ to constants ${\boldsymbol x}$, denoted by $do({\boldsymbol x})$, 
An intervention $do({\boldsymbol x})$ of setting  endogenous variables ${\boldsymbol X}$ to constants ${\boldsymbol x}$ 
replaces the functions of ${\boldsymbol X}$
 by the constants ${\boldsymbol x}$ and induces a \textit{sub-model}  ${\cal M}_{{\boldsymbol x}}$.
We denote the potential outcome $Y$ under intervention $do({\boldsymbol x})$ by $Y_{{\boldsymbol x}}({\boldsymbol u})$, which is the solution of $Y$ in the sub-model ${\cal M}_{{\boldsymbol x}}$ given ${\boldsymbol U}={\boldsymbol u}$.

%{\bf Average Partial Causal Effect (APCE).} Considering a continuous treatment $X$, the APCE $\mathbb{E}[\partial_x Y_x]:=\mathbb{E}_{\boldsymbol U}[\frac{\partial}{\partial x}Y_x({\boldsymbol U})]$ have been introduced by \citep{Chamberlain1984,Wooldridge2005,Graham2012}. APCE is a function from $x \in \Omega_X$ to $\mathbb{R}$, and a natural generalization of the average causal effect of a binary treatment $\mathbb{E}[Y_{1}]-\mathbb{E}[Y_{0}]$. % for $x,x' \in \Omega_X$. $\mathbb{E}[\partial_x Y_x]$ enables us to evaluate a popular target, the average causal effect of changing treatments from $x'$ to $x''$, by $\mathbb{E}[Y_{x'}]-\mathbb{E}[Y_{x''}]=\int_{x''}^{x'} \mathbb{E}[\partial_x Y_x] dx$. It is sufficient to evaluate APCE to reveal causal relationships.
\begin{comment}
\textbf{Conditional average causal effect (CACE).} 
%Heterogeneity of causal effect is the nonrandom and explainable variability in causal effects across subsets of the population \citep{Varadhan2013}.
CACE $\mathbb{E}[Y_1-Y_0|{\boldsymbol w}]:=\mathbb{E}_{\boldsymbol U}[Y_1({\boldsymbol U})-Y_0({\boldsymbol U})|{\boldsymbol W}={\boldsymbol w}]$  is the most common quantity for evaluating heterogeneous causal effects of  a binary treatment across subsets of the population where ${\boldsymbol W}$ is a set of covariates \citep{Athey2016,Ding2016,Kunzel2019,Wager2018}.
%We define the CACE below:
%\begin{definition}[Conditional average causal effect (CACE)]
%$\mathbb{E}[Y_1-Y_0|{\boldsymbol W}={\boldsymbol w}]$
%\end{definition}
%Here, ${\boldsymbol W}$ is the subjects covariates.
\end{comment}

{\bf Instrumental Variable (IV) Model with Covariates.}
We consider the IV model represented by the causal graph in Fig \ref{DAG1}, with the following SCM ${\cal M}_{IV}$ over ${\boldsymbol V}=\{Z,X,Y,{\boldsymbol W}\}$ and ${\boldsymbol U}=\{{\boldsymbol H},{\boldsymbol u}_X,{\boldsymbol u}_Y,{\boldsymbol u}_Z,{\boldsymbol u}_{\boldsymbol W}\}$: %\jin{Do the results hold with Z = f(W, ) or an edge from W to Z in Figure 1?} \yuta{[If there is an edge from W to Z in Figure 1, the distribution $P(X,W|Z)$, say $P(X|Z)$, is biased.]}
%The SCM of the IV model, ${\cal M}_{IV}$, is defined as
%\begin{eqnarray}
%\vspace{-0.9cm}
%{%\small
\begin{equation}
\begin{gathered}
Y:=f_Y(X,{\boldsymbol W},{\boldsymbol H},{\boldsymbol u}_Y),\  X:=f_X(Z,{\boldsymbol W},{\boldsymbol H},{\boldsymbol u}_X),\\
{\boldsymbol W}:=f_{\boldsymbol W}({\boldsymbol H},{\boldsymbol u}_{\boldsymbol W}),\  
Z:=f_Z({\boldsymbol u}_Z),
%\left\{
%\begin{array}{l}
%    Y:=f_Y(X,{\boldsymbol W},{\boldsymbol H},{\boldsymbol u}_Y)\\
%    X:=f_X(Z,{\boldsymbol W},{\boldsymbol H},{\boldsymbol u}_X)\\
%    {\boldsymbol W}:=f_X({\boldsymbol H},{\boldsymbol u}_{\boldsymbol W})
%\end{array}
%\right.
%\end{eqnarray}
\end{gathered}
\end{equation}
%\normalsize}
%with the conditional joint distribution $\mathbb{P}_{\{X,Y\}|Z}$.
where %$f_Y$, $f_X$, and $f_Z$ are scalar functions, and 
$f_{\boldsymbol W}$  is a vector function.  % ${\boldsymbol U}=\{{\boldsymbol H},{\boldsymbol u}_X,{\boldsymbol u}_Y,{\boldsymbol u}_Z,{\boldsymbol u}_{\boldsymbol W}\}$, and .
We assume all variables are continuous, %$Z$ are $M$-dimensional instrumental variables,  
${\boldsymbol W}$ are $d$-dimensional pre-treatment covariates, and ${\boldsymbol H}$ stands for unmeasured confounders. %and $d$-dimensional pre-treatment covariates ${\boldsymbol W}$ are generated from ${\boldsymbol H}$. 
{This IV model has been studied in e.g., \citep{Hartford2017,Klein2020}. We further consider an IV model with an additional edge ${\boldsymbol W} \rightarrow Z$ in Appendix \ref{appA2}. }
{\bf Related work.} 
%With covariates ${\boldsymbol W}$ and IVs $Z$, 
Under the IV model, 
\citet{Whitney2003} introduced %sieve nonparametric two-stage least square (
sieve NTSLS for identifying and estimating $\mathbb{E}[Y_{x}|{\boldsymbol w}]$ via an integral equation, $\displaystyle \mathbb{E}[Y|Z=z]=\int_{\Omega_{\boldsymbol W}}\int_{\Omega_X} \mathfrak{p}(X=x,{\boldsymbol W}={\boldsymbol w}|Z=z)\mathbb{E}[Y_{x}|{\boldsymbol w}]dxd{\boldsymbol w}$, under the following assumption called \emph{separability}: 
%\vspace{-0.9cm}
\begin{equation}
\begin{gathered}
%\begin{eqnarray}
\label{eq-sep}
%\text{Separability:}
f_Y(X,{\boldsymbol W},{\boldsymbol H},{\boldsymbol u}_Y)=f_Y^1(X,{\boldsymbol W},{\boldsymbol u}_Y)+f_Y^2({\boldsymbol H}, {\boldsymbol u}_Y),\\
\mathbb{E}[f_Y^2({\boldsymbol H},{\boldsymbol u}_Y)|{\boldsymbol W}]=0, 
%\end{eqnarray}
\end{gathered}
\end{equation}
which says the function $f_Y(X,{\boldsymbol W},{\boldsymbol H},{\boldsymbol u}_Y)$ is in the form of a summation of two functions, one over $(X,{\boldsymbol W})$ and one over ${\boldsymbol H}$.   {Parametric PTSLS \citep{Angrist2009,Wooldridge2010} and Kernel IV \citep{Singh2019} methods for estimating $\mathbb{E}[Y_{x}|{\boldsymbol w}]$ have also been developed under the separability assumption.} 

%separability $f_Y(X,{\boldsymbol W},{\boldsymbol H},{\boldsymbol u}_Y)=f_Y^1(X,{\boldsymbol W},{\boldsymbol u}_Y)+f_Y^2({\boldsymbol H},{\boldsymbol u}_Y)$ and $\mathbb{E}[f_Y^2({\boldsymbol H},{\boldsymbol u}_Y)|Z]=0$. This separability is a strong assumption since all covariates and the treatment must be separable from all unmeasured confounders.
%This integral equation is widely used in the field of machine learning \citep{Hartford2017,Singh2019,Muandet2020}.
%Previously, two integral equations have been introduced when ${\boldsymbol W}$ is an empty set.
%First, an integral equation for estimating $\mathbb{E}[Y_x]$ have been introduced by \citep{Whitney2003}, which is $$\mathbb{E}[Y|Z=z]=\int_{\Omega_X} \mathbb{P}(X=x|Z=z)\mathbb{E}[Y_x]dx$$ 
%under assumptions that $f_Y(X,{\boldsymbol H})=f_Y^1(X)+f_Y^2({\boldsymbol H})$ and $\mathbb{E}[f_Y^2({\boldsymbol H})|Z]=0$, where $\mathbb{P}(X=x|Z=z)$ is a conditional density function, which is widely used in machine learning fields \citep{Hartford2017,Singh2019,Muandet2020}. 
Recently, \citet{Wong2022} introduced an integral equation for identifying APCE $\mathbb{E}[\partial_x Y_x]:=\mathbb{E}_{\boldsymbol U}[\partial_x Y_x({\boldsymbol U})]$ under the IV model with no covariates ${\boldsymbol W}$: %using an IV $Z$:
$\displaystyle \mathbb{E}[Y|Z=z]-\mathbb{E}[Y|Z=z_0]=-\int_{\Omega_X}\{\mathbb{P}(X\leq x|Z=z)-\mathbb{P}(X\leq x|Z=z_0)\}\mathbb{E}[\partial_x Y_x]dx$. 
\citet{Kawakami2023} has developed parametric (P-APCE) and Picard iteration-based (N-APCE) estimators for APCE. In this paper, we extend their results and develop three families of methods for estimating CACPE $\mathbb{E}[\partial_x Y_{x}|{\boldsymbol w}]$. Our parametric estimator reduces to P-APCE  when $\boldsymbol{W}$ is empty. The sieve and RKHS estimators in this paper were not provided in \citep{Kawakami2023}. We note that Picard-iteration estimator in \citep{Kawakami2023} is not suitable here because  equation~(\ref{IE6}) uses a PDF in the integral kernel instead of a CDF in the  equation for APCE.

%due to the use of a density function in the integral kernel instead of a cumulative distribution function (CDF).
%Especially, sieve and RKHS-based estimators in this paper are not in \citep{Kawakami2023}, even for estimating APCE.
%without any subject's covariates, ${\boldsymbol W}$.
%APCE is sufficient if we compare outcomes of the two treatments $\mathbb{E}[Y_{x''}]-\mathbb{E}[Y_{x'}]$.
%This paper will introduce conditions for directly identifying average partial causal effects (CAPCE) with the subject's covariates under weaker assumption than \citep{Whitney2003}. 

%Due to space constraints, all the proofs are provided in the Appendix in the supplementary material. 
%All the proofs are provided in the Appendix.

\section{Identification of CAPCE}

%First, we introduce a new concept \emph{conditional average partial causal effect (CAPCE)} to capture the heterogeneous causal effects of a continuous treatment.
First, we formally define \emph{conditional average partial causal effect (CAPCE)} to capture the heterogeneous causal effects of a continuous treatment. 
Then we present a theorem for identifying CAPCE under the IV model. 
\begin{definition}[CAPCE]
$\displaystyle \mathbb{E}[\partial_x Y_x|{\boldsymbol w}]:=\mathbb{E}_{\boldsymbol U}\left[\frac{\partial}{\partial x} Y_{x}({\boldsymbol U})\Big|{\boldsymbol W}={\boldsymbol w}\right]$.
\end{definition}
%{We also denote $\mathbb{E}[Y_x|{\boldsymbol w}]:=\mathbb{E}_{\boldsymbol U}[ Y_{x}({\boldsymbol U})|{\boldsymbol W}={\boldsymbol w}]$.}
CAPCE is a real-valued function from $x \in \Omega_X$ and ${\boldsymbol w} \in \Omega_{\boldsymbol W}$ to $\mathbb{R}$.
It is a generalization of CACE for continuous treatment. %conditional average causal effect $\mathbb{E}[Y_1-Y_0|{\boldsymbol W}={\boldsymbol w}]$ \citep{Athey2016,Kunzel2019}.
It is also a generalization of APCE $\mathbb{E}[\partial_x Y_x]$ 
 %$:=\mathbb{E}_{\boldsymbol U}[\frac{\partial}{\partial x}Y_x({\boldsymbol U})]$ have been introduced by 
 %\citep{Chamberlain1984,Wooldridge2005,Graham2012} 
 to represent heterogeneous causal effects. 
Next, we present conditions for identifying CAPCE under the IV model. %using IVs from distributions $\mathbb{P}(X, {\boldsymbol W}|Z)$ and $\mathbb{P}( Y |Z)$.
%$\mathbb{P}(X, Y ,{\boldsymbol W}|Z)$.
%We assume the following conditions:
\begin{assumption}
Under the SCM ${\cal M}_{IV}$, given $ {\boldsymbol W}={\boldsymbol w}$, 
\label{AS1}
\begin{enumerate}
\vspace{-0cm}
  \setlength{\parskip}{0.cm}
  \setlength{\itemsep}{0.25cm}
   \item Instrument relevance: IV $Z$ has a causal effect on $X$, i.e., $\mathbb{E}[X_{z}]$ is not a constant function of $z$.
   %the instrument $Z$ has a causal effect on $X$, i.e., $X_{z}$ is not a constant function by varying $z$ for each subject.}
    \item %the potential outcome 
    $Y_{x}$ is differentiable and bounded in $x \in \Omega_X$.
    \item %the potential outcome 
    %$X_{z}$ is not a zero function \yuta{by varying $z$ for each subjects}, and 
$\displaystyle \sup_{x,z,{\boldsymbol w}}\mathfrak{p}(X_{z}=x|{\boldsymbol W}={\boldsymbol w}) < \infty$. 
    %, where $p$ denotes the density function.
    \item {The set of distributions $\mathbb{P}(X|Z=z,{\boldsymbol W}={\boldsymbol w})$} induced by varying $z$ is a complete set.
\vspace{-0.2cm}
\end{enumerate}
\end{assumption}
%These assumptions are related to the conditions for identifying APCE \citep{Wong2022}, and 
%These assumptions are needed just to set up the model and are not restrictive. 
The first assumption is standard for the IV setting.
The second assumption means that there exists CAPCE for all subjects for $x \in \Omega_X$ and ${\boldsymbol w} \in \Omega_{\boldsymbol W}$. 
The third assumption means the density function of $X_{z,{\boldsymbol w}}$ is bounded.
The fourth assumption implies that $h$ is a zero function if $\mathbb{E}[h(X)|Z=z,{\boldsymbol W}={\boldsymbol w}]$ does not depend on $z$ for all ${\boldsymbol w} \in \Omega_{\boldsymbol W}$, which is also assumed in \citep{Whitney2003} for identifying $\mathbb{E}[Y_{x}|{\boldsymbol w}]$.

%Next, we assume the following condition:
\begin{assumption}[Separability on $X$]
\label{AS2}
$f_Y(X,{\boldsymbol W},{\boldsymbol H},{\boldsymbol u}_Y)$ is in the form of a summation of two functions over $X$ and ${\boldsymbol H}$ separately, i.e., 
    $f_Y(X,{\boldsymbol W},{\boldsymbol H},{\boldsymbol u}_Y)=f_Y^1(X,{\boldsymbol W},{\boldsymbol u}_Y)+f_Y^2({\boldsymbol W},{\boldsymbol H},{\boldsymbol u}_Y)$.
\end{assumption}
We obtain the following result. 
\begin{theorem}[Identification of CAPCE]
\label{TEO2}
Under SCM ${\cal M}_{IV}$ and Assumptions \ref{AS1} and \ref{AS2}, CAPCE $\mathbb{E}[\partial_x Y_{x}|{\boldsymbol w}]$ is identifiable from distributions $\mathbb{P}(X, {\boldsymbol W}|Z)$ and $\mathbb{P}( Y |Z)$ via the  integral equation:
%\vspace{-0.6cm}
\begin{eqnarray}
\label{IE6}
\mu(z)=\int_{\Omega_{\boldsymbol W}}\int_{\Omega_X} k(z,x,{\boldsymbol w})\mathbb{E}[\partial_x Y_{x}|{\boldsymbol w}] dxd{\boldsymbol w},
\end{eqnarray}
where $\mu(z)=\mathbb{E}[Y|Z=z_0]-\mathbb{E}[Y|Z=z], k(z,x,{\boldsymbol w})=\mathfrak{p}(X\leq x,{\boldsymbol W}={\boldsymbol w}|Z=z)-\mathfrak{p}(X\leq x,{\boldsymbol W}={\boldsymbol w}|Z=z_0)$, and $z_0$ is a {arbitrary} fixed value.
%\begin{eqnarray}
%\label{EQ3}
%\begin{array}{l}
%    \mu(z)=\mathbb{E}[Y|Z=z_0]-\mathbb{E}[Y|Z=z],\\
%    k(z,x,{\boldsymbol w})=\mathfrak{p}(X\leq x,{\boldsymbol W}={\boldsymbol w}|Z=z)-\mathfrak{p}(X\leq x,{\boldsymbol W}={\boldsymbol w}|Z=z_0).
%    \end{array}
%\end{eqnarray}
\end{theorem}
\textbf{Remark:} Assumption~\ref{AS2} is weaker than the assumption (\ref{eq-sep}) needed by existing work sieve NTSLS \citep{Whitney2003}, PTSLS \citep{Wooldridge2010}, and Kernel IV  \citep{Singh2019} for identifying $\mathbb{E}[Y_{x}|{\boldsymbol w}]$, which require both covariates $\boldsymbol{W}$ and the treatment $X$ to be separable from the unmeasured confounders $\boldsymbol{H}$. Assumption~\ref{AS2} is particularly less restrictive  when there are many covariates. Theorem~\ref{TEO2} states that \emph{CAPCE $\mathbb{E}[\partial_x Y_{x}|{\boldsymbol w}]$ is identifiable under a weaker assumption than required by $\mathbb{E}[Y_x|{\boldsymbol w}]$.} The result enables us to compute causal effects in IV models where Assumption~\ref{AS2} holds but assumption (\ref{eq-sep}) does not such that existing methods are not applicable. Theorem~\ref{TEO2} extends the results in \citep{Wong2022,Kawakami2023} for identifying APCE  $\mathbb{E}[\partial_x Y_{x}]$; however, it is worth noting that this important point about weaker separability assumption does not arise in the work of \citet{Wong2022} and \citet{Kawakami2023} because they study the setting with no covariates $\boldsymbol{W}$.

%The integral equation (\ref{IE6})  is a ``Fredholm Integral Equation of the First Kind” with $k$ called an integral kernel \citep{Bocher1926}. Equation (\ref{IE6}) is ill-posed since the integral operator ${\cal K}$ is not guaranteed to be compact,  where ${\cal K}(f)(z)=\int_{\Omega_{\boldsymbol W}}\int_{\Omega_X} k(z,x,{\boldsymbol w})f(x,{\boldsymbol w}) dxd{\boldsymbol w}$.
%A well-posed problem satisfies the following three properties \citep{Tikhonov1995}: the solution's existence, uniqueness, and stability.
%Problems, where one or more of these conditions do not hold, are called ill-posed problems.

\section{Estimation of  CAPCE}

In this section, we develop three families of methods for estimating CAPCE from  data based on Theorem \ref{TEO2}.  We do not need 
%i.i.d. 
samples from the joint $\mathfrak{p}(Z,X,Y,{\boldsymbol W})$, %single dataset ${\cal D} = \{y_i,x_i,{\boldsymbol w}_i,z_i\}_{i=1}^N$, 
but rather two datasets ${\cal D}^{(1)} = \{x^{(1)}_i,{\boldsymbol w}^{(1)}_i,z^{(1)}_i\}_{i=1}^{N_1}$ and ${\cal D}^{(2)} = \{y^{(2)}_i,z^{(2)}_i\}_{i=1}^{N_2}$ known as two-samples IV methods \citep{Singh2019,Angrist1992}.
%The integral equation (\ref{IE6}) holds for each individual IV $Z$ in the $M$-dimensional IV $\boldsymbol{Z} = \{Z^1,\ldots, Z^M\}$.  We note that   using many valid  IVs can improve the precision of estimation \citep{Hansen2008}. We will usually derive results for an individual IV $Z$ 

%\yuta{We note that using valid many IVs can improve precision of estimation \citep{Hansen2008}.}
%While only one IV is sufficient for CAPCE identification, multiple IVs can be available for CAPCE estimation. 
%\jin{Unclear what "can be available" means. Multiple IVs are more helpful for estimation?} \jin{We assume multiple IVs $Z^1,\ldots,Z$ such that for $m=1, \dots, M$, (2) holds for $Z=Z$ - Your vector $Z$ notation gives the impression of replacing Z in (2) with a vector Z.} \yuta{[I've fixed Theorem 1 by replacing Z in (2) with a vector Z.]}

%\jin{Why not just stay with a single IV variable? Section 4.1 is much easier to read with the index m dropped. In fact, you could derive the result with a single IV, then extend the results to multiple IVs in the end.}

\subsection{Sieve CAPCE Estimator}
%Next, we introduce the \yuta{sieve CAPCE estimator}.
Sieve estimators are a class of non-parametric estimators that use progressively more complex models to estimate an unknown function as more data becomes available \citep{Geman1982}. 
%In this section we use $W^{l,2}$ norm $(0\leq l \leq \infty)$.
%, which is introduced by \citep{Whitney2003} and \citep{Ai2003}
%, and we can estimate CAPCE directly.
%We consider the following model:
%\footnotesize
%\setlength{\abovedisplayskip}{0pt}
%\begin{equation}
%\label{EQ1}
%\mathbb{E}[\partial_x Y_{x,{\boldsymbol w}}]=g(x,{\boldsymbol w},{\boldsymbol \pi})
%\end{equation}

{\bf Approximation by Orthonormal Basis Functions.} We approximate the CAPCE $\mathbb{E}[\partial_x Y_{x}|{\boldsymbol w}]$ by a set of orthonormal basis functions, such as Hermite polynomial functions \citep{Hermite2009}. Specifically, 
%we approximate CAPCE $g_0 \in {\cal G}$ by
%Let $L$ dimensional vector $x,{\boldsymbol w}=(x,{\boldsymbol w}^T)^T$.
%\footnotesize
%\setlength{\abovedisplayskip}{0pt}
%\vspace{-0.6cm}
\begin{align}
\label{eq-orth-basis}
  %\mathbb{E}[\partial_x Y_{x,{\boldsymbol w}}] =  \sum_{j=1}^J\beta_{j} \psi_j(x,{\boldsymbol w})+g_0(x,{\boldsymbol w}) =  \sum_{j=1}^J\beta_{j} \psi_j(x,{\boldsymbol w})+\sum_{k=1}^{\infty}\gamma_{k} \sigma_k(x,{\boldsymbol w}),
   \mathbb{E}[\partial_x Y_{x}|{\boldsymbol w}] \equiv g_0(x,{\boldsymbol w}) \approx g(x,{\boldsymbol w})= \sum_{j=1}^{J}\beta_{j} \phi_j(x,{\boldsymbol w}),
\end{align}
where %${\cal G}$ is a function space, and 
$\displaystyle \{\phi_j(x,{\boldsymbol w})\}_{j=1}^{\infty}$ is a set of infinite basis functions that satisfy the following conditions where Sobolev norm $W^{l,2}$ norm $(0\leq l \leq \infty)$ is used:
\begin{assumption}
\label{B1}
The basis functions $\displaystyle \{\phi_j(x,{\boldsymbol w})\}_{j=1}^{\infty}$ are orthonormal basis functions, and satisfy $\|\phi_j(x,{\boldsymbol w})\|_{W^{l,2}} < \infty$ for all $j=1,2,\ldots.$
\end{assumption}
%Let parameters be ${\boldsymbol \pi}=({\boldsymbol \beta}^T,g_0)$, where ${\boldsymbol \pi} \in {\boldsymbol \pi}$.
%This model is nonparametric if ${\boldsymbol \beta}$ is null; on the other hand, this model is parametric if $g_0$ is null.
%We call CAPCE estimator a \textbf{parametric CAPCE (P-CAPCE) estimator} if $g_0$ is null, and a (nonparametric) \textbf{sieve CAPCE (S-CAPCE) estimator} if ${\boldsymbol \beta}$ is null.

%We approximate CAPCE by $\sum_{j=1}^J\beta_{j} \phi_j(x,{\boldsymbol w})$. Let ${\boldsymbol \beta} =(\beta_1,\ldots,\beta_J)^T$.
%\normalsize
%We assume the following condition: 
\begin{assumption}
\label{AS3}
$\displaystyle \sum_{j=1}^J\beta_j\phi_j(x,{\boldsymbol w})$ convergences uniformly to $g_0(x,{\boldsymbol w})$ if $J \rightarrow \infty$. 
\end{assumption}
%In this section we use Sobolev norm $W^{l,2}$ norm $(0\leq l \leq \infty)$.
%\yuta{From Assumption \ref{AS3}, the interchange of integration and limit
%\footnotesize
%\begin{eqnarray}
%\int_{\Omega_{\boldsymbol W}}\int_{\Omega_X}k(z,x,{\boldsymbol w})\lim_{J \rightarrow \infty}\sum_{j=1}^J\beta_{j}\phi_j(x,{\boldsymbol w}) dxd{\boldsymbol w}=\lim_{J \rightarrow \infty}\sum_{j=1}^J\gamma_{j}\int_{\Omega_{\boldsymbol W}}\int_{\Omega_X}k(z,x,{\boldsymbol w})\phi_j(x,{\boldsymbol w}) dxd{\boldsymbol w} 
%\end{eqnarray}
%\normalsize
%holds.} 
We note that Hermite polynomial functions satisfy Assumption~\ref{AS3} for any bounded and continuous function $g_0$ \citep{Damelin2001}.

%converge uniformly to any $g_0$ if $J \rightarrow \infty$ \citep{Damelin2001}.
%Hermite polynomial functions are $\sigma_k(x,{\boldsymbol w})=exp(-x,{\boldsymbol w}^Tx,{\boldsymbol w})x,{\boldsymbol w}^{{\boldsymbol \kappa}(k)}$, where $\|{\boldsymbol \kappa}(k)\|$ is increasing in $k$.
%Here, ${\boldsymbol \kappa}$ is a vector of non-negative integers and $\|{\boldsymbol \kappa}\|=\sum_{l=1}^L \kappa_l$ and $x,{\boldsymbol w}^{\boldsymbol \kappa}=\prod_{i=1}^L x^{\kappa_i}$.
%We denote parameter space ${\boldsymbol \pi}_K=({\boldsymbol \beta}^T,\sum_{k=1}^K\gamma_{k} \sigma_k(x,{\boldsymbol w}))$, \jin{why $\sigma_k$ here?} which is a subspace of ${\boldsymbol \pi}$.
%using the basis functions $\{\phi_i(x,{\boldsymbol w})\}_{p=1,\ldots,P}$ \citep{Bishop2006}, where ${\boldsymbol \beta}=\{\beta_{1},\ldots,\beta_{i}\}$ are the model parameters to be estimated from data. 
%For example, $\phi_j(x,{w})=\beta_0+ \beta_1 x +\beta_2w +\beta_3 x^2+ \beta_4xw+ \beta_5 w^2+\cdots.$
%This is a generalization of a linear regression that replaces each explanatory variable with some appropriate functions.

%The integral equation (\ref{IE6}) holds for each individual IV $Z$ in the $M$-dimensional IV $\boldsymbol{Z} = \{Z^1,\ldots, Z^M\}$. We note that   using many valid  IVs can improve the precision of estimation \citep{Hansen2008}. 
%{\bf Restriction for Compactness.} The restriction on ${\boldsymbol \beta}$ and ${\boldsymbol \gamma}$ are used to imposed compactness.
%\jin{Should "u" be "c"? What happened to $\gamma$, or should $\beta$ be $\pi$? }

{\bf Compactness Restriction.} 
The integral equation (\ref{IE6}), known as   a ``Fredholm Integral Equation of the First Kind” %with $k$ called an integral kernel 
\citep{Bocher1926}, 
 is ill-posed  since the integral operator ${\cal K}$,  where $\displaystyle {\cal K}(f)(z)=\int_{\Omega_{\boldsymbol W}}\int_{\Omega_X} k(z,x,{\boldsymbol w})f(x,{\boldsymbol w}) dxd{\boldsymbol w}$, is not guaranteed to be compact. 
Problems where one or more of the three properties - existence, uniqueness, and stability of the solution - do not hold are called ill-posed problems \citep{Tikhonov1995} and  lead to severe estimation difficulties. {To relieve the issue, we put restrictions on the functional space of $g_0(x,{\boldsymbol w})$.}
%\yuta{We restrict the integral operator ${\cal K}$ to be in the compact set since the integral equation (\ref{IE6}) is ill-posed.} \citep{Whitney2003} introduced nonparametric compact restriction using Sobolev norm \citep{Gallant1987}. 
Let $\displaystyle \mathfrak{g}(X,{\boldsymbol W})=\int_{\Omega_{\boldsymbol W}}\int_{\Omega_X}\{\mathbbm{1}_{X\leq x,{\boldsymbol W}={\boldsymbol w}}-\mathbb{E}[\mathbbm{1}_{X\leq x,{\boldsymbol W}={\boldsymbol w}}|Z=z_0]\}g(X,{\boldsymbol W})dxd{\boldsymbol w}$, and define regularized Sobolev norm $\tilde{W}^{l,2}$, which is  called ``consistency norm" in \citep{Gallant1987}, as follows
\begin{equation}
\begin{aligned}
\label{RES1}
    &\left\|\mathfrak{g}(x,{\boldsymbol w})\right\|_{\tilde{W}^{l,2}}^2=\sum_{|\lambda|\leq l} \int \left\{D^{\lambda}\mathfrak{g}(x,{\boldsymbol w})\right\}^2\\
    &\hspace{1.5cm}\times \{1+(x,{\boldsymbol w}^T)(x,{\boldsymbol w}^T)^T\}^{\kappa}  dxd{\boldsymbol w},  
    \end{aligned}
\end{equation}
where $l\ge 1$ is an integer and $\kappa$ is a constant satisfying $\kappa>(1+d)/2$ where $d$ is the dimension of ${\boldsymbol W}$.
We make  the following assumption:
\begin{assumption}
\label{COM}
Given a positive regularization parameter $B_S$, $g_0(x,{\boldsymbol w})$ is in the functional space ${\cal G}_{B_S}=\{g:\|\mathfrak{g}(x,{\boldsymbol w})\|_{\tilde{W}^{l,2}}^2 \leq B_S\}$.    
\end{assumption}

%The relationship ${\boldsymbol \pi}_B \subset \{{\boldsymbol \pi}:{\boldsymbol \beta}^T{\boldsymbol \beta}\leq B_{\beta},\|\mathfrak{g}_0(x,{\boldsymbol w})\|_{\tilde{W}^{l,2}}^2 \leq B_{\gamma}\}$ holds for all $m=1,\ldots,M$.

Using the approximation in (\ref{eq-orth-basis}), equation (\ref{IE6}) reduces to 

\vspace{-0.6cm}
%\footnotesize
\begin{align}
%\mu(z)=\sum_{j=1}^J\beta_{j}\int_{\Omega_{\boldsymbol W}}\int_{\Omega_X}k(z,x,{\boldsymbol w})\phi_j(x,{\boldsymbol w}) dxd{\boldsymbol w}+\sum_{k=1}^K\gamma_{k}\int_{\Omega_{\boldsymbol W}}\int_{\Omega_X}k(z,x,{\boldsymbol w})\sigma_k(x,{\boldsymbol w}) dxd{\boldsymbol w}
\mu(z)=\sum_{j=1}^J\beta_{j}\int_{\Omega_{\boldsymbol W}}\int_{\Omega_X}k(z,x,{\boldsymbol w})\phi_j(x,{\boldsymbol w}) dxd{\boldsymbol w}.
\end{align}
\normalsize
Letting the anti-derivative of the basis functions be $\displaystyle \varphi_j(x,{\boldsymbol w})=\int \phi_j(x,{\boldsymbol w})dx$.% for $j=1,\ldots,J$.
\footnote{We will simply write the antiderivative $\displaystyle \varphi_j(x,{\boldsymbol w})=\int_{-\infty}^x \phi_j(x',{\boldsymbol w})dx'$ as $\varphi_j(x,{\boldsymbol w})=\int \phi_j(x,{\boldsymbol w})dx$ in the paper.} % because the constant of integration is irrelevant since we take the difference between the antiderivatives.}  % where constant of integration is $0$.
Then, the equation becomes
\begin{equation}
\label{EQ7}
\begin{aligned}
&\mathbb{E}[Y|Z=z]-\mathbb{E}[Y|Z=z_0]=\\
&\sum_{j=1}^J\beta_{j}\{\mathbb{E}[\varphi_j(X,{\boldsymbol W})|Z=z]-\mathbb{E}[\varphi_j(X,{\boldsymbol W})|Z=z_0]\}%\\
%&&\hspace{4cm}+\sum_{k=1}^K\gamma_{k}\{\mathbb{E}[\pi_k(X,{\boldsymbol W})|Z=z]-\mathbb{E}[\pi_k(X,{\boldsymbol W})|Z=z_0]\}.
\end{aligned}
\end{equation}
%Next, we show that the estimation problem reduces to a linear equation. 
Let 
%\begin{eqnarray}
%\begin{array}{l}
${c}=\mathbb{E}[Y|Z=z]-\mathbb{E}[Y|z=z_0]$, ${\boldsymbol \beta} =(\beta_1,\ldots,\beta_J)^T$, 
and ${\boldsymbol d}=({d}^{1},\ldots,{d}^{J})^T$ where 
${d}^{j}=\mathbb{E}[\varphi_j(X,{\boldsymbol W})|Z=z]-\mathbb{E}[\varphi_j(X,{\boldsymbol W})|Z=z_0]$. %and ${e}^{k}=\mathbb{E}[\pi_k(X,{\boldsymbol W})|Z=z]-\mathbb{E}[\pi_k(X,{\boldsymbol W})|Z=z_0]$
%\end{array}
%\end{eqnarray}
%for $j=1,\ldots,J$. %, $k=1,\ldots,K$ and $m=1,\ldots,M$.
%Furthermore, denote %${\boldsymbol f}=({d}^{m,1},\ldots,{d}^{j},{e}^{m,1},\ldots,{e}^{k})^T$
%${\boldsymbol d}=({d}^{1},\ldots,{d}^{J})^T$.
Then, the integral equation (\ref{IE6}) finally reduces to a linear equation $c={\boldsymbol \beta}^T{\boldsymbol d}$.

{\bf Sieve CAPCE (S-CAPCE) estimator.} Given datasets ${\cal D}^{(1)} = \{x^{(1)}_i,{\boldsymbol w}^{(1)}_i,z^{(1)}_i\}_{i=1}^{N_1}$ and ${\cal D}^{(2)} = \{y^{(2)}_i,z^{(2)}_i\}_{i=1}^{N_2}$, 
%We explain CAPCE estimation method from observations. 
%We do not need single dataset ${\cal D} = \{y_i,x_i,{\boldsymbol w}_i,z_i\}_{i=1}^N$, but rather two datasets ${\cal D}^{(1)} = \{x^{(1)}_i,{\boldsymbol w}^{(1)}_i,z^{(1)}_i\}_{i=1}^{N_1}$ and ${\cal D}^{(2)} = \{y^{(2)}_i,z^{(2)}_i\}_{i=1}^{N_2}$ as two-samples IV estimator \citep{Singh2019,Angrist1992}.
%, where $z^{(1)}_i=(z^{(1)}_1,\ldots,z^{(1)}_1)$.
%let $N=N_1+N_2$ and  $(z_1,\ldots,z_N)=(z_1^{(1)},\ldots,z_{N_1}^{(1)},z_1^{(2)},\ldots,z_{N_2}^{(2)})$. \\ %for $m=1,\ldots,M$, 
our S-CAPCE estimator consists of two stages. 
{In Stage 1, we learn models $\hat{\mathbb{E}}[Y|Z=z]$ and $\hat{\mathbb{E}}[\varphi_j(X,{\boldsymbol W})|Z=z]$  from the datasets by regression. 
Then in Stage 2, we estimate parameters ${\boldsymbol \beta}$ by solving Eq. (\ref{EQ7})}.
%$\tildex,{\boldsymbol w}=\hat{\Sigma}^{1/2}(x,{\boldsymbol w}-\overlinex,{\boldsymbol w})$, where $\hat{\Sigma}$ and $\overlinex,{\boldsymbol w}$ are the variance and mean of $x,{\boldsymbol w}$.
%There are four steps in Algorithm \ref{alg1}.

\noindent{\bf Stage 1.} We learn  prediction models %based on covariates 
$\hat{\mathbb{E}}[Y|Z=z]$
%for $m=1,\ldots,M$ 
using ${\cal D}^{(2)}$ and
%Explanatory variables are only IV, $Z$, and response variables are the outcome, $Y$.
%create prediction models 
$\hat{\mathbb{E}}[\varphi_j(X,{\boldsymbol W})|Z=z]$ for $j=1,\ldots,J$
%and $m=1,\ldots,M$, and $\hat{\mathbb{E}}[\varsigma_k(X,{\boldsymbol W})|Z=z]$ for $k=1,\ldots,K$ and $m=1,\ldots,M$
using ${\cal D}^{(1)}$. Any regression method can be used. We select an IV value $z_0$. 
Denote $\hat{c}_i=\hat{\mathbb{E}}[Y|Z=z_i]-\hat{\mathbb{E}}[Y|Z=z_0]$ and $\hat{d}_i^{j}=\hat{\mathbb{E}}[\varphi_j(X,{\boldsymbol W})|Z=z_i]-\hat{\mathbb{E}}[\varphi_j(X,{\boldsymbol W})|Z=z_0]$. % for $i=1,\ldots,N$ and $j=1,\ldots,J$.
%for each IV $Z \in \{Z^1,\ldots,Z^M\}$.
%We can use any machine learning method to estimate conditional expectations.

Specifically, we perform the regression using the power series basis functions in this paper. %\yuta{such as power series or splines}. \jin{What basis functions are actually used?}
%For each IV $Z \in \{Z^1,\ldots,Z^M\}$, 
Let basis functions be ${\boldsymbol q}(z)=(q_1(z),q_2(z),\ldots,q_P(z))^T$, and consider the model $\hat{\mathbb{E}}[Y|Z=z]=\sum_{p=1}^P \omega_p q_p(z)$, $\hat{\mathbb{E}}[\varphi_j(X,{\boldsymbol W})|Z=z]=\sum_{p=1}^P \nu_p^j q_p(z)$ for $j=1,\ldots,J$. Denote ${\boldsymbol \omega}=(\omega_1,\ldots,\omega_P)^T$ and  ${\boldsymbol \nu}^j=(\nu_1^j,\ldots,\nu_P^j)^T$.
Then, we optimize the error functions below:
\begin{equation}
\begin{aligned}
    &Q_1({\boldsymbol \nu}^j;{\cal D}^{(1)})\\
    &\hspace{0.5cm}=\frac{1}{N_1}\sum_{i=1}^{N_1}(\varphi_j(x_i^{(1)},{\boldsymbol w}^{(1)}_i)-{\boldsymbol q}(z_i^{(1)})^T{\boldsymbol \nu}^j)^2,
    \end{aligned}
\end{equation}
\begin{equation}
\begin{aligned}
    &Q_2({\boldsymbol \omega};{\cal D}^{(2)})=\frac{1}{N_2}\sum_{i=1}^{N_2}(y_i^{(2)}-{\boldsymbol q}(z_i^{(2)})^T{\boldsymbol \omega})^2.
\end{aligned}
\end{equation}
Let variance-covariance matrices be $\hat{\bf M}^{(1)}=\sum_{i=1}^{N_1} N_1^{-1}{\boldsymbol q}(z^{(1)}_i){\boldsymbol q}(z^{(1)}_i)^T$ and  $\hat{\bf M}^{(2)}=\sum_{i=1}^{N_2} N_2^{-1}{\boldsymbol q}(z^{(2)}_i){\boldsymbol q}(z^{(2)}_i)^T$. 
%for $m=1,\ldots,M$.
We obtain %the following prediction values %$\hat{c}_i=\hat{\mathbb{E}}[Y|Z=z_i]-\hat{\mathbb{E}}[Y|Z=z_0]$, $\hat{d}_i^{j}=\hat{\mathbb{E}}[\varphi_j(X,{\boldsymbol W})|Z=z_i]-\hat{\mathbb{E}}[\varphi_j(X,{\boldsymbol W})|Z=z_0]$ for $i=1,\ldots,N$ and $j=1,\ldots,J$.
%and $\hat{e}_i^{k}=\hat{\mathbb{E}}[\varsigma_k(X,{\boldsymbol W})|Z=z_i]-\hat{\mathbb{E}}[\varsigma_k(X,{\boldsymbol W})|Z=z_0]$ for $i=1,\ldots,N$, $m=1,\ldots,M$, $j=1,\ldots,J$, and $k=1,\ldots,L$. These values can be calculated as below:
%\small
\begin{equation}
\label{eq-pred}
\left\{
\begin{array}{l}
\renewcommand{\arraystretch}{1}
    %&&c_i=\hat{\mathbb{E}}[Y|Z=z_i]-\hat{\mathbb{E}}[Y|Z=z_0]=({\boldsymbol q}(z_i)-{\boldsymbol q}(z_0))^T\hat{\bf M}^{-}\sum_{l=1}^N \frac{1}{N} {\boldsymbol q}(z_l)y_l\\
    \hat{c}_i%=\hat{\mathbb{E}}[Y|Z=z_i]-\hat{\mathbb{E}}[Y|Z=z_0]
    =({\boldsymbol q}(z_i)-{\boldsymbol q}(z_0))^T\hat{\bf M}^{(2)-}\sum_{l=1}^{N_2} \frac{1}{N_2} {\boldsymbol q}(z^{(2)}_l)y^{(2)}_l\\
    \hat{d}_i^{j}%=\hat{\mathbb{E}}[\varphi_j(X,{\boldsymbol W})|Z=z_i]-\hat{\mathbb{E}}[\varphi_j(X,{\boldsymbol W})|Z=z_0]
    =({\boldsymbol q}(z_i)-{\boldsymbol q}(z_0))^T\hat{\bf M}^{(1)-}\\
    \hspace{2cm}\times\sum_{l=1}^{N_1} \frac{1}{N_1} {\boldsymbol q}(z^{(1)}_l)\varphi_j(x^{(1)}_l,{\boldsymbol w}^{(1)}_l)\\
    %\hat{e}_i^{k}%=\hat{\mathbb{E}}[\varsigma_k(X,{\boldsymbol W})|Z=z_i]-\hat{\mathbb{E}}[\varsigma_k(X,{\boldsymbol W})|Z=z_0]
    %=({\boldsymbol q}(z_i)-{\boldsymbol q}(z_0))^T\hat{\bf M}^{(1)-}\sum_{l=1}^{N_1} \frac{1}{N_1} {\boldsymbol q}(z^{(1)}_l)\varsigma_k(x^{(1)}_l,{\boldsymbol w}^{(1)}_l)\nonumber
\end{array}
\right.
\end{equation}
for $j=1,\ldots,J$,
%, $m=1,\ldots,M$, $j=1,\ldots,J$, and $k=1,\ldots,L$
where $\hat{\bf M}^{-}$ denotes the generalized inverse that satisfies $\hat{\bf M}\hat{\bf M}^{-}\hat{\bf M}=\hat{\bf M}$.
{Let $N=N_1+N_2$ and  $(z_1,\ldots,z_N)=(z_1^{(1)},\ldots,z_{N_1}^{(1)},z_1^{(2)},\ldots,z_{N_2}^{(2)})$. We will compute predicted values in (\ref{eq-pred})  for all $i=1,\ldots,N$. }

\noindent{\bf Stage 2.} 
%Consider multiple IVs $Z=(Z^1,\ldots,Z^M)$.
Estimate parameters ${\boldsymbol \beta}$ based on the  linear equation $c={\boldsymbol \beta}^T{\boldsymbol d}$. 
 Let $\hat{\boldsymbol c}=(\hat{c}_1,\ldots,\hat{c}_N)^T$, $\hat{\boldsymbol d}_i=(\hat{d}^{1}_i,\ldots,\hat{d}^{J}_i)^T$, 
%and $\hat{\boldsymbol e}^k_i=(\hat{e}^{1,j}_i,\ldots,\hat{e}^{j}_i)^T$.
$\hat{\bf D}=(\hat{\boldsymbol d}_1,\ldots,\hat{\boldsymbol d}_N)^T$, and
%$\hat{\bf E}_i=(\hat{\boldsymbol e}^{1}_i,\ldots,\hat{\boldsymbol e}^{K}_i)$, and $\hat{\bf F}_i=(\hat{\bf D}_i,\hat{\bf E}_i)$.\\
%We denote ${\boldsymbol \delta}=({\boldsymbol \beta}^T,{\boldsymbol \gamma}^T)^T$ and 
the empirical risk be 
\begin{eqnarray}
\label{Q1}
    {Q}_3({\boldsymbol \beta}\ ;{\cal D}^{(1)},{\cal D}^{(2)})=\frac{1}{N}\sum_{i=1}^N(\hat{c}_i-\hat{\boldsymbol d}_i^T{\boldsymbol \beta})^2.
\end{eqnarray}
Under Assumption \ref{COM}, our estimator $\hat{\boldsymbol \beta}$ is given  by the optimization problem below: 
\begin{eqnarray}
\label{OPT1}
    \min_{\boldsymbol \beta}{Q}_3({\boldsymbol \beta}\ ;{\cal D}^{(1)},{\cal D}^{(2)})\text{ subject to } {\boldsymbol \beta}^T{\boldsymbol \Lambda}{\boldsymbol \beta}\leq B_S,
\end{eqnarray}
where %\jin{Is $\Lambda$ something easy to compute? Do you need to describe how to compute $\Lambda$?}
%\vspace{-0.8cm}
\begin{equation}
\begin{aligned}
    &{\Lambda}_{i,j}=\sum_{|\lambda|\leq l} \int \left\{D^{\lambda}{\varphi}_i(x,{\boldsymbol w})-D^{\lambda}\mathbb{E}[{\varphi}_i(X,{\boldsymbol W})|Z=z_0]\right\}\\    &\hspace{0.8cm}\times\left\{D^{\lambda}{\varphi}_j(x,{\boldsymbol w})-D^{\lambda}\mathbb{E}[{\varphi}_j(X,{\boldsymbol W})|Z=z_0]\right\}\\
    &\hspace{3cm}\times\{1+\|(x,{\boldsymbol w}^T)\|^2\}^{\kappa}dxd{\boldsymbol w}
\end{aligned}
\end{equation}
\noindent for $i,j=1,\ldots,J$, and ${\boldsymbol \Lambda}=\{{\Lambda}_{i,j}\}_{i,j=1}^{J}$.
%and ${\boldsymbol \varphi}(x,{\boldsymbol w})=(\varphi_1(x,{\boldsymbol w}),\ldots,\varphi_J(x,{\boldsymbol w}))^T$.
${\boldsymbol \Lambda}$ can be calculated by Monte Carlo integration $\hat{\boldsymbol \Lambda}$ \citep{Kroese2011}. %if ${\Lambda}$ is hard to calculate directly.}\jin{How was $\Lambda$ computed in the experiments?}
%The regularization restriction is defined in \citep{Whitney2003}, which are used to impose compactness of ${\boldsymbol \pi}$.
%$\hat{\bf A}_3$ is a positive definite matrix and is involved in the estimator's variance. 
%In this paper, we do not discuss it. 
%\jin{You must specify what $\hat{\bf A}$ is.}
%\yuta{We follow the procedure proposed in \citep{Ai2003} to determine $\hat{\bf A}_3$; (i) compute identity weighted estimator using ${Q}_3({\boldsymbol \beta};{\cal D})=\sum_{i=1}^N \frac{1}{N}[\hat{\boldsymbol c}_i-\hat{\bf D}_i{\boldsymbol \beta}]^T[\hat{\boldsymbol c}_i-\hat{\bf D}_i{\boldsymbol \beta}]$; (ii) compute the $(i,j)$-th element of $\hat{\bf A}_{3}$ by the covariance of residuals given $Z=z_i$ and residuals given $Z=z_j$ for $i,j=1,\ldots,N$.}
The optimization problem (\ref{OPT1}) can be solved by a ridge regression method with the following solution {\citep{Hilt1977}}: 
%The optimization (\ref{OPT1}) have a ridge regression form when $\hat{\bf A}_3$ is an identity matrix \yuta{\citep{Whitney2003}}.
%and ${\bf S}$ be a diagonal matrix whose diagonal elements are $\lambda\text{diag}[{\boldsymbol \Lambda}]$, where ${\boldsymbol 1}_J$ is $J$-th vector $(1,1,\ldots,1)$ and $\text{diag}[{\boldsymbol \Lambda}]$ is a diagonal vector of ${\boldsymbol \Lambda}$. 
%Then, the estimator is 
%\jin{Are you assuming $\hat{\bf A}$ is an identity matrix here? Are you saying (10) is a ridge regression problem? Citation for the following conclusion? }
\begin{eqnarray}
    \hat{\boldsymbol \beta}=(\hat{\bf D}^T\hat{\bf D}+\zeta_S\text{diag}[{\boldsymbol \Lambda}])^{-1}\hat{\bf D}^T\hat{\boldsymbol c}, 
\end{eqnarray}
where  $\zeta_S$ is a regularization parameter called 
Lagrange multipliers.  
%This is a ridge regression form. \jin{Where does this $\beta$ come from? No clue what you are talking about here.}
%Then, estimator of the parameter $\hat{\boldsymbol \pi}$ is given as $(\hat{\boldsymbol \beta},g(\hat{\boldsymbol \gamma}))$. \jin{What is $g(\gamma)$?}
Then, our proposed sieve CAPCE estimator  is given by $\displaystyle \hat{\mathbb{E}}[\partial_x{Y}_{x}|{\boldsymbol w}]=\sum_{j=1}^{J}\hat{\beta}_{j} \phi_j(x,{\boldsymbol w})$.

{\bf Model Selection.} %Finally, we explain the performance metric of the trained models by Algorithm \ref{alg1}. We have the trained parameters $\hat{\boldsymbol \pi}$ and available test observations ${\cal D'}=\{z'^{(i)},x'^{(i)},y'^{(i)}\}_{i=1}^{N'}$.
The model selection in Stage 1 is a standard regression problem, and we presume the models in Stage 1 have been selected appropriately according to standard machine learning methods. 
We can use the empirical risk in equation~(\ref{Q1}) as a performance metric of the trained model in Stage 2 with parameters $\hat{\boldsymbol \beta}$ if given  separate test datasets %${\cal D'}=\{x'_i,y'_i,z'_i,{\boldsymbol w}'_i\}_{i=1}^{N'}$ or 
${\cal D}^{(1)'}=\{x^{(1)'}_i,z^{(1)'}_i,{\boldsymbol w}^{(1)'}_i\}_{i=1}^{N_1'}$  and ${\cal D}^{(2)'}=\{y^{(2)'}_i,z^{(2)'}_i\}_{i=1}^{N_2'}$.
Let $N'= N_1' + N_2'$. Assume $\hat{c}_i'$ and $\hat{\boldsymbol d}_i'$ for $i=1,\ldots,N'$ are computed using ${\cal D}^{(1)'}$ and ${\cal D}^{(2)'}$.
%By substituting ${\cal D}$ to ${\cal D'}$, we obtain $\hat{\boldsymbol H}'$ and $\hat{\bf D}'$.
Then, we can evaluate the trained model by the  test error  % mean squared error (MSE):
%\begin{eqnarray}
%\label{Q3}
$\displaystyle \hat{Q}_3(\hat{\boldsymbol \beta}\ ;{\cal D}^{(1)'},{\cal D}^{(2)'})=\frac{1}{N'}\sum_{i=1}^{N'}(\hat{c}'_i-\hat{\boldsymbol d}_i^{'T}\hat{\boldsymbol \beta})^2$. 
%\end{eqnarray}
Given separate datasets, this performance metric can  be used for model selection from various candidate  basis functions or the number $J$ or $P$ of basis terms.

{\bf Property of sieve CAPCE estimator.} 
We show that sieve CAPCE estimator is consistent under assumptions similar to sieve NTSLS \citep{Whitney2003}.
%\jin{citation?} %, shown in Appendix.
%The rate of convergence is given under similar assumptions in \citep{Xiaohong2018}, shown in Appendix.
Assumptions B.1 - 4 are shown in Appendix \ref{appB}. 

\begin{theorem}[Consistency]
\label{STHEO1}
    Under SCM ${\cal M}_{IV}$ and Assumptions \ref{AS1}, \ref{AS2},  \ref{B1}, \ref{AS3}, \ref{COM}, \ref{A1}, \ref{A2}, \ref{A3}, and \ref{A6}, 
    %B.1, B.2, B.3, and B.4,
    letting $P \rightarrow \infty$ and $J\rightarrow \infty$, then $\|\hat{g}-g_0\|_{W^{l,\infty}}\xrightarrow{p} 0$.
\end{theorem}
\begin{theorem}[Rate of Convergence]
\label{STHEO2}
    Under SCM ${\cal M}_{IV}$ and Assumptions \ref{AS1}, \ref{AS2},  \ref{B1}, \ref{AS3}, \ref{COM}, \ref{RA1}, \ref{RA2}, \ref{RA4}, \ref{RA5}, \ref{RA6}, \ref{RA7}, and \ref{RA8},
    %C.1, C.2, C.3, C.4, C.5, C.6, and C.7,
    setting $N=N_1=N_2$, then $\|\hat{g}-{g}_0\|_{A}={o}_p(N^{-1/4})$.
\end{theorem}
{Assumpts. C.1-7 and norm $\|\cdot\|_A$ are defined in Appendix~\ref{appC}.} %$o_p$ is order in probability notation \citep{Bishop1975}.

\subsection{Parametric CAPCE estimator}
Next, we develop a parametric CAPCE (P-CAPCE) estimator.
We consider the setting that the CAPCE $\mathbb{E}[\partial_x Y_{x}|{\boldsymbol w}]$ takes the form of  the following parametric model: 
\begin{equation}
 \mathbb{E}[\partial_x Y_{x}|{\boldsymbol w}]=   \sum_{k=1}^K\gamma_k\theta_k(x,{\boldsymbol w}), 
\end{equation}
where $\displaystyle\{\theta_k(x,{\boldsymbol w})\}_{k=1}^{K}$ are a set of known  functions, and  ${\boldsymbol \gamma}=(\gamma_1,\ldots,\gamma_K)^T$ are unknown model parameters to be estimated from data. 
%Let ${\boldsymbol \gamma}_0$ be the parameters which satisfies $\mathbb{E}[\partial_x Y_{x}|{\boldsymbol w}]=\sum_{k=1}^K\gamma_{0 k}\theta_k(x,{\boldsymbol w})$. \jin{Why are you introducing $\gamma_0$? The distinction between $\gamma$ and $\hat{\gamma}$ is not enough?}

The derivation of the P-CAPCE estimator is very similar to that of the sieve CAPCE estimator, so we skip the details in the following. Denote the anti-derivatives $\displaystyle \vartheta_k(x,{\boldsymbol w})=\int\theta_k(x,{\boldsymbol w})dx$ for $k=1,\ldots,K$. 
%The integral equation (\ref{IE6}) becomes 
%\begin{eqnarray}
%\mathbb{E}[Y|Z=z]-\mathbb{E}[Y|Z=z_0]=\sum_{k=1}^K\gamma_{k}\{\mathbb{E}[\vartheta_k(X,{\boldsymbol W})|Z=z]-\mathbb{E}[\vartheta_k(X,{\boldsymbol W})|Z=z_0]\}%\nonumber\\
%&&\hspace{4cm}+\sum_{k=1}^K\gamma_{k}\{\mathbb{E}[\pi_k(X,{\boldsymbol W})|Z=z]-\mathbb{E}[\pi_k(X,{\boldsymbol W})|Z=z_0]\}.
%\end{eqnarray}
%Next, we show that the estimation problem reduces to a linear equation. 
Let 
%\begin{eqnarray}
%\begin{array}{l}
${c}=\mathbb{E}[Y|Z=z]-\mathbb{E}[Y|z=z_0]$, 
and ${\boldsymbol e}=({e}^{1},\ldots,{e}^{K})^T$ where 
${e}^{k}=\mathbb{E}[\vartheta_k(X,{\boldsymbol W})|Z=z]-\mathbb{E}[\vartheta_k(X,{\boldsymbol W})|Z=z_0]$. %and ${e}^{k}=\mathbb{E}[\pi_k(X,{\boldsymbol W})|Z=z]-\mathbb{E}[\pi_k(X,{\boldsymbol W})|Z=z_0]$
%\end{array}
%\end{eqnarray}
%for $j=1,\ldots,J$. %, $k=1,\ldots,K$ and $m=1,\ldots,M$.
%Furthermore, denote %${\boldsymbol f}=({d}^{m,1},\ldots,{d}^{j},{e}^{m,1},\ldots,{e}^{k})^T$
%${\boldsymbol d}=({d}^{1},\ldots,{d}^{J})^T$.
Then, %the integral 
equation (\ref{IE6})  reduces to a linear equation $c={\boldsymbol \gamma}^T{\boldsymbol e}$.

{\bf P-CAPCE estimator.} Given datasets ${\cal D}^{(1)} = \{x^{(1)}_i,{\boldsymbol w}^{(1)}_i,z^{(1)}_i\}_{i=1}^{N_1}$ and ${\cal D}^{(2)} = \{y^{(2)}_i,z^{(2)}_i\}_{i=1}^{N_2}$, our P-CAPCE estimator consists of two stages.

\noindent{\bf Stage 1.} %We learn prediction models $\hat{\mathbb{E}}[Y|Z=z]$ using ${\cal D}^{(2)}$ and $\hat{\mathbb{E}}[\vartheta_k(X,{\boldsymbol W})|Z=z]$ for $k=1,\ldots,K$ using ${\cal D}^{(1)}$.
%We perform the regression using the power series basis functions. 
Let basis functions be ${\boldsymbol q}(z)=(q_1(z),q_2(z),\ldots,q_P(z))^T$. % and consider the model $\hat{\mathbb{E}}[Y|Z=z]=\sum_{p=1}^P \omega_p q_p(z)$, $\hat{\mathbb{E}}[\vartheta_k(X,{\boldsymbol W})|Z=z]=\sum_{p=1}^P \nu_p^k q_p(z)$ for $k=1,\ldots,K$. Denote ${\boldsymbol \omega}=(\omega_1,\ldots,\omega_P)^T$ and  ${\boldsymbol \nu}^k=(\nu_1^k,\ldots,\nu_P^k)^T$.
%We select an IV value $z_0$. 
Denote $\hat{c}_i=\hat{\mathbb{E}}[Y|Z=z_i]-\hat{\mathbb{E}}[Y|Z=z_0]$ and $\hat{e}_i^{k}=\hat{\mathbb{E}}[\vartheta_k(X,{\boldsymbol W})|Z=z_i]-\hat{\mathbb{E}}[\vartheta_k(X,{\boldsymbol W})|Z=z_0]$. 
Let variance-covariance matrices $\hat{\bf M}^{(1)}=\sum_{i=1}^{N_1} N_1^{-1}{\boldsymbol q}(z^{(1)}_i){\boldsymbol q}(z^{(1)}_i)^T$ and  $\hat{\bf M}^{(2)}=\sum_{i=1}^{N_2} N_2^{-1}{\boldsymbol q}(z^{(2)}_i){\boldsymbol q}(z^{(2)}_i)^T$.
%for $m=1,\ldots,M$.
We obtain the following predication values %$\hat{c}_i=\hat{\mathbb{E}}[Y|Z=z_i]-\hat{\mathbb{E}}[Y|Z=z_0]$, $\hat{e}_i^{k}=\hat{\mathbb{E}}[\vartheta_k(X,{\boldsymbol W})|Z=z_i]-\hat{\mathbb{E}}[\vartheta_k(X,{\boldsymbol W})|Z=z_0]$ for $i=1,\ldots,N$ and $k=1,\ldots,K$.
%and $\hat{e}_i^{k}=\hat{\mathbb{E}}[\varsigma_k(X,{\boldsymbol W})|Z=z_i]-\hat{\mathbb{E}}[\varsigma_k(X,{\boldsymbol W})|Z=z_0]$ for $i=1,\ldots,N$, $m=1,\ldots,M$, $j=1,\ldots,J$, and $k=1,\ldots,L$. These values can be calculated as below:
\begin{equation}
\label{eq-pred2}
\left\{
\begin{array}{l}
%\renewcommand{\arraystretch}{1}
    %&&c_i=\hat{\mathbb{E}}[Y|Z=z_i]-\hat{\mathbb{E}}[Y|Z=z_0]=({\boldsymbol q}(z_i)-{\boldsymbol q}(z_0))^T\hat{\bf M}^{-}\sum_{l=1}^N \frac{1}{N} {\boldsymbol q}(z_l)y_l\\
    \hat{c}_i%=\hat{\mathbb{E}}[Y|Z=z_i]-\hat{\mathbb{E}}[Y|Z=z_0]
    =({\boldsymbol q}(z_i)-{\boldsymbol q}(z_0))^T\hat{\bf M}^{(2)-}\sum_{l=1}^{N_2} \frac{1}{N_2} {\boldsymbol q}(z^{(2)}_l)y^{(2)}_l\\
    \hat{e}_i^{k}%=\hat{\mathbb{E}}[\varphi_j(X,{\boldsymbol W})|Z=z_i]-\hat{\mathbb{E}}[\varphi_j(X,{\boldsymbol W})|Z=z_0]
    =({\boldsymbol q}(z_i)-{\boldsymbol q}(z_0))^T\hat{\bf M}^{(1)-}\\
 \hspace{2cm}\times\sum_{l=1}^{N_1} \frac{1}{N_1} {\boldsymbol q}(z^{(1)}_l)\vartheta_k(x^{(1)}_l,{\boldsymbol w}^{(1)}_l)\\
    %\hat{e}_i^{k}%=\hat{\mathbb{E}}[\varsigma_k(X,{\boldsymbol W})|Z=z_i]-\hat{\mathbb{E}}[\varsigma_k(X,{\boldsymbol W})|Z=z_0]
    %=({\boldsymbol q}(z_i)-{\boldsymbol q}(z_0))^T\hat{\bf M}^{(1)-}\sum_{l=1}^{N_1} \frac{1}{N_1} {\boldsymbol q}(z^{(1)}_l)\varsigma_k(x^{(1)}_l,{\boldsymbol w}^{(1)}_l)\nonumber
\end{array}
\right.
\end{equation}
for  $k=1,\ldots,K$. 
%, $m=1,\ldots,M$, $j=1,\ldots,J$, and $k=1,\ldots,L$,where $\hat{\bf M}^{-}$ denotes a generalized inverse, which satisfies $\hat{\bf M}\hat{\bf M}^{-}\hat{\bf M}=\hat{\bf M}$.
Let $N=N_1+N_2$ and  $(z_1,\ldots,z_N)=(z_1^{(1)},\ldots,z_{N_1}^{(1)},z_1^{(2)},\ldots,z_{N_2}^{(2)})$. We will compute predicted values in (\ref{eq-pred2}) for all $i=1,\ldots,N$.

\noindent{\bf Stage 2.} 
%Consider multiple IVs $Z=(Z^1,\ldots,Z^M)$.
Estimate parameters ${\boldsymbol \gamma}$ based on the linear equation $c={\boldsymbol \gamma}^T{\boldsymbol e}$.
Let $\hat{\boldsymbol c}=(\hat{c}_1,\ldots,\hat{c}_{N})^T$, $\hat{\boldsymbol e}_i=(\hat{e}^{1}_i,\ldots,\hat{e}^{K}_i)^T$,
%and $\hat{\boldsymbol e}^k_i=(\hat{e}^{1,j}_i,\ldots,\hat{e}^{j}_i)^T$.
$\hat{\bf E}=(\hat{\boldsymbol e}_1,\ldots,\hat{\boldsymbol e}_N)^T$, and
%$\hat{\bf E}_i=(\hat{\boldsymbol e}^{1}_i,\ldots,\hat{\boldsymbol e}^{K}_i)$, and $\hat{\bf F}_i=(\hat{\bf D}_i,\hat{\bf E}_i)$.\\
%We denote ${\boldsymbol \delta}=({\boldsymbol \beta}^T,{\boldsymbol \gamma}^T)^T$ and 
the empirical risk be
\begin{eqnarray}
\label{Q-P}
    {Q}_4({\boldsymbol \gamma}\ ;{\cal D}^{(1)},{\cal D}^{(2)})=\sum_{i=1}^N \frac{1}{N}(\hat{c}_i-\hat{\boldsymbol e}_i^T{\boldsymbol \gamma})^2.
\end{eqnarray}
We make the following assumption:
\begin{assumption}
\label{COM2}
Given a positive regularization parameter $B_P$, ${\boldsymbol \gamma}$ satisfies ${\boldsymbol \gamma}^T{\boldsymbol \gamma} \leq B_P$.    
\end{assumption}
Under Assumption \ref{COM2}, our estimator $\hat{\boldsymbol \gamma}$ is given by the optimization problem below:
\begin{eqnarray}
\label{OPT-P}
    %\hat{\boldsymbol \gamma}=\arg
    \min_{\boldsymbol \gamma}{Q_4}({\boldsymbol \gamma}\ ;{\cal D}^{(1)},{\cal D}^{(2)})\text{ subject to } {\boldsymbol \gamma}^T{\boldsymbol \gamma}\leq B_P.
\end{eqnarray}
This problem can be solved by the ridge regression method with the following solution {\citep{Hilt1977}}: 
%and ${\bf S}$ be a diagonal matrix whose diagonal elements are $\lambda\text{diag}[{\boldsymbol \Lambda}]$, where ${\boldsymbol 1}_J$ is $J$-th vector $(1,1,\ldots,1)$ and $\text{diag}[{\boldsymbol \Lambda}]$ is a diagonal vector of ${\boldsymbol \Lambda}$. 
\begin{eqnarray}
    \hat{\boldsymbol \gamma}=(\hat{\bf E}^T\hat{\bf E}+\zeta_P{\bf I}_K)^{-1}\hat{\bf E}^T\hat{\boldsymbol c},
\end{eqnarray}
where $\zeta_P$ is a regularization parameter, and ${\bf I}_K$ is a $K \times K$ identity matrix. Then, our proposed P-CAPCE estimator  is given by $\displaystyle \hat{\mathbb{E}}[\partial_x{Y}_{x}|{\boldsymbol w}]=\sum_{k=1}^{K}\hat{\gamma}_{k} \theta_k(x,{\boldsymbol w})$.
%Then, estimator of the parameter $\hat{\boldsymbol \pi}$ is given as $(\hat{\boldsymbol \beta},g(\hat{\boldsymbol \gamma}))$. \jin{What is $g(\gamma)$?}

{\bf Model Selection.} %Finally, we explain the performance metric of the trained models by Algorithm \ref{alg1}. We have the trained parameters $\hat{\boldsymbol \pi}$ and available test observations ${\cal D'}=\{z'^{(i)},x'^{(i)},y'^{(i)}\}_{i=1}^{N'}$.
We presume the models in Stage 1 have been selected appropriately. We can use the empirical risk in equation~(\ref{Q-P}) as a performance metric of the trained model in Stage 2 with parameters $\hat{\boldsymbol \gamma}$ if given separate test datasets %${\cal D'}=\{x'_i,y'_i,z'_i,{\boldsymbol w}'_i\}_{i=1}^{N'}$ or 
${\cal D}^{(1)'}=\{x^{(1)'}_i,z^{(1)'}_i,{\boldsymbol w}^{(1)'}_i\}_{i=1}^{N_1'}$  and ${\cal D}^{(2)'}=\{y^{(2)'}_i,z^{(2)'}_i\}_{i=1}^{N_2'}$. 
Let $N'= N_1' + N_2'$. Assume $\hat{c}'_i$ and $\hat{\boldsymbol e}'_i$ for $i=1,\ldots,N$ are computed using ${\cal D}^{(1)'}$ and ${\cal D}^{(2)'}$.
%By substituting ${\cal D}$ to ${\cal D'}$, we obtain $\hat{\boldsymbol H}'$ and $\hat{\bf D}'$.
Then, we can evaluate the trained model by the  test error %mean squared error (MSE):
%\begin{eqnarray}
%\label{Q2}
$\displaystyle \hat{Q}_4(\hat{\boldsymbol \gamma}\ ;{\cal D}^{(1)'},{\cal D}^{(2)'})=\frac{1}{N'}\sum_{i=1}^{N'} (\hat{c}'_i-\hat{\boldsymbol e}_i^{'T}\hat{\boldsymbol \gamma})^2$. 
%\end{eqnarray}
Given a separate dataset, this performance metric can be used for model selection from various candidate  basis functions or the number $K$ or $P$ of basis terms.

{\bf Property of P-CAPCE estimator.} 
We show that P-CAPCE estimator is consistent. % and the rate of convergence.
\begin{theorem}[Consistency]
\label{PTHEO1}
    Under SCM ${\cal M}_{IV}$ and Assumptions \ref{AS1}, \ref{AS2},  \ref{COM2},
    \ref{PA1}, \ref{PA2}, \ref{PA3}, and \ref{PA6},
    %D.1, D.2, D.3, and D.4,
    letting $P \rightarrow \infty$, then $\|\hat{\boldsymbol \gamma}-{\boldsymbol \gamma}\|\xrightarrow{p} 0$.
\end{theorem}
\begin{theorem}[Rate of Convergence]
\label{PTHEO2}
   Under SCM ${\cal M}_{IV}$ and Assumptions \ref{AS1}, 
 \ref{AS2}, \ref{COM2},
 \ref{PRA1}, \ref{PRA2}, \ref{PRA4}, \ref{PRA5}, and \ref{PRA8},
 %E.1, E.2, E.3, E.4, and E.5,
 setting $N=N_1=N_2$, then $\|\hat{\boldsymbol \gamma}-{\boldsymbol \gamma}\|={o}_p(N^{-1/4})$.
\end{theorem}
%The rate of convergence is given under similar assumptions in \citep{Xiaohong2018}, shown in Appendix. %\jin{These appeared copied from the Sieve estimator. They needs updates.}
{Assumptions D.1 - 4 are shown in Appendix \ref{appD}. Assumptions E.1 - 5 are in Appendix \ref{appE}.}

%We do not need Assumption \ref{B1}, \ref{AS3} and \ref{COM} for consistency since it is always satisfied when $K<\infty$. 
%We can estimate CAPCE under weaker assumptions than PTSLS \citep{Wooldridge2010}.

\subsection{RKHS CAPCE estimator}

Finally, we develop a reproducing kernel Hilbert space (RKHS) CAPCE estimator. 
RKHS models are popular and widely used in nonparametric regression \citep{Theodoridis2006,Scholkopf2013}. 

% $\hat{\mathbb{E}}[\partial_x{Y}_{x}|{\boldsymbol W}={\boldsymbol w}]$. % following  \citep{Singh2019}.
%We consider two datasets ${\cal D}^{(1)} = \{x^{(1)}_i,{\boldsymbol w}^{(1)}_i,z^{(1)}_i\}_{i=1}^{N_1}$ and ${\cal D}^{(2)} = \{y^{(2)}_i,z^{(2)}_i\}_{i=1}^{N_2}$.

{\bf RKHS model.} Let $k_{X,{\boldsymbol W}}: \Omega_{X,{\boldsymbol W}} \times \Omega_{X,{\boldsymbol W}} \rightarrow \mathbb{R}$ and $k_Z: \Omega_Z \times \Omega_Z \rightarrow \mathbb{R}$ be measurable positive definitive kernels corresponding to RKHSs ${\cal H}_{X,{\boldsymbol W}}$ and ${\cal H}_Z$.
%, and denote $\left<\cdot,\cdot\right>_{{\cal H}_{X,{\boldsymbol W}}}$ and $\left<\cdot,\cdot\right>_{{\cal H}_{Z}}$ be their inner products. %\citep{Taylor2004}:
%\begin{definition}[Positive definitive kernels]
    A symmetric function $k: \Omega \times \Omega \rightarrow \mathbb{R}$ is called positive-definite kernel if
    %\begin{eqnarray}
    $\displaystyle \sum_{i=1}^n\sum_{j=1}^n c_ic_j k({\boldsymbol a}_i,{\boldsymbol a}_j)\geq 0$
    %\end{eqnarray}
    for all ${\boldsymbol a}_1,\ldots,{\boldsymbol a}_n \in \Omega$ given any $n \in \mathbb{N}$ and $c_1,\ldots,c_n \in \mathbb{R}$ \citep{Taylor2004}.
%\end{definition}
Denote the feature map $\eta: \Omega_{X,{\boldsymbol W}}  \rightarrow {\cal H}_{X,{\boldsymbol W}}$, $(x,{\boldsymbol w}) \mapsto k'_{X,{\boldsymbol W}}(x,{\boldsymbol w},\cdot,\cdot)$ and $\psi: \Omega_Z  \rightarrow {\cal H}_Z$, $z \mapsto k_Z(z,\cdot).$
{In addition, we denote the antiderivative feature function $\pi: \Omega_{X,{\boldsymbol W}}  \rightarrow {\cal H}_{X,{\boldsymbol W}}, (x,{\boldsymbol w}) \mapsto k_{X,{\boldsymbol W}}(x,{\boldsymbol w},\cdot,\cdot)$ with $\displaystyle \pi(x,{\boldsymbol w})=-\int \eta(x,{\boldsymbol w})dx$ and the antiderivative kernel function $\displaystyle k_{X,{\boldsymbol W}}(x,{\boldsymbol w},x',{\boldsymbol w}')=\int k'_{X,{\boldsymbol W}}(x,{\boldsymbol w},x',{\boldsymbol w}')dxdx'$. 
%Denote the feature map
%\begin{eqnarray}
%$\pi: \Omega_{X,{\boldsymbol W}}  \rightarrow {\cal H}_{X,{\boldsymbol W}}, (x,{\boldsymbol w}) \mapsto k_{X,{\boldsymbol W}}(x,{\boldsymbol w},\cdot,\cdot)$ and $\psi: \Omega_Z  \rightarrow {\cal H}_Z, \boldsymbol z \mapsto k_Z(z,\cdot).$
Assume that the CAPCE takes the form 
%\jin{Do the following notation need to be changed given that you have used a different notation in describing the two stagtes?}
\begin{equation}
 \mathbb{E}[\partial_x Y_{x}|{\boldsymbol w}]= H(\pi(x,{\boldsymbol w}))
\end{equation}
for some operator $H\in {\cal L}_2({\cal H}_{X,{\boldsymbol W}},\Omega_Y)$, where ${\cal L}_2(\Omega_1,\Omega_2)$ is the ${\cal L}_2$ measurable function space from $\Omega_1$ to $\Omega_2$, and $H(\pi(x,{\boldsymbol w}))$ is a composition function $H \circ \pi: \Omega_{X,{\boldsymbol W}}  \rightarrow \Omega_Y$.}
%\yuta{The, the RKHS estimator is given by $\left<\hat{H},\pi(x,{\boldsymbol w}) \right>$ for $x \in \Omega_X$ and ${\boldsymbol w} \in {\boldsymbol W}$.}
%\end{eqnarray}
%\jin{The notation $\pi$ and $\pi$ have been used in the last section. Use different notation.}
%\begin{eqnarray}
%    {\bf K}_{Z^{(1)}Z^{(=1=
%    \left(
%    \begin{array}{ccc}
%     k_Z(z_1^{(1)},z_1^{(1)})    & \cdots & k_Z(z_1^{(1)},z_{N_1}^{(1)}) \\
%        \vdots & \ddots & \vdots\\
%    k_Z(z_{N_1}^{(1)},z_1^{(1)}) & \cdots & k_Z(z_{N_1}^{(1)},z_{N_1}^{(1)})
%    \end{array}
%    \right).
%\end{eqnarray}
Our RKHS CAPCE estimator consists of two stages (a detailed derivation is provided in Appendix \ref{appA2}). 
%\yuta{We introduce $G_1, G_2 \in {\cal H}_{Z}$ and $H \in {\cal H}_{X,{\boldsymbol W}}$ for learning.}\\

\noindent{\bf Stage 1.} We learn an operator $G_1\in {\cal L}_2({\cal H}_{Z},{\cal H}_{X,{\boldsymbol W}})$  
%\jin{why "linear" operator?} \yuta{[Comment: This is because it is define by the inner product $G_1(\psi(z))=<\psi(z),g_1>$ with some $g_1 \in {\cal H}_Z$.  It is linear about the input $\psi(z)$. But I will delete ``linear" since it is confusing.]}
that satisfies $\mathbb{E}[\pi(X,{\boldsymbol W})|Z=z]=G_1 (\psi(z))$, 
%\yuta{[Comment: $G_1 (\psi(z))$ is a composition function $G_1 \circ \psi: \Omega_Z  \rightarrow {\cal H}_{X,{\boldsymbol W}}$ of $\psi: \Omega_Z  \rightarrow {\cal H}_Z$ and $G_1:{\cal H}_{Z} \rightarrow {\cal H}_{X,{\boldsymbol W}}$.[]}
%where $\pi(X,{\boldsymbol W})=-\int_{-\infty}^X \pi(x,{\boldsymbol w})dx$.
%, and we do not specify the form of $\psi(x)$.\jin{Isn't $\pi()$  a function of X and W? "we don't specify $\psi$"???}
and learn an operator $G_2\in {\cal L}_2({\cal H}_{Z},\Omega_Y)$ that satisfies $\mathbb{E}[Y|Z=z]=G_2(\psi(z))$. % in the first stage.

\noindent{\bf Stage 2.} 
%Then, in the second stage,
We learn an operator $H \in {\cal L}_2({\cal H}_{X,{\boldsymbol W}},\Omega_Y)$ that satisfies $\hat{\mathbb{E}}[Y|Z=z]-\hat{\mathbb{E}}[Y|Z=z_0]=H(\hat{\mathbb{E}}[\pi(X,{\boldsymbol W})|Z=z]-\hat{\mathbb{E}}[\pi(X,{\boldsymbol W})|Z=z_0]) \Leftrightarrow \hat{G}_1(\psi(z)-\psi(z_0))=H(\hat{G}_2(\psi(z)-\psi(z_0)))$, where $\hat{G}_1$ and $\hat{G}_2$ are learned in Stage 1.

We learn $\hat{G}_1$, $\hat{G}_2$, and $\hat{H}$ by the following optimization problems using datasets ${\cal D}^{(1)}$ and ${\cal D}^{(2)}$:
%\vspace{-0.6cm}
\begin{equation}
\begin{aligned}
    &\min_{G_1%\in {\cal L}_2({\cal H}_{Z},{\cal H}_{X,{\boldsymbol W}})
    } \frac{1}{N_1}\sum_{i=1}^{N_1}\left\|\pi(x_i^{(1)},{\boldsymbol w}_i^{(1)})-G_1(\psi(z_i^{(1)}))\right\|^2_{{\cal H}_{X,{\boldsymbol W}}}\\
    &\hspace{3.5cm}+\lambda_1\left\|G_1\right\|^2_{{\cal L}_2({\cal H}_Z,{\cal H}_{X,{\boldsymbol W}})},
  \end{aligned}
\end{equation}
\begin{equation}
\begin{aligned}
&\min_{G_2%\in {\cal L}_2({\cal H}_{Z},\Omega_Y)
} \frac{1}{N_2}\sum_{i=1}^{N_2}\left\|y_i^{(2)}-G_2(\psi(z_i^{(2)}))\right\|^2\\
&\hspace{4cm}+\lambda_2\left\|G_2\right\|^2_{{\cal L}_2({\cal H}_{Z},\Omega_Y)},
  \end{aligned}
\end{equation}
\begin{equation}
\begin{aligned}
  &\min_{H% \in {\cal L}_2(\Omega_Y,{\cal H}_{X,{\boldsymbol W}})
  } \frac{1}{N_2}\sum_{i=1}^{N_2}\Big\|\hat{G}_2(\psi(z_i^{(2)})-\psi(z_0))\\
  &\hspace{3cm}-H(\hat{G}_1( \psi(z_i^{(2)})-\psi(z_0)))\Big\|^2\\
  &\hspace{0.6cm}+\xi\left\|H\right\|^2_{{\cal L}_2({\cal H}_{X,{\boldsymbol W}},\Omega_Y)}+\lambda_3\left\|H\circ \hat{G}_1\right\|^2_{{\cal L}_2({\cal H}_Z,\Omega_Y)},
  \end{aligned}
\end{equation}
\noindent where $(\lambda_1,\lambda_2,\lambda_3,\xi)$ are regularization parameters. From the representer theorem \citep{Schlkopf2001}, the optimal $G_1$ exists in $\text{span}\{\psi(z_1^{(1)}),\ldots,\psi(z_{N_1}^{(1)})\}$, and the optimal $G_2$ and $H$ exist in $\text{span}\{\psi(z_1^{(2)}),\ldots,\psi(z_{N_2}^{(2)})\}$.
%\yuta{The, the RKHS estimator is given by $\left<\hat{H},\pi(x,{\boldsymbol w}) \right>$ for $x \in \Omega_X$ and ${\boldsymbol w} \in {\boldsymbol W}$.}

We denote %${\bf K}$ is a 
gram matrices ${\bf K}_{Z^{(1)}Z^{(1)}}=\{ k_Z(z_i^{(1)},z_j^{(1)}) \}_{i,j=1}^{N_1}$; 
${\bf K}_{Z^{(1)}z_0}$ is $N_1 \times N_1$ matrix $\{ k_Z(z_i^{(1)},z_0) \}_{i,j=1}^{N_1}$; and ${\bf K}_{(X,{\boldsymbol W})^{(1)}(x,{\boldsymbol w})}$ is $N_1$-dimension vector $\{ k_{X,{\boldsymbol W}}(x_i^{(1)},{\boldsymbol w}_i^{(1)},x,{\boldsymbol w}) \}_{i=1}^{N_1}$. 
%We give a closed form of RKHS CAPCE estimator as below: 
%\begin{alg}
Then, the RKHS CAPCE estimator is given by 
\begin{align}\label{eq-rkhs}
\hat{\mathbb{E}}[\partial_x{Y}_{x}|{\boldsymbol w}]=\hat{\boldsymbol \alpha}^T{\bf K}_{(X,{\boldsymbol W})^{(1)}(x,{\boldsymbol w})}, 
\end{align} 
where
%\vspace{-0.8cm}
\begin{equation}\label{eq-rkhs-inverse}
\begin{aligned}
&\hat{\boldsymbol \alpha}=(\hat{\bf O}\hat{\bf O}^T+N_2\xi {\bf K}_{(X,{\boldsymbol W})^{(1)}(X,{\boldsymbol W})^{(1)}}+N_2\lambda_3 {\bf I}_{N_2})^{-1}\\
&\hspace{1cm}\times\hat{\bf O}\{{\boldsymbol y}^{(2)T}({\bf K}_{Z^{(2)}Z^{(2)}}+N_2\lambda_2 {\bf I}_{N_2})^{-1}\\
&\hspace{3cm}\times({\bf K}_{Z^{(2)}Z^{(2)}}-{\bf K}_{Z^{(2)}z_0})\},
\end{aligned}
\end{equation}
\begin{equation}
\begin{aligned}
&\hat{\bf O}={\bf K}_{(X,{\boldsymbol W})^{(1)}(X,{\boldsymbol W})^{(1)}}({\bf K}_{Z^{(1)}Z^{(1)}}+N_1\lambda_1 {\bf I}_{N_1})^{-1}\\
&\hspace{2.5cm}\times({\bf K}_{Z^{(1)}Z^{(2)}}-{\bf K}_{Z^{(1)}z_0}),
\end{aligned}
\end{equation}
 and ${\bf I}_N$ is a $N \times N$ identity matrix.
%\end{alg}

{\bf Model Selection.} 
We presume the models in Stage 1 have been selected appropriately, and introduce a model selection method in Stage 2 following  \citep{Singh2019}.  
Assume we have separate datasets ${\cal D}^{(1)'} = \{x^{(1)'}_i,{\boldsymbol w}^{(1)'}_i,z^{(1)'}_i\}_{i=1}^{N'_1}$ and ${\cal D}^{(2)} = \{y^{(2)'}_i,z^{(2)'}_i\}_{i=1}^{N'_2}$.
%Train stage 1 estimator $\lambda_1$ on stage 1 observations $\{x_i^{(1)},{\boldsymbol w}^{(1)}_i,z_i^{(1)}\}$ then select stage 1 regularization parameter value $\lambda_1^*$ to minimize out-of-sample loss, calculated from observations $\{x_i^{(2)},{\boldsymbol w}^{(2)}_i,z_i^{(2)}\}$. 
%Also, train stage 1 estimator $\lambda_2$ on stage 1 observations $\{x_i^{(2)},{\boldsymbol w}^{(2)}_i,z_i^{(2)}\}$ then select stage 1 regularization parameter value $\lambda_2^*$ to minimize out-of-sample loss, calculated from observations$\{x_i^{(1)},{\boldsymbol w}^{(1)}_i,z_i^{(1)}\}$. 
%Train stage 2 estimator $\hat{H}$ on stage 2 observations $\{y_i^{(2)},{\boldsymbol w}^{(2)}_i,z_i^{(2)}\}$ then select $\xi$ stage 2 regularization parameter value $\xi^*$ to minimize out-of-sample loss, calculated from observations $\{y_i^{(1)},{\boldsymbol w}^{(1)}_i,z_i^{(1)}\}$.
%We introduce the evaluation algorithm of RKHS CAPCE estimator below:
%\begin{alg}
%\label{ALG2}
We determine the optimal $\lambda_1^*$ by minimizing
\begin{equation}
\begin{aligned}
&L_1(\lambda_1)=\frac{1}{N'_1}\text{Trace}\Big[{\bf K}_{(X,{\boldsymbol W})^{(1)'}(X,{\boldsymbol W})^{(1)'}}\\
&\hspace{0cm}-2{\bf K}_{(X,{\boldsymbol W})^{(1)'}(X,{\boldsymbol W})^{(1)}}{\bf P}_1+{\bf P}_1^T {\bf K}_{(X,{\boldsymbol W})^{(1)}(X,{\boldsymbol W})^{(1)}}{\bf P}_1 \Big],
\end{aligned}
\end{equation}
where ${\bf P}_1=({\bf K}_{Z^{(1)}Z^{(1)}}+N'_1\lambda_1{\bf I}_{N_1})^{-1}{\bf K}_{Z^{(1)}Z^{(2)}}$. We determine the optimal $\lambda_2^*$ by minimizing
\begin{equation}
\begin{aligned}
&L_2(\lambda_2)=\frac{1}{N'_2}\text{Trace}\Big[{\boldsymbol y}^{(2)'}{\boldsymbol y}^{(2)'T}\\
&\hspace{1.5cm}-2{\boldsymbol y}^{(2)'}{\boldsymbol y}^{(2)T}{\bf P}_2+{\bf P}_2^T {\boldsymbol y}^{(2)}{\boldsymbol y}^{(2)T}{\bf P}_2 \Big],
\end{aligned}
\end{equation}
where %${\boldsymbol y}^{(2)'}=(y_1^{(2)'},\ldots,y_{N'_2}^{(2)'})^T$, ${\boldsymbol y}^{(2)}=(y_1^{(2)},\ldots,y_{N_2}^{(2)})^T$ and
${\bf P}_2=({\bf K}_{Z^{(1)}Z^{(1)}}+N_1\lambda_2{\bf I}_{N_1})^{-1}{\bf K}_{Z^{(1)}Z^{(2)}}$.
Finally, we determine the optimal $\xi^*$ and $\lambda_3^*$ by minimizing test error
%\begin{eqnarray}
$\displaystyle L(\lambda_3,\xi)=\frac{1}{N'_2}\sum_{i=1}^{N'_2}\|{\boldsymbol y}^{(2)'T}({\bf K}_{Z^{(2)'}Z^{(2)'}}+N'_2\lambda_2^* {\bf I}_{N'_2})^{-1}({\bf K}_{Z^{(2)'}Z^{(2)'}}-{\bf K}_{Z^{(2)'}z_0})-\hat{H}_{\lambda_3,\xi}(x_i^{(1)'},{\boldsymbol w}_i^{(1)'})\|^2$
%\end{eqnarray}
where $\hat{H}_{\lambda_3,\xi}$ is learned with  $\lambda_1=\lambda_1^*$ and $\lambda_2=\lambda_2^*$ using ${\cal D}^{(1)}$  and ${\cal D}^{(2)}$.
%\end{alg}

{\bf Properties of RKHS CAPCE estimator.}  The RKHS CAPCE estimator requires ${\cal O}(N_1^3)+{\cal O}(N_2^3)$ time \citep{Saunders1998}.
%and ${\cal O}(N_1^2)+{\cal O}(N_2^2)$ memory. 
%When $\lambda_3$ is 0, the consistency of RKHS CAPCE estimator is guaranteed by similar assumptions in \citep{Singh2019}, shown in Appendix. \jin{Can you write a formal consistency theorem here?}
%Furthermore, the following theorem holds under Assumptions F.1-8 shown in Appendix F.
{We show that RKHS CAPCE is consistent under assumptions similar to Kernel IV \citep{Singh2019}. Assumptions F.1 - 8 are shown in Appendix \ref{appF}.}
\begin{theorem}[Consistency]
\label{RTEO1}
    Under SCM ${\cal M}_{IV}$ and Assumptions \ref{AS1}, \ref{AS2},  
    \ref{RAS1}, \ref{RAS2}, \ref{RAS3}, \ref{RAS4}, \ref{RAS5}, \ref{RAS6}, \ref{RAS7} and \ref{RAS8},
    %F.1, F.2, F.3, F.4, F.5, F.6, F.7, and F.8,
   the  RKHS CAPCE estimator in (\ref{eq-rkhs}) %$\hat{\mathbb{E}}[\partial_x{Y}_{x}|{\boldsymbol W}={\boldsymbol w}]=\hat{\boldsymbol \alpha}^T{\bf K}_{(X,{\boldsymbol W})^{(1)}(x,{\boldsymbol w})}$ 
    converges pointwise to CAPCE %$\mathbb{E}[\partial_xY_{x}|{\boldsymbol W}={\boldsymbol w}]$ 
    when $\lambda_3=0$.
\end{theorem}
When $\lambda_3=0$, the inverse of the matrix $\hat{\bf O}\hat{\bf O}^T+N_2 \xi  {\bf K}_{(X,{\boldsymbol W})^{(1)}(X,{\boldsymbol W})^{(1)}}$ in Eq.~(\ref{eq-rkhs-inverse}) is numerically unstable. In practice, regularization leads to bias, but we must consider the bias-variance trade-off.
%Assumptions C.1 $\sim$ C.8 are shown in Appendix.
%\jin{is a pointwise consistent estimator of CAPCE (almost everywhere?) or converge pointwise to CAPCE (almost everywhere?)}

\begin{table*}[!t]
\renewcommand{\arraystretch}{1.2}
%\small
\centering
\caption{Means of estimated coefficients by PTSLS and P-CAPCE estimators in setting (A).}
%\vspace{-5pt}
\label{tab:TAB1}
\renewcommand{\arraystretch}{1.1}
\begin{tabular}{l|lll|lll}
\hline
Estimated coefficients & \multicolumn{3}{c|}{$N=1000$}& \multicolumn{3}{c}{$N=10000$}\\
 \hline
Terms      & 1 & $W$      & $X$ & 1 & $W$      & $X$ \\
                           \hline \hline
PTSLS                      & 1.248     & 50.032 & 27.862               & 1.101      & 51.181 & 19.763               \\
P-CAPCE & -1.651    & 10.383 & 19.293               & 1.226      & 0.963  & 19.971 \\
\hdashline
True Coefficients          & 1         & 1      & 20                   & 1          & 1      & 20                   \\
\hline
\end{tabular}
\vspace{-0cm}
\end{table*}

\begin{table*}[t]
\renewcommand{\arraystretch}{1.2}
%\small
\centering
\caption{MSE and run time of estimators in settings (A) and (B).}
%\vspace{-5pt}
\label{tab:TAB2}
\begin{tabular}{l|lll:lll}
\hline
   \multicolumn{1}{c|}{MSE}      & PTSLS &NTSLS & Kernel IV & P-CAPCE & S-CAPCE & RKHS CAPCE \\
        \hline \hline
(A) $N=1000$  & 925.139 & 418.396 & 548.821 & 104.990 & 203.079 & {\bf 87.853} \\
Time (second) & 0.126 & 0.361 & 6.105 & 0.132 & 0.596 & 6.410 \\
(A) $N=10000$ & 817.074 & 357.777 & 495.742 & {\bf 69.185}  & 185.056 &    71.276    \\
Time (second) & 0.372 & 1.127 & 2814.018 & 0.452 & 1.883 & 4530.765\\
\hline
(B) $N=1000$  & 290.340 & 46.405  & 45.734  & 202.313 & {\bf 8.600}   & 11.612 \\
Time (second) & 0.127 & 0.356 & 6.019 & 0.143 & 0.454 & 6.540 \\
(B) $N=10000$ & 265.400 & 20.990  & 51.470  & 54.124  & {\bf 3.579}   & 8.985   \\ 
Time (second) & 0.367& 1.031 & 2951.841 & 0.485 & 1.836 & 4360.991\\
\hline
\end{tabular}
\end{table*}

\section{Experiments}

In this section, we present numerical experiments to demonstrate the performance of the proposed P-CAPCE, sieve CAPCE,  and RKHS CAPCE estimators. {Detailed settings are  in Appendix \ref{appG}.} 
The experiments are performed using an Apple M1 (16GB).

{\bf Baselines.} We compare with the most widely used methods PTSLS (parametric), NTSLS (sieve), and Kernel IV. These methods compute $\mathbb{E}[Y_{x}|{w}]$ which we differentiate to compute CAPCE  $\mathbb{E}[\partial_x Y_{x}|{w}]$.
%Additional information shown in Appendix.
%\subsection{Parametric Estimation}
%First, we compare the PTSLS and P-CAPCE estimator when $g_0$ is null in Eq. (\ref{EQ1}).\\

{\bf SCM Settings.} We consider the following two SCMs:  $W:=H+E_1, X:=Z+W+H+E_2$, and 
%\vspace{-0.1cm}
%{%\small
\begin{equation}
    \begin{aligned}
\label{eq-scm}
\left\{
\begin{array}{l}
Y:=10X^2+WX+X+W+50 f(W)H+E_3\ \hfill\text{(A)}\\
Y:=\text{exp}(X)\text{exp}(W)+25 f(W)H+E_3\hfill\text{(B)}
\end{array}
\right.
\end{aligned}
\end{equation}
%\normalsize}
where $f(W)=W^5+W^4+W^3+W^2$. The SCMs satisfy separability Assumption~\ref{AS2} but not (\ref{eq-sep}). 
We use setting (A) as a parametric setting and setting (B) as a nonparametric setting. 
Values of $Z$, $H$, $E_1$, $E_2$, and $E_3$ are sampled i.i.d. from a uniform distribution on $[-1,1]$.
%$U[-1,1]$.
True CAPCE  is $20x+w+1$ in setting (A) and $\text{exp}(x)\text{exp}(w)$ in setting (B).
%\yuta{The sample sizes are $N=1000$ and $N=10000$. We choose the parameters of each method using test errors from candidates shown in Appendix G.}

{\bf Setting of P-CAPCE  and PTSLS.} %We learn the conditional expectations of basis functions
%$\mathbb{E}[Y|Z=z]$, $\mathbb{E}[X|Z=z]$, $\mathbb{E}[WX|Z=z]$ and $\mathbb{E}[X^2|Z=z]$
%by the nonlinear model, 
%\begin{eqnarray}
%    $b_0+ b_1Z+b_2Z^2$.
%\end{eqnarray}
We used the basis terms $\{1,W,X\}$ for P-CAPCE and $\{1,W,X,WX,X^2\}$ for PTSLS, which match setting (A). 
%and let $z_0=-1$.
%We regularize the matrix $\displaystyle \hat{\bf G}^T \hat{\bf G}$ by adding $0.001 {\bf I}$ for PTSLS estimator and $0.1 {\bf I}$ for P-CAPCE estimator, where ${\bf I}$ is an identity matrix of size $M$.
%Regularize value is determined by test MSE from $\{1,10^{-1},10^{-2},10^{-3}\}$.
%The results of the test errors are shown in Table 1.

{\bf Setting of NTSLS and sieve CAPCE.} %We learn the conditional expectations by the nonlinear model, 
%\begin{eqnarray}
%    $b_0+ b_1Z+b_2Z^2+b_3Z^3$,
%\end{eqnarray}
We consider the  basis terms $h_p(X)h_q(W)$ for $p=0,1,2$ and $q=0,1,2$, where $h_p$ is Hermite polynomial functions: $h_0(t)=1$, $h_1(t)=t$, $h_2(t)=t^2-1$, and $h_3(t)=t^3-3t$. 
%and let $z_0=-1$.
%Let $\kappa=2$, and we calculate $\hat{\Lambda}$ by Monte Carlo integration using uniform distribution $(x,w)=(U(-4,4),U(-2,2))$, where $\Omega_X \subseteq [-4,4]$ and $\Omega_X \subseteq [-2,2]$. Regularize value is determined by test MSE from $\{1,10^{-1},10^{-2},10^{-3}\}$.\\
%We regularize the matrix $\displaystyle \hat{\bf G}^T \hat{\bf G}$ by adding $10^{-2} \hat{\Lambda}$.
%In addition, we give an experiment using multivariate linear basis function $\{1,W,X\}$, which is a minimal basis function to build CAPCE.
%Results of the test errors are shown in Table 5.
%We estimate CAPCE via differentiating estimated $\mathbb{E}[Y_{x}|{W}={w}]$.\\
%{\bf Setting of sieve NTSLS estimator.} We learn $\mathbb{E}[Y|Z=z], \mathbb{E}[h_p(X)h_q(W)|Z=z]$ for any $p=0,1,2,3$, $q=0,1,2,3$ and $q=0,1$ by the nonlinear model, 
%\begin{eqnarray}
%    $b_0+ b_1Z+b_2Z^2+b_3Z^3$.
%\end{eqnarray}
%where $h_0(t)=1$, $h_1(t)=t$, $h_2(t)=t^2-1$ and $h_3(t)=t^3-3t$.
%Multivariate linear basis function are $\{1,W,X,WX,X^2\}$.
%In this situation, the function $f_Y^2$ is mis-specified.
%Let $\kappa=2$, and we calculate $\hat{\Lambda}$ by Monte Carlo integration using uniform distridution $(x,w)=(U(-4,4),U(-2,2))$.
%Regularize value is determined by test error from $\{1,10^{-1},10^{-2},10^{-3},\ldots\}$.
%We regularize the matrix $\displaystyle \hat{\bf G}^T \hat{\bf G}$ by adding $10^{-3} \hat{\Lambda}$.
%We estimate CAPCE via differentiating estimated $\mathbb{E}[Y_{x}|{W}={w}]$. \jin{Explain why NTSLS uses different settings than S-CAPCE.}
%Results of the test errors are shown in Table 6.\\

{\bf Setting of Kernel IV and RKHS CAPCE.} We use polynomial kernel function $k_Z(z,z')=(z^Tz'+C_1)^{C_2}$ and $k_{X,W}(x,w,x',w')=((x,w)^T(x',w')+C_3)^{C_4}$. 
%We select the kernel parameters $(C_1,C_2)$ from $\{1,2,3,4,5\} \times \{1,2,3,4,5\}$.
%, and determined $(\zeta_1,\zeta_2)=(4,5)$ by test MSE.
%We select the regularize values $\lambda_1,\lambda_2$ from $\{1,10^{-1},10^{-2},10^{-3}\}$, respectively, and $(\lambda_3,\xi)$ is from $\{100,10,1\} \times \{100,10,1\}$. \\
%Then, we determine $(\lambda_1,\lambda_2,\lambda_3,\xi)=(0.01,0.01,1,100)$.\\
%{\bf Setting of PTSLS estimator.} We learn $\mathbb{E}[Y|Z=z]$, $\mathbb{E}[W|Z=z]$, $\mathbb{E}[X|Z=z]$, $\mathbb{E}[WX|Z=z]$ and $\mathbb{E}[X^2|Z=z]$ by the nonlinear model, 
%\begin{eqnarray}
%    $b_0+ b_1Z+b_2Z^2$.
%\end{eqnarray}
%We consider the following terms, $\{1,W,X,WX,X^2\}$ to build model of $\mathbb{E}[Y_{x}|{W}={w}]$.
%We regularize the matrix $\displaystyle \hat{\bf G}^T \hat{\bf G}$ by adding $0.1 {\bf I}$.
%Regularize value is determined by test error from $\{1,10^{-1},10^{-2},10^{-3},\ldots\}$.
%We estimate CAPCE via differentiating estimated $\mathbb{E}[Y_{x}|{W}={w}]$. \jin{The settings have a lot of overlap, no need to repeat. Write a single Settings paragraph.} 
%Results of the test errors are shown in Table 3.\\
%{\bf Results.}

%\begin{wrapfigure}{r}[1pt]{0.65\textwidth}
\begin{figure}[!t]
\vspace{-1cm}
%\Huge
    \centering
    \begin{minipage}[c]{0.475\textwidth}
    \hspace{-0.5cm}
    \scalebox{1}{
% Created by tikzDevice version 0.12.3.1 on 2023-05-16 11:19:23
% !TEX encoding = UTF-8 Unicode
\begin{tikzpicture}[x=1pt,y=1pt]
\definecolor{fillColor}{RGB}{255,255,255}
\path[use as bounding box,fill=fillColor,fill opacity=0.00] (0,0) rectangle (252.94,216.81);
\begin{scope}
\path[clip] (  0.00,  0.00) rectangle (252.94,216.81);
\definecolor{drawColor}{RGB}{0,0,0}

\path[draw=drawColor,line width= 0.4pt,line join=round,line cap=round] ( 55.81, 61.20) -- (221.13, 61.20);

\path[draw=drawColor,line width= 0.4pt,line join=round,line cap=round] ( 55.81, 61.20) -- ( 55.81, 55.20);

\path[draw=drawColor,line width= 0.4pt,line join=round,line cap=round] ( 97.14, 61.20) -- ( 97.14, 55.20);

\path[draw=drawColor,line width= 0.4pt,line join=round,line cap=round] (138.47, 61.20) -- (138.47, 55.20);

\path[draw=drawColor,line width= 0.4pt,line join=round,line cap=round] (179.80, 61.20) -- (179.80, 55.20);

\path[draw=drawColor,line width= 0.4pt,line join=round,line cap=round] (221.13, 61.20) -- (221.13, 55.20);

\node[text=drawColor,anchor=base,inner sep=0pt, outer sep=0pt, scale=  1.00] at ( 55.81, 39.60) {-1.0};

\node[text=drawColor,anchor=base,inner sep=0pt, outer sep=0pt, scale=  1.00] at ( 97.14, 39.60) {-0.5};

\node[text=drawColor,anchor=base,inner sep=0pt, outer sep=0pt, scale=  1.00] at (138.47, 39.60) {0.0};

\node[text=drawColor,anchor=base,inner sep=0pt, outer sep=0pt, scale=  1.00] at (179.80, 39.60) {0.5};

\node[text=drawColor,anchor=base,inner sep=0pt, outer sep=0pt, scale=  1.00] at (221.13, 39.60) {1.0};

\path[draw=drawColor,line width= 0.4pt,line join=round,line cap=round] ( 49.20, 74.10) -- ( 49.20,163.67);

\path[draw=drawColor,line width= 0.4pt,line join=round,line cap=round] ( 49.20, 74.10) -- ( 43.20, 74.10);

\path[draw=drawColor,line width= 0.4pt,line join=round,line cap=round] ( 49.20, 92.01) -- ( 43.20, 92.01);

\path[draw=drawColor,line width= 0.4pt,line join=round,line cap=round] ( 49.20,109.93) -- ( 43.20,109.93);

\path[draw=drawColor,line width= 0.4pt,line join=round,line cap=round] ( 49.20,127.84) -- ( 43.20,127.84);

\path[draw=drawColor,line width= 0.4pt,line join=round,line cap=round] ( 49.20,145.75) -- ( 43.20,145.75);

\path[draw=drawColor,line width= 0.4pt,line join=round,line cap=round] ( 49.20,163.67) -- ( 43.20,163.67);

\node[text=drawColor,rotate= 90.00,anchor=base,inner sep=0pt, outer sep=0pt, scale=  1.00] at ( 34.80, 74.10) {-20};

\node[text=drawColor,rotate= 90.00,anchor=base,inner sep=0pt, outer sep=0pt, scale=  1.00] at ( 34.80, 92.01) {0};

\node[text=drawColor,rotate= 90.00,anchor=base,inner sep=0pt, outer sep=0pt, scale=  1.00] at ( 34.80,109.93) {20};

\node[text=drawColor,rotate= 90.00,anchor=base,inner sep=0pt, outer sep=0pt, scale=  1.00] at ( 34.80,145.75) {60};

\path[draw=drawColor,line width= 0.4pt,line join=round,line cap=round] ( 49.20, 61.20) --
	(227.75, 61.20) --
	(227.75,167.61) --
	( 49.20,167.61) --
	cycle;
\end{scope}
\begin{scope}
\path[clip] ( 49.20, 61.20) rectangle (227.75,167.61);
\definecolor{drawColor}{RGB}{0,0,0}

\path[draw=drawColor,line width= 0.8pt,line join=round,line cap=round] ( 55.81, 75.89) --
	( 57.48, 76.25) --
	( 59.15, 76.61) --
	( 60.82, 76.98) --
	( 62.49, 77.34) --
	( 64.16, 77.70) --
	( 65.83, 78.06) --
	( 67.50, 78.42) --
	( 69.17, 78.78) --
	( 70.84, 79.15) --
	( 72.51, 79.51) --
	( 74.18, 79.87) --
	( 75.85, 80.23) --
	( 77.52, 80.59) --
	( 79.19, 80.96) --
	( 80.86, 81.32) --
	( 82.53, 81.68) --
	( 84.20, 82.04) --
	( 85.87, 82.40) --
	( 87.54, 82.77) --
	( 89.21, 83.13) --
	( 90.88, 83.49) --
	( 92.55, 83.85) --
	( 94.22, 84.21) --
	( 95.89, 84.58) --
	( 97.56, 84.94) --
	( 99.23, 85.30) --
	(100.90, 85.66) --
	(102.57, 86.02) --
	(104.24, 86.38) --
	(105.91, 86.75) --
	(107.58, 87.11) --
	(109.25, 87.47) --
	(110.92, 87.83) --
	(112.59, 88.19) --
	(114.26, 88.56) --
	(115.93, 88.92) --
	(117.60, 89.28) --
	(119.27, 89.64) --
	(120.94, 90.00) --
	(122.61, 90.37) --
	(124.28, 90.73) --
	(125.95, 91.09) --
	(127.62, 91.45) --
	(129.29, 91.81) --
	(130.96, 92.18) --
	(132.63, 92.54) --
	(134.30, 92.90) --
	(135.97, 93.26) --
	(137.64, 93.62) --
	(139.31, 93.98) --
	(140.98, 94.35) --
	(142.65, 94.71) --
	(144.32, 95.07) --
	(145.99, 95.43) --
	(147.66, 95.79) --
	(149.33, 96.16) --
	(151.00, 96.52) --
	(152.67, 96.88) --
	(154.34, 97.24) --
	(156.01, 97.60) --
	(157.68, 97.97) --
	(159.35, 98.33) --
	(161.02, 98.69) --
	(162.69, 99.05) --
	(164.36, 99.41) --
	(166.03, 99.78) --
	(167.70,100.14) --
	(169.37,100.50) --
	(171.04,100.86) --
	(172.71,101.22) --
	(174.38,101.58) --
	(176.05,101.95) --
	(177.71,102.31) --
	(179.38,102.67) --
	(181.05,103.03) --
	(182.72,103.39) --
	(184.39,103.76) --
	(186.06,104.12) --
	(187.73,104.48) --
	(189.40,104.84) --
	(191.07,105.20) --
	(192.74,105.57) --
	(194.41,105.93) --
	(196.08,106.29) --
	(197.75,106.65) --
	(199.42,107.01) --
	(201.09,107.38) --
	(202.76,107.74) --
	(204.43,108.10) --
	(206.10,108.46) --
	(207.77,108.82) --
	(209.44,109.18) --
	(211.11,109.55) --
	(212.78,109.91) --
	(214.45,110.27) --
	(216.12,110.63) --
	(217.79,110.99) --
	(219.46,111.36) --
	(221.13,111.72);
\definecolor{drawColor}{RGB}{100,149,237}

\path[draw=drawColor,line width= 0.8pt,dash pattern=on 1pt off 3pt on 4pt off 3pt ,line join=round,line cap=round] ( 55.81, 52.57) --
	( 57.48, 53.37) --
	( 59.15, 54.17) --
	( 60.82, 54.98) --
	( 62.49, 55.78) --
	( 64.16, 56.59) --
	( 65.83, 57.39) --
	( 67.50, 58.19) --
	( 69.17, 59.00) --
	( 70.84, 59.80) --
	( 72.51, 60.61) --
	( 74.18, 61.41) --
	( 75.85, 62.22) --
	( 77.52, 63.03) --
	( 79.19, 63.89) --
	( 80.86, 64.81) --
	( 82.53, 65.43) --
	( 84.20, 66.25) --
	( 85.87, 67.13) --
	( 87.54, 68.01) --
	( 89.21, 68.89) --
	( 90.88, 69.76) --
	( 92.55, 70.64) --
	( 94.22, 71.52) --
	( 95.89, 72.40) --
	( 97.56, 73.22) --
	( 99.23, 73.89) --
	(100.90, 74.70) --
	(102.57, 75.36) --
	(104.24, 76.02) --
	(105.91, 76.68) --
	(107.58, 77.41) --
	(109.25, 78.21) --
	(110.92, 79.00) --
	(112.59, 79.62) --
	(114.26, 80.58) --
	(115.93, 81.19) --
	(117.60, 81.54) --
	(119.27, 82.22) --
	(120.94, 83.15) --
	(122.61, 83.72) --
	(124.28, 84.05) --
	(125.95, 84.52) --
	(127.62, 84.79) --
	(129.29, 85.01) --
	(130.96, 85.29) --
	(132.63, 85.59) --
	(134.30, 85.94) --
	(135.97, 86.25) --
	(137.64, 86.56) --
	(139.31, 86.89) --
	(140.98, 86.77) --
	(142.65, 87.17) --
	(144.32, 87.68) --
	(145.99, 88.15) --
	(147.66, 88.19) --
	(149.33, 87.87) --
	(151.00, 88.17) --
	(152.67, 88.47) --
	(154.34, 88.77) --
	(156.01, 89.10) --
	(157.68, 89.05) --
	(159.35, 89.00) --
	(161.02, 88.95) --
	(162.69, 88.90) --
	(164.36, 89.05) --
	(166.03, 88.99) --
	(167.70, 88.85) --
	(169.37, 88.72) --
	(171.04, 88.58) --
	(172.71, 88.44) --
	(174.38, 88.36) --
	(176.05, 88.17) --
	(177.71, 88.02) --
	(179.38, 87.86) --
	(181.05, 87.71) --
	(182.72, 87.56) --
	(184.39, 87.40) --
	(186.06, 87.19) --
	(187.73, 86.80) --
	(189.40, 86.41) --
	(191.07, 86.02) --
	(192.74, 85.63) --
	(194.41, 85.24) --
	(196.08, 84.85) --
	(197.75, 84.47) --
	(199.42, 84.10) --
	(201.09, 83.73) --
	(202.76, 83.36) --
	(204.43, 82.98) --
	(206.10, 82.61) --
	(207.77, 82.24) --
	(209.44, 81.98) --
	(211.11, 81.89) --
	(212.78, 81.80) --
	(214.45, 81.72) --
	(216.12, 81.63) --
	(217.79, 81.55) --
	(219.46, 81.46) --
	(221.13, 81.37);
\definecolor{drawColor}{RGB}{0,0,255}

\path[draw=drawColor,line width= 0.8pt,dash pattern=on 4pt off 4pt ,line join=round,line cap=round] ( 55.81, 79.01) --
	( 57.48, 79.33) --
	( 59.15, 79.64) --
	( 60.82, 79.95) --
	( 62.49, 80.27) --
	( 64.16, 80.58) --
	( 65.83, 80.90) --
	( 67.50, 81.21) --
	( 69.17, 81.52) --
	( 70.84, 81.84) --
	( 72.51, 82.15) --
	( 74.18, 82.47) --
	( 75.85, 82.78) --
	( 77.52, 83.09) --
	( 79.19, 83.41) --
	( 80.86, 83.72) --
	( 82.53, 84.04) --
	( 84.20, 84.35) --
	( 85.87, 84.66) --
	( 87.54, 84.98) --
	( 89.21, 85.29) --
	( 90.88, 85.61) --
	( 92.55, 85.92) --
	( 94.22, 86.24) --
	( 95.89, 86.55) --
	( 97.56, 86.86) --
	( 99.23, 87.18) --
	(100.90, 87.49) --
	(102.57, 87.81) --
	(104.24, 88.12) --
	(105.91, 88.43) --
	(107.58, 88.75) --
	(109.25, 89.06) --
	(110.92, 89.38) --
	(112.59, 89.69) --
	(114.26, 90.00) --
	(115.93, 90.32) --
	(117.60, 90.63) --
	(119.27, 90.95) --
	(120.94, 91.26) --
	(122.61, 91.57) --
	(124.28, 91.89) --
	(125.95, 92.20) --
	(127.62, 92.52) --
	(129.29, 92.83) --
	(130.96, 93.14) --
	(132.63, 93.46) --
	(134.30, 93.77) --
	(135.97, 94.09) --
	(137.64, 94.40) --
	(139.31, 94.72) --
	(140.98, 95.03) --
	(142.65, 95.34) --
	(144.32, 95.66) --
	(145.99, 95.97) --
	(147.66, 96.29) --
	(149.33, 96.60) --
	(151.00, 96.91) --
	(152.67, 97.23) --
	(154.34, 97.54) --
	(156.01, 97.86) --
	(157.68, 98.17) --
	(159.35, 98.48) --
	(161.02, 98.80) --
	(162.69, 99.11) --
	(164.36, 99.43) --
	(166.03, 99.74) --
	(167.70,100.05) --
	(169.37,100.37) --
	(171.04,100.68) --
	(172.71,101.00) --
	(174.38,101.31) --
	(176.05,101.62) --
	(177.71,101.94) --
	(179.38,102.25) --
	(181.05,102.57) --
	(182.72,102.88) --
	(184.39,103.19) --
	(186.06,103.51) --
	(187.73,103.82) --
	(189.40,104.14) --
	(191.07,104.45) --
	(192.74,104.77) --
	(194.41,105.08) --
	(196.08,105.39) --
	(197.75,105.71) --
	(199.42,106.02) --
	(201.09,106.34) --
	(202.76,106.65) --
	(204.43,106.96) --
	(206.10,107.28) --
	(207.77,107.59) --
	(209.44,107.91) --
	(211.11,108.22) --
	(212.78,108.53) --
	(214.45,108.85) --
	(216.12,109.16) --
	(217.79,109.48) --
	(219.46,109.79) --
	(221.13,110.10);
\definecolor{drawColor}{RGB}{100,149,237}

\path[draw=drawColor,line width= 0.8pt,dash pattern=on 1pt off 3pt on 4pt off 3pt ,line join=round,line cap=round] ( 55.81,105.80) --
	( 57.48,105.66) --
	( 59.15,105.52) --
	( 60.82,105.39) --
	( 62.49,105.26) --
	( 64.16,105.16) --
	( 65.83,105.06) --
	( 67.50,104.97) --
	( 69.17,104.87) --
	( 70.84,104.78) --
	( 72.51,104.68) --
	( 74.18,104.59) --
	( 75.85,104.49) --
	( 77.52,104.39) --
	( 79.19,104.30) --
	( 80.86,104.11) --
	( 82.53,103.88) --
	( 84.20,103.53) --
	( 85.87,103.34) --
	( 87.54,103.20) --
	( 89.21,102.97) --
	( 90.88,102.75) --
	( 92.55,102.53) --
	( 94.22,102.31) --
	( 95.89,102.08) --
	( 97.56,101.86) --
	( 99.23,101.64) --
	(100.90,101.41) --
	(102.57,101.19) --
	(104.24,101.04) --
	(105.91,100.95) --
	(107.58,100.86) --
	(109.25,100.77) --
	(110.92,100.68) --
	(112.59,100.60) --
	(114.26,100.51) --
	(115.93,100.42) --
	(117.60,100.81) --
	(119.27,100.74) --
	(120.94,100.64) --
	(122.61,100.52) --
	(124.28,100.37) --
	(125.95,100.28) --
	(127.62,100.58) --
	(129.29,100.48) --
	(130.96,100.75) --
	(132.63,101.09) --
	(134.30,101.37) --
	(135.97,101.76) --
	(137.64,102.59) --
	(139.31,102.95) --
	(140.98,102.79) --
	(142.65,103.13) --
	(144.32,103.47) --
	(145.99,103.84) --
	(147.66,104.41) --
	(149.33,104.99) --
	(151.00,105.59) --
	(152.67,106.19) --
	(154.34,106.79) --
	(156.01,107.39) --
	(157.68,108.05) --
	(159.35,108.75) --
	(161.02,109.44) --
	(162.69,110.31) --
	(164.36,111.23) --
	(166.03,112.18) --
	(167.70,113.17) --
	(169.37,114.17) --
	(171.04,114.92) --
	(172.71,115.81) --
	(174.38,116.74) --
	(176.05,117.67) --
	(177.71,118.61) --
	(179.38,119.54) --
	(181.05,120.46) --
	(182.72,121.38) --
	(184.39,122.29) --
	(186.06,123.09) --
	(187.73,123.65) --
	(189.40,124.20) --
	(191.07,124.75) --
	(192.74,125.30) --
	(194.41,125.85) --
	(196.08,126.40) --
	(197.75,126.96) --
	(199.42,127.57) --
	(201.09,128.50) --
	(202.76,129.43) --
	(204.43,130.36) --
	(206.10,131.29) --
	(207.77,132.22) --
	(209.44,133.15) --
	(211.11,133.92) --
	(212.78,134.64) --
	(214.45,135.36) --
	(216.12,136.08) --
	(217.79,136.86) --
	(219.46,137.74) --
	(221.13,138.63);
\definecolor{drawColor}{RGB}{255,192,203}

\path[draw=drawColor,line width= 0.8pt,dash pattern=on 2pt off 2pt on 6pt off 2pt ,line join=round,line cap=round] ( 55.81,105.75) --
	( 57.48,106.10) --
	( 59.15,106.45) --
	( 60.82,106.79) --
	( 62.49,107.14) --
	( 64.16,107.49) --
	( 65.83,107.84) --
	( 67.50,108.19) --
	( 69.17,108.54) --
	( 70.84,108.89) --
	( 72.51,109.24) --
	( 74.18,109.59) --
	( 75.85,109.94) --
	( 77.52,110.28) --
	( 79.19,110.63) --
	( 80.86,110.98) --
	( 82.53,111.33) --
	( 84.20,111.68) --
	( 85.87,112.03) --
	( 87.54,112.38) --
	( 89.21,112.73) --
	( 90.88,113.08) --
	( 92.55,113.42) --
	( 94.22,113.77) --
	( 95.89,114.12) --
	( 97.56,114.47) --
	( 99.23,114.82) --
	(100.90,115.18) --
	(102.57,115.53) --
	(104.24,115.89) --
	(105.91,116.24) --
	(107.58,116.59) --
	(109.25,116.95) --
	(110.92,117.30) --
	(112.59,117.65) --
	(114.26,118.01) --
	(115.93,118.36) --
	(117.60,118.72) --
	(119.27,119.07) --
	(120.94,119.42) --
	(122.61,119.78) --
	(124.28,120.13) --
	(125.95,120.46) --
	(127.62,120.77) --
	(129.29,121.09) --
	(130.96,121.40) --
	(132.63,121.71) --
	(134.30,122.03) --
	(135.97,122.34) --
	(137.64,122.65) --
	(139.31,122.97) --
	(140.98,123.28) --
	(142.65,123.59) --
	(144.32,123.91) --
	(145.99,124.22) --
	(147.66,124.53) --
	(149.33,124.91) --
	(151.00,125.36) --
	(152.67,125.81) --
	(154.34,126.26) --
	(156.01,126.72) --
	(157.68,127.17) --
	(159.35,127.51) --
	(161.02,127.86) --
	(162.69,128.21) --
	(164.36,128.55) --
	(166.03,128.90) --
	(167.70,129.25) --
	(169.37,129.59) --
	(171.04,129.94) --
	(172.71,130.28) --
	(174.38,130.62) --
	(176.05,130.86) --
	(177.71,131.11) --
	(179.38,131.35) --
	(181.05,131.60) --
	(182.72,131.84) --
	(184.39,132.09) --
	(186.06,132.34) --
	(187.73,132.60) --
	(189.40,132.85) --
	(191.07,133.16) --
	(192.74,133.64) --
	(194.41,134.12) --
	(196.08,134.62) --
	(197.75,135.22) --
	(199.42,135.86) --
	(201.09,136.26) --
	(202.76,136.61) --
	(204.43,137.03) --
	(206.10,137.50) --
	(207.77,137.98) --
	(209.44,138.46) --
	(211.11,138.80) --
	(212.78,139.03) --
	(214.45,139.26) --
	(216.12,139.50) --
	(217.79,139.75) --
	(219.46,140.09) --
	(221.13,140.43);
\definecolor{drawColor}{RGB}{255,0,0}

\path[draw=drawColor,line width= 0.8pt,dash pattern=on 1pt off 3pt ,line join=round,line cap=round] ( 55.81,129.51) --
	( 57.48,129.68) --
	( 59.15,129.86) --
	( 60.82,130.04) --
	( 62.49,130.21) --
	( 64.16,130.39) --
	( 65.83,130.56) --
	( 67.50,130.74) --
	( 69.17,130.92) --
	( 70.84,131.09) --
	( 72.51,131.27) --
	( 74.18,131.45) --
	( 75.85,131.62) --
	( 77.52,131.80) --
	( 79.19,131.97) --
	( 80.86,132.15) --
	( 82.53,132.33) --
	( 84.20,132.50) --
	( 85.87,132.68) --
	( 87.54,132.86) --
	( 89.21,133.03) --
	( 90.88,133.21) --
	( 92.55,133.38) --
	( 94.22,133.56) --
	( 95.89,133.74) --
	( 97.56,133.91) --
	( 99.23,134.09) --
	(100.90,134.27) --
	(102.57,134.44) --
	(104.24,134.62) --
	(105.91,134.79) --
	(107.58,134.97) --
	(109.25,135.15) --
	(110.92,135.32) --
	(112.59,135.50) --
	(114.26,135.67) --
	(115.93,135.85) --
	(117.60,136.03) --
	(119.27,136.20) --
	(120.94,136.38) --
	(122.61,136.56) --
	(124.28,136.73) --
	(125.95,136.91) --
	(127.62,137.08) --
	(129.29,137.26) --
	(130.96,137.44) --
	(132.63,137.61) --
	(134.30,137.79) --
	(135.97,137.97) --
	(137.64,138.14) --
	(139.31,138.32) --
	(140.98,138.49) --
	(142.65,138.67) --
	(144.32,138.85) --
	(145.99,139.02) --
	(147.66,139.20) --
	(149.33,139.37) --
	(151.00,139.55) --
	(152.67,139.73) --
	(154.34,139.90) --
	(156.01,140.08) --
	(157.68,140.26) --
	(159.35,140.43) --
	(161.02,140.61) --
	(162.69,140.78) --
	(164.36,140.96) --
	(166.03,141.14) --
	(167.70,141.31) --
	(169.37,141.49) --
	(171.04,141.67) --
	(172.71,141.84) --
	(174.38,142.02) --
	(176.05,142.19) --
	(177.71,142.37) --
	(179.38,142.55) --
	(181.05,142.72) --
	(182.72,142.90) --
	(184.39,143.08) --
	(186.06,143.25) --
	(187.73,143.43) --
	(189.40,143.60) --
	(191.07,143.78) --
	(192.74,143.96) --
	(194.41,144.13) --
	(196.08,144.31) --
	(197.75,144.48) --
	(199.42,144.66) --
	(201.09,144.84) --
	(202.76,145.01) --
	(204.43,145.19) --
	(206.10,145.37) --
	(207.77,145.54) --
	(209.44,145.72) --
	(211.11,145.89) --
	(212.78,146.07) --
	(214.45,146.25) --
	(216.12,146.42) --
	(217.79,146.60) --
	(219.46,146.78) --
	(221.13,146.95);
\definecolor{drawColor}{RGB}{255,192,203}

\path[draw=drawColor,line width= 0.8pt,dash pattern=on 2pt off 2pt on 6pt off 2pt ,line join=round,line cap=round] ( 55.81,152.51) --
	( 57.48,152.51) --
	( 59.15,152.50) --
	( 60.82,152.50) --
	( 62.49,152.49) --
	( 64.16,152.49) --
	( 65.83,152.48) --
	( 67.50,152.48) --
	( 69.17,152.47) --
	( 70.84,152.47) --
	( 72.51,152.46) --
	( 74.18,152.46) --
	( 75.85,152.45) --
	( 77.52,152.45) --
	( 79.19,152.45) --
	( 80.86,152.44) --
	( 82.53,152.44) --
	( 84.20,152.43) --
	( 85.87,152.43) --
	( 87.54,152.43) --
	( 89.21,152.45) --
	( 90.88,152.48) --
	( 92.55,152.50) --
	( 94.22,152.52) --
	( 95.89,152.54) --
	( 97.56,152.57) --
	( 99.23,152.59) --
	(100.90,152.57) --
	(102.57,152.49) --
	(104.24,152.42) --
	(105.91,152.39) --
	(107.58,152.39) --
	(109.25,152.39) --
	(110.92,152.39) --
	(112.59,152.39) --
	(114.26,152.39) --
	(115.93,152.39) --
	(117.60,152.39) --
	(119.27,152.39) --
	(120.94,152.39) --
	(122.61,152.39) --
	(124.28,152.40) --
	(125.95,152.40) --
	(127.62,152.40) --
	(129.29,152.40) --
	(130.96,152.40) --
	(132.63,152.40) --
	(134.30,152.40) --
	(135.97,152.40) --
	(137.64,152.39) --
	(139.31,152.39) --
	(140.98,152.39) --
	(142.65,152.38) --
	(144.32,152.38) --
	(145.99,152.38) --
	(147.66,152.37) --
	(149.33,152.37) --
	(151.00,152.37) --
	(152.67,152.39) --
	(154.34,152.41) --
	(156.01,152.43) --
	(157.68,152.41) --
	(159.35,152.35) --
	(161.02,152.36) --
	(162.69,152.36) --
	(164.36,152.36) --
	(166.03,152.36) --
	(167.70,152.36) --
	(169.37,152.35) --
	(171.04,152.31) --
	(172.71,152.25) --
	(174.38,152.33) --
	(176.05,152.34) --
	(177.71,152.35) --
	(179.38,152.35) --
	(181.05,152.35) --
	(182.72,152.36) --
	(184.39,152.36) --
	(186.06,152.36) --
	(187.73,152.37) --
	(189.40,152.37) --
	(191.07,152.37) --
	(192.74,152.37) --
	(194.41,152.44) --
	(196.08,152.52) --
	(197.75,152.60) --
	(199.42,152.68) --
	(201.09,152.75) --
	(202.76,152.82) --
	(204.43,152.91) --
	(206.10,152.99) --
	(207.77,153.07) --
	(209.44,153.16) --
	(211.11,153.14) --
	(212.78,153.19) --
	(214.45,153.36) --
	(216.12,153.39) --
	(217.79,153.42) --
	(219.46,153.45) --
	(221.13,153.48);
\definecolor{drawColor}{RGB}{0,0,0}
\definecolor{fillColor}{RGB}{255,255,255}

\path[draw=drawColor,line width= 0.4pt,line join=round,line cap=round,fill=fillColor] ( 49.20,167.61) rectangle (112.39,100.41);

\path[draw=drawColor,line width= 0.4pt,line join=round,line cap=round] ( 55.50,159.21) -- ( 68.10,159.21);
\definecolor{drawColor}{RGB}{100,149,237}

\path[draw=drawColor,line width= 0.4pt,dash pattern=on 1pt off 3pt on 4pt off 3pt ,line join=round,line cap=round] ( 55.50,150.81) -- ( 68.10,150.81);
\definecolor{drawColor}{RGB}{0,0,255}

\path[draw=drawColor,line width= 0.4pt,dash pattern=on 4pt off 4pt ,line join=round,line cap=round] ( 55.50,142.41) -- ( 68.10,142.41);
\definecolor{drawColor}{RGB}{100,149,237}

\path[draw=drawColor,line width= 0.4pt,dash pattern=on 1pt off 3pt on 4pt off 3pt ,line join=round,line cap=round] ( 55.50,134.01) -- ( 68.10,134.01);
\definecolor{drawColor}{RGB}{255,192,203}

\path[draw=drawColor,line width= 0.4pt,dash pattern=on 2pt off 2pt on 6pt off 2pt ,line join=round,line cap=round] ( 55.50,125.61) -- ( 68.10,125.61);
\definecolor{drawColor}{RGB}{255,0,0}

\path[draw=drawColor,line width= 0.4pt,dash pattern=on 1pt off 3pt ,line join=round,line cap=round] ( 55.50,117.21) -- ( 68.10,117.21);
\definecolor{drawColor}{RGB}{255,192,203}

\path[draw=drawColor,line width= 0.4pt,dash pattern=on 2pt off 2pt on 6pt off 2pt ,line join=round,line cap=round] ( 55.50,108.81) -- ( 68.10,108.81);
\definecolor{drawColor}{RGB}{0,0,0}

\node[text=drawColor,anchor=base west,inner sep=0pt, outer sep=0pt, scale=  0.70] at ( 74.40,156.80) {True Curve};

\node[text=drawColor,anchor=base west,inner sep=0pt, outer sep=0pt, scale=  0.70] at ( 74.40,140.00) {P-CAPCE};

\node[text=drawColor,anchor=base west,inner sep=0pt, outer sep=0pt, scale=  0.70] at ( 74.40,114.80) {PTSLS};
\end{scope}
\end{tikzpicture}

}
\vspace{-1.75cm}
\captionsetup{labelformat=empty}
\caption*{\footnotesize \hspace{1cm}(a) Parametric setting  (A) (Means, $95 \%$ CI)}
\end{minipage}
    \centering
    \vspace{-1.25cm}
    
    \hspace{0.1cm}
\begin{minipage}[c]{0.475\textwidth}
    \hspace{-0.5cm}
    \scalebox{1}{
% Created by tikzDevice version 0.12.3.1 on 2023-05-15 16:24:13
% !TEX encoding = UTF-8 Unicode
\begin{tikzpicture}[x=1pt,y=1pt]
\definecolor{fillColor}{RGB}{255,255,255}
\path[use as bounding box,fill=fillColor,fill opacity=0.00] (0,0) rectangle (252.94,216.81);
\begin{scope}
\path[clip] (  0.00,  0.00) rectangle (252.94,216.81);
\definecolor{drawColor}{RGB}{0,0,0}

\path[draw=drawColor,line width= 0.4pt,line join=round,line cap=round] ( 55.81, 61.20) -- (221.13, 61.20);

\path[draw=drawColor,line width= 0.4pt,line join=round,line cap=round] ( 55.81, 61.20) -- ( 55.81, 55.20);

\path[draw=drawColor,line width= 0.4pt,line join=round,line cap=round] ( 97.14, 61.20) -- ( 97.14, 55.20);

\path[draw=drawColor,line width= 0.4pt,line join=round,line cap=round] (138.47, 61.20) -- (138.47, 55.20);

\path[draw=drawColor,line width= 0.4pt,line join=round,line cap=round] (179.80, 61.20) -- (179.80, 55.20);

\path[draw=drawColor,line width= 0.4pt,line join=round,line cap=round] (221.13, 61.20) -- (221.13, 55.20);

\node[text=drawColor,anchor=base,inner sep=0pt, outer sep=0pt, scale=  1.00] at ( 55.81, 39.60) {-1.0};

\node[text=drawColor,anchor=base,inner sep=0pt, outer sep=0pt, scale=  1.00] at ( 97.14, 39.60) {-0.5};

\node[text=drawColor,anchor=base,inner sep=0pt, outer sep=0pt, scale=  1.00] at (138.47, 39.60) {0.0};

\node[text=drawColor,anchor=base,inner sep=0pt, outer sep=0pt, scale=  1.00] at (179.80, 39.60) {0.5};

\node[text=drawColor,anchor=base,inner sep=0pt, outer sep=0pt, scale=  1.00] at (221.13, 39.60) {1.0};

\path[draw=drawColor,line width= 0.4pt,line join=round,line cap=round] ( 49.20, 65.14) -- ( 49.20,163.67);

\path[draw=drawColor,line width= 0.4pt,line join=round,line cap=round] ( 49.20, 65.14) -- ( 43.20, 65.14);

\path[draw=drawColor,line width= 0.4pt,line join=round,line cap=round] ( 49.20, 81.56) -- ( 43.20, 81.56);

\path[draw=drawColor,line width= 0.4pt,line join=round,line cap=round] ( 49.20, 97.98) -- ( 43.20, 97.98);

\path[draw=drawColor,line width= 0.4pt,line join=round,line cap=round] ( 49.20,114.41) -- ( 43.20,114.41);

\path[draw=drawColor,line width= 0.4pt,line join=round,line cap=round] ( 49.20,130.83) -- ( 43.20,130.83);

\path[draw=drawColor,line width= 0.4pt,line join=round,line cap=round] ( 49.20,147.25) -- ( 43.20,147.25);

\path[draw=drawColor,line width= 0.4pt,line join=round,line cap=round] ( 49.20,163.67) -- ( 43.20,163.67);

\node[text=drawColor,rotate= 90.00,anchor=base,inner sep=0pt, outer sep=0pt, scale=  1.00] at ( 34.80, 65.14) {-20};

\node[text=drawColor,rotate= 90.00,anchor=base,inner sep=0pt, outer sep=0pt, scale=  1.00] at ( 34.80, 97.98) {0};

\node[text=drawColor,rotate= 90.00,anchor=base,inner sep=0pt, outer sep=0pt, scale=  1.00] at ( 34.80,114.41) {10};

\node[text=drawColor,rotate= 90.00,anchor=base,inner sep=0pt, outer sep=0pt, scale=  1.00] at ( 34.80,147.25) {30};

\path[draw=drawColor,line width= 0.4pt,line join=round,line cap=round] ( 49.20, 61.20) --
	(227.75, 61.20) --
	(227.75,167.61) --
	( 49.20,167.61) --
	cycle;
\end{scope}
\begin{scope}
\path[clip] ( 49.20, 61.20) rectangle (227.75,167.61);
\definecolor{drawColor}{RGB}{0,0,0}

\path[draw=drawColor,line width= 0.8pt,line join=round,line cap=round] ( 55.81,102.45) --
	( 57.48,102.54) --
	( 59.15,102.63) --
	( 60.82,102.73) --
	( 62.49,102.82) --
	( 64.16,102.92) --
	( 65.83,103.02) --
	( 67.50,103.13) --
	( 69.17,103.23) --
	( 70.84,103.34) --
	( 72.51,103.45) --
	( 74.18,103.56) --
	( 75.85,103.67) --
	( 77.52,103.79) --
	( 79.19,103.91) --
	( 80.86,104.03) --
	( 82.53,104.15) --
	( 84.20,104.28) --
	( 85.87,104.41) --
	( 87.54,104.54) --
	( 89.21,104.67) --
	( 90.88,104.81) --
	( 92.55,104.95) --
	( 94.22,105.09) --
	( 95.89,105.23) --
	( 97.56,105.38) --
	( 99.23,105.53) --
	(100.90,105.69) --
	(102.57,105.84) --
	(104.24,106.00) --
	(105.91,106.17) --
	(107.58,106.33) --
	(109.25,106.50) --
	(110.92,106.68) --
	(112.59,106.86) --
	(114.26,107.04) --
	(115.93,107.22) --
	(117.60,107.41) --
	(119.27,107.60) --
	(120.94,107.80) --
	(122.61,108.00) --
	(124.28,108.20) --
	(125.95,108.41) --
	(127.62,108.62) --
	(129.29,108.84) --
	(130.96,109.06) --
	(132.63,109.29) --
	(134.30,109.52) --
	(135.97,109.76) --
	(137.64,110.00) --
	(139.31,110.24) --
	(140.98,110.49) --
	(142.65,110.75) --
	(144.32,111.01) --
	(145.99,111.27) --
	(147.66,111.54) --
	(149.33,111.82) --
	(151.00,112.10) --
	(152.67,112.39) --
	(154.34,112.68) --
	(156.01,112.98) --
	(157.68,113.29) --
	(159.35,113.60) --
	(161.02,113.92) --
	(162.69,114.25) --
	(164.36,114.58) --
	(166.03,114.92) --
	(167.70,115.26) --
	(169.37,115.62) --
	(171.04,115.98) --
	(172.71,116.34) --
	(174.38,116.72) --
	(176.05,117.10) --
	(177.71,117.49) --
	(179.38,117.89) --
	(181.05,118.29) --
	(182.72,118.71) --
	(184.39,119.13) --
	(186.06,119.56) --
	(187.73,120.00) --
	(189.40,120.45) --
	(191.07,120.91) --
	(192.74,121.38) --
	(194.41,121.86) --
	(196.08,122.34) --
	(197.75,122.84) --
	(199.42,123.35) --
	(201.09,123.87) --
	(202.76,124.39) --
	(204.43,124.93) --
	(206.10,125.48) --
	(207.77,126.04) --
	(209.44,126.62) --
	(211.11,127.20) --
	(212.78,127.80) --
	(214.45,128.41) --
	(216.12,129.03) --
	(217.79,129.66) --
	(219.46,130.31) --
	(221.13,130.97);
\definecolor{drawColor}{RGB}{255,0,0}

\path[draw=drawColor,line width= 0.4pt,dash pattern=on 4pt off 4pt ,line join=round,line cap=round] ( 55.81,105.77) --
	( 57.48,105.59) --
	( 59.15,105.41) --
	( 60.82,105.23) --
	( 62.49,105.07) --
	( 64.16,104.92) --
	( 65.83,104.77) --
	( 67.50,104.64) --
	( 69.17,104.51) --
	( 70.84,104.39) --
	( 72.51,104.27) --
	( 74.18,104.17) --
	( 75.85,104.07) --
	( 77.52,103.98) --
	( 79.19,103.90) --
	( 80.86,103.83) --
	( 82.53,103.77) --
	( 84.20,103.71) --
	( 85.87,103.66) --
	( 87.54,103.62) --
	( 89.21,103.59) --
	( 90.88,103.57) --
	( 92.55,103.55) --
	( 94.22,103.54) --
	( 95.89,103.54) --
	( 97.56,103.54) --
	( 99.23,103.56) --
	(100.90,103.58) --
	(102.57,103.61) --
	(104.24,103.65) --
	(105.91,103.70) --
	(107.58,103.75) --
	(109.25,103.81) --
	(110.92,103.88) --
	(112.59,103.96) --
	(114.26,104.05) --
	(115.93,104.15) --
	(117.60,104.25) --
	(119.27,104.36) --
	(120.94,104.48) --
	(122.61,104.61) --
	(124.28,104.75) --
	(125.95,104.90) --
	(127.62,105.05) --
	(129.29,105.22) --
	(130.96,105.39) --
	(132.63,105.58) --
	(134.30,105.77) --
	(135.97,105.97) --
	(137.64,106.19) --
	(139.31,106.41) --
	(140.98,106.64) --
	(142.65,106.88) --
	(144.32,107.14) --
	(145.99,107.40) --
	(147.66,107.67) --
	(149.33,107.96) --
	(151.00,108.26) --
	(152.67,108.56) --
	(154.34,108.88) --
	(156.01,109.21) --
	(157.68,109.55) --
	(159.35,109.90) --
	(161.02,110.27) --
	(162.69,110.65) --
	(164.36,111.04) --
	(166.03,111.44) --
	(167.70,111.85) --
	(169.37,112.28) --
	(171.04,112.73) --
	(172.71,113.18) --
	(174.38,113.65) --
	(176.05,114.14) --
	(177.71,114.63) --
	(179.38,115.15) --
	(181.05,115.67) --
	(182.72,116.22) --
	(184.39,116.78) --
	(186.06,117.35) --
	(187.73,117.94) --
	(189.40,118.55) --
	(191.07,119.17) --
	(192.74,119.81) --
	(194.41,120.47) --
	(196.08,121.14) --
	(197.75,121.83) --
	(199.42,122.54) --
	(201.09,123.27) --
	(202.76,124.02) --
	(204.43,124.79) --
	(206.10,125.58) --
	(207.77,126.38) --
	(209.44,127.21) --
	(211.11,128.05) --
	(212.78,128.92) --
	(214.45,129.81) --
	(216.12,130.72) --
	(217.79,131.65) --
	(219.46,132.61) --
	(221.13,133.59);
\definecolor{drawColor}{RGB}{0,0,255}

\path[draw=drawColor,line width= 0.4pt,dash pattern=on 1pt off 3pt ,line join=round,line cap=round] ( 55.81, 81.82) --
	( 57.48, 82.50) --
	( 59.15, 83.17) --
	( 60.82, 83.84) --
	( 62.49, 84.49) --
	( 64.16, 85.13) --
	( 65.83, 85.77) --
	( 67.50, 86.39) --
	( 69.17, 87.01) --
	( 70.84, 87.62) --
	( 72.51, 88.23) --
	( 74.18, 88.82) --
	( 75.85, 89.42) --
	( 77.52, 90.00) --
	( 79.19, 90.59) --
	( 80.86, 91.17) --
	( 82.53, 91.75) --
	( 84.20, 92.32) --
	( 85.87, 92.90) --
	( 87.54, 93.47) --
	( 89.21, 94.04) --
	( 90.88, 94.62) --
	( 92.55, 95.19) --
	( 94.22, 95.77) --
	( 95.89, 96.34) --
	( 97.56, 96.92) --
	( 99.23, 97.51) --
	(100.90, 98.09) --
	(102.57, 98.69) --
	(104.24, 99.28) --
	(105.91, 99.89) --
	(107.58,100.50) --
	(109.25,101.11) --
	(110.92,101.74) --
	(112.59,102.37) --
	(114.26,103.01) --
	(115.93,103.66) --
	(117.60,104.32) --
	(119.27,104.99) --
	(120.94,105.67) --
	(122.61,106.36) --
	(124.28,107.06) --
	(125.95,107.78) --
	(127.62,108.51) --
	(129.29,109.26) --
	(130.96,110.02) --
	(132.63,110.79) --
	(134.30,111.59) --
	(135.97,112.39) --
	(137.64,113.22) --
	(139.31,114.06) --
	(140.98,114.92) --
	(142.65,115.80) --
	(144.32,116.70) --
	(145.99,117.62) --
	(147.66,118.55) --
	(149.33,119.51) --
	(151.00,120.50) --
	(152.67,121.50) --
	(154.34,122.53) --
	(156.01,123.58) --
	(157.68,124.65) --
	(159.35,125.75) --
	(161.02,126.87) --
	(162.69,128.02) --
	(164.36,129.20) --
	(166.03,130.40) --
	(167.70,131.63) --
	(169.37,132.88) --
	(171.04,134.17) --
	(172.71,135.49) --
	(174.38,136.83) --
	(176.05,138.20) --
	(177.71,139.61) --
	(179.38,141.05) --
	(181.05,142.52) --
	(182.72,144.02) --
	(184.39,145.55) --
	(186.06,147.12) --
	(187.73,148.72) --
	(189.40,150.36) --
	(191.07,152.03) --
	(192.74,153.74) --
	(194.41,155.48) --
	(196.08,157.26) --
	(197.75,159.08) --
	(199.42,160.94) --
	(201.09,162.83) --
	(202.76,164.77) --
	(204.43,166.74) --
	(206.10,168.76) --
	(207.77,170.81) --
	(209.44,172.91) --
	(211.11,175.05) --
	(212.78,177.23) --
	(214.45,179.45) --
	(216.12,181.72) --
	(217.79,184.03) --
	(219.46,186.38) --
	(221.13,188.78);
\definecolor{drawColor}{RGB}{0,100,0}

\path[draw=drawColor,line width= 0.4pt,dash pattern=on 1pt off 3pt on 4pt off 3pt ,line join=round,line cap=round] ( 55.81,110.46) --
	( 57.48,110.53) --
	( 59.15,110.60) --
	( 60.82,110.67) --
	( 62.49,110.74) --
	( 64.16,110.81) --
	( 65.83,110.87) --
	( 67.50,110.94) --
	( 69.17,111.01) --
	( 70.84,111.08) --
	( 72.51,111.16) --
	( 74.18,111.23) --
	( 75.85,111.30) --
	( 77.52,111.37) --
	( 79.19,111.45) --
	( 80.86,111.52) --
	( 82.53,111.60) --
	( 84.20,111.67) --
	( 85.87,111.75) --
	( 87.54,111.83) --
	( 89.21,111.91) --
	( 90.88,111.99) --
	( 92.55,112.08) --
	( 94.22,112.16) --
	( 95.89,112.25) --
	( 97.56,112.33) --
	( 99.23,112.42) --
	(100.90,112.51) --
	(102.57,112.61) --
	(104.24,112.70) --
	(105.91,112.80) --
	(107.58,112.90) --
	(109.25,113.00) --
	(110.92,113.10) --
	(112.59,113.20) --
	(114.26,113.31) --
	(115.93,113.42) --
	(117.60,113.53) --
	(119.27,113.64) --
	(120.94,113.76) --
	(122.61,113.88) --
	(124.28,114.00) --
	(125.95,114.12) --
	(127.62,114.25) --
	(129.29,114.38) --
	(130.96,114.51) --
	(132.63,114.65) --
	(134.30,114.79) --
	(135.97,114.93) --
	(137.64,115.07) --
	(139.31,115.22) --
	(140.98,115.37) --
	(142.65,115.52) --
	(144.32,115.68) --
	(145.99,115.84) --
	(147.66,116.01) --
	(149.33,116.18) --
	(151.00,116.35) --
	(152.67,116.52) --
	(154.34,116.70) --
	(156.01,116.89) --
	(157.68,117.07) --
	(159.35,117.27) --
	(161.02,117.46) --
	(162.69,117.66) --
	(164.36,117.86) --
	(166.03,118.07) --
	(167.70,118.28) --
	(169.37,118.50) --
	(171.04,118.72) --
	(172.71,118.95) --
	(174.38,119.18) --
	(176.05,119.41) --
	(177.71,119.65) --
	(179.38,119.89) --
	(181.05,120.14) --
	(182.72,120.40) --
	(184.39,120.66) --
	(186.06,120.92) --
	(187.73,121.19) --
	(189.40,121.46) --
	(191.07,121.74) --
	(192.74,122.03) --
	(194.41,122.32) --
	(196.08,122.61) --
	(197.75,122.91) --
	(199.42,123.22) --
	(201.09,123.53) --
	(202.76,123.85) --
	(204.43,124.17) --
	(206.10,124.50) --
	(207.77,124.84) --
	(209.44,125.18) --
	(211.11,125.53) --
	(212.78,125.88) --
	(214.45,126.24) --
	(216.12,126.60) --
	(217.79,126.98) --
	(219.46,127.35) --
	(221.13,127.74);
\definecolor{drawColor}{RGB}{255,165,0}

\path[draw=drawColor,line width= 0.4pt,dash pattern=on 2pt off 2pt on 6pt off 2pt ,line join=round,line cap=round] ( 55.81, 70.10) --
	( 57.48, 70.25) --
	( 59.15, 70.41) --
	( 60.82, 70.57) --
	( 62.49, 70.74) --
	( 64.16, 70.91) --
	( 65.83, 71.09) --
	( 67.50, 71.27) --
	( 69.17, 71.46) --
	( 70.84, 71.66) --
	( 72.51, 71.86) --
	( 74.18, 72.06) --
	( 75.85, 72.28) --
	( 77.52, 72.50) --
	( 79.19, 72.72) --
	( 80.86, 72.95) --
	( 82.53, 73.19) --
	( 84.20, 73.43) --
	( 85.87, 73.67) --
	( 87.54, 73.93) --
	( 89.21, 74.18) --
	( 90.88, 74.45) --
	( 92.55, 74.72) --
	( 94.22, 74.99) --
	( 95.89, 75.27) --
	( 97.56, 75.56) --
	( 99.23, 75.85) --
	(100.90, 76.15) --
	(102.57, 76.45) --
	(104.24, 76.76) --
	(105.91, 77.08) --
	(107.58, 77.40) --
	(109.25, 77.73) --
	(110.92, 78.06) --
	(112.59, 78.40) --
	(114.26, 78.74) --
	(115.93, 79.09) --
	(117.60, 79.44) --
	(119.27, 79.80) --
	(120.94, 80.17) --
	(122.61, 80.54) --
	(124.28, 80.92) --
	(125.95, 81.30) --
	(127.62, 81.69) --
	(129.29, 82.09) --
	(130.96, 82.49) --
	(132.63, 82.89) --
	(134.30, 83.31) --
	(135.97, 83.72) --
	(137.64, 84.15) --
	(139.31, 84.58) --
	(140.98, 85.01) --
	(142.65, 85.45) --
	(144.32, 85.90) --
	(145.99, 86.35) --
	(147.66, 86.81) --
	(149.33, 87.27) --
	(151.00, 87.74) --
	(152.67, 88.21) --
	(154.34, 88.69) --
	(156.01, 89.18) --
	(157.68, 89.67) --
	(159.35, 90.17) --
	(161.02, 90.67) --
	(162.69, 91.18) --
	(164.36, 91.69) --
	(166.03, 92.21) --
	(167.70, 92.74) --
	(169.37, 93.27) --
	(171.04, 93.81) --
	(172.71, 94.35) --
	(174.38, 94.90) --
	(176.05, 95.45) --
	(177.71, 96.01) --
	(179.38, 96.58) --
	(181.05, 97.15) --
	(182.72, 97.73) --
	(184.39, 98.31) --
	(186.06, 98.90) --
	(187.73, 99.49) --
	(189.40,100.09) --
	(191.07,100.70) --
	(192.74,101.31) --
	(194.41,101.93) --
	(196.08,102.55) --
	(197.75,103.18) --
	(199.42,103.81) --
	(201.09,104.45) --
	(202.76,105.10) --
	(204.43,105.75) --
	(206.10,106.40) --
	(207.77,107.06) --
	(209.44,107.73) --
	(211.11,108.41) --
	(212.78,109.09) --
	(214.45,109.77) --
	(216.12,110.46) --
	(217.79,111.16) --
	(219.46,111.86) --
	(221.13,112.57);
\definecolor{drawColor}{RGB}{0,0,0}
\definecolor{fillColor}{RGB}{255,255,255}

\path[draw=drawColor,line width= 0.4pt,line join=round,line cap=round,fill=fillColor] ( 49.20,167.61) rectangle (113.74,124.41);

\path[draw=drawColor,line width= 0.4pt,line join=round,line cap=round] ( 54.60,160.41) -- ( 65.40,160.41);
\definecolor{drawColor}{RGB}{255,0,0}

\path[draw=drawColor,line width= 0.4pt,dash pattern=on 4pt off 4pt ,line join=round,line cap=round] ( 54.60,153.21) -- ( 65.40,153.21);
\definecolor{drawColor}{RGB}{0,0,255}

\path[draw=drawColor,line width= 0.4pt,dash pattern=on 1pt off 3pt ,line join=round,line cap=round] ( 54.60,146.01) -- ( 65.40,146.01);
\definecolor{drawColor}{RGB}{0,100,0}

\path[draw=drawColor,line width= 0.4pt,dash pattern=on 1pt off 3pt on 4pt off 3pt ,line join=round,line cap=round] ( 54.60,138.81) -- ( 65.40,138.81);
\definecolor{drawColor}{RGB}{255,165,0}

\path[draw=drawColor,line width= 0.4pt,dash pattern=on 2pt off 2pt on 6pt off 2pt ,line join=round,line cap=round] ( 54.60,131.61) -- ( 65.40,131.61);
\definecolor{drawColor}{RGB}{0,0,0}

\node[text=drawColor,anchor=base west,inner sep=0pt, outer sep=0pt, scale=  0.60] at ( 70.80,158.34) {True Curve};

\node[text=drawColor,anchor=base west,inner sep=0pt, outer sep=0pt, scale=  0.60] at ( 70.80,151.14) {RKHS CAPCE};

\node[text=drawColor,anchor=base west,inner sep=0pt, outer sep=0pt, scale=  0.60] at ( 70.80,143.94) {Kernel IV};

\node[text=drawColor,anchor=base west,inner sep=0pt, outer sep=0pt, scale=  0.60] at ( 70.80,136.74) {S-CAPCE};

\node[text=drawColor,anchor=base west,inner sep=0pt, outer sep=0pt, scale=  0.60] at ( 70.80,129.54) {NTSLS};
\end{scope}
\end{tikzpicture}

}
\vspace{-1.75cm}
\captionsetup{labelformat=empty}
\caption*{\footnotesize \hspace{1cm}(b) Nonparametric setting (B) (Means)}
\end{minipage}
\vspace{0cm}
    \caption{Plots of CAPCE curves with $W=1$.
    X-axis represents treatment ($X$); Y-axis is CAPCE value. Dot-dashed curves in (a) represent $95\%$ pointwise confidence interval (CI).
    %$\mathbb{E}[\partial_x Y_{x}|W=w]$.
    }
    \label{fig:FIG02}
\end{figure}
%\end{wrapfigure}

%\textbf{Results: Parametric setting (A).}
\textbf{Results.} The means of estimated coefficients by %$100$ time simulations of 
PTSLS and P-CAPCE in the parametric setting (A)  are shown in Table \ref{tab:TAB1}. % \ref{tab:PNUM_EXMP1} and \ref{tab:PNUM_EXMP2}.
%We see that the estimated coefficients of P-CAPCE are converging to the true values when the sample size $N=10000$, while one of the PTSLS estimates is still biased.
%The means of coefficients of P-CAPCE estimators are closer to true coefficients than PTSLS; on the other hand, PTSLS is biased largely due to the violation of the separability when $N=10000$.
%Both P-CAPCE and PTSLS estimates have large standard deviations (SD) when $N=1000$ (shown in Appendix \ref{appG}). 
{We observe that, when $N=1000$, both P-CAPCE and PTSLS estimates have large standard deviations (SD) (shown in Appendix \ref{appG}) such that the differences in estimated values are not statistically significant. The estimated coefficients of P-CAPCE are converging to the true values when the sample size $N=10000$, while the coefficient for $W$ estimated by  PTSLS  is still biased.} 
We plotted the true and estimated CAPCE curves given $W=1$ in Figure \ref{fig:FIG02}(a). It is clear that the estimated curve by P-CAPCE is much closer to the true curve than PTSLS. The true and estimated CAPCE surfaces over $(X, W)$ are shown in Appendix \ref{appG}.

%\textbf{Results: Nonparametric setting (B).}
We computed the mean-squared-error (MSE) between estimated and true CAPCE values for each estimator, where MSE is computed as $\displaystyle \frac{1}{N_1'}\sum_{i=1}^{N_1'}\{\hat{g}(x_i^{(1)'},w_i^{(1)'})-g(x_i^{(1)'},w_i^{(1)'})\}^2$ with test dataset ${\cal D}^{(1)'}$, and the results are shown in Table \ref{tab:TAB2}.
%The mean of test MSE of each estimator by $100$ time simulations are shown in Table \ref{tab:TAB2}.
%We note that TSLS (PTSLS and NTSLS) and CAPCE estimators (parametric, sieve and RKHS) have different risk functions.
We observed that our sieve and RKHS CAPCE estimators are superior to the existing methods; sieve and RKHS CAPCE estimators are superior to P-CAPCE  in the nonparametric setting (B); and kernel-based methods are much slower than other methods.  
We plotted the true and estimated CAPCE curves given $W=1$ in Figure \ref{fig:FIG02}(b), which shows the estimated curves by sieve and RKHS CAPCE are much closer to the true curve than NTSLS and Kernel IV. 
The true and estimated CAPCE surfaces over $(X, W)$ are shown in Appendix \ref{appG}.
%\yuta{We have performed additional experiments in settings (C) and (D) under the separability assumption (\ref{eq-sep}) in Appendix G.}

%, $f_Y(X,{\boldsymbol W},{\boldsymbol H},{\boldsymbol u}_Y)=f_Y^1(X,{\boldsymbol W},{\boldsymbol u}_Y)+f_Y^2({\boldsymbol H},{\boldsymbol u}_Y)$.
Overall, the results of settings (A) and (B) show that our proposed methods (P-CAPCE, sieve CAPCE,  RKHS CAPCE) are superior to the previous works (PTSLS, NTSLS, Kernel IV). % in terms of unbiasedness.
%In the setting (A), all CAPCE estimators are almost the same MSE; on the other hand, sieve and RKHS CAPCE estimators are superior to P-CAPCE estimator in the setting (B).
{The advantage of our proposed methods stems from that the underlying models (A) and (B) do not satisfy the separability assumption (\ref{eq-sep}) needed by the existing works. Indeed, we have performed experiments in settings where  the interaction between the covariates $W$ and unobserved confounders $H$ (the $f(W)H$ term in (\ref{eq-scm})) is absent, and the results (presented in Appendix \ref{appG}) show that the performances of the existing methods PTSLS, NTSLS, Kernel IV are comparable with our proposed methods under this situation.} 
%\yuta{The run time of P-CAPCE, S-CAPCE, and RKHS CAPCE is slightly larger than PTSLS, NTSLS, and Kernel IV, respectively.}
 %Sieve CAPCE may require many  basis functions if ${\boldsymbol W}$ is high dimensional. 
%\yuta{In addition, from the results of setting (C) and (D) shown in Appendix G, the MSE of all estimators are almost the same under separability assumption (\ref{eq-sep}) and parametric setting (setting (C)). The MSE of NTSLS, Kernel IV, S-CAPCE and RKHS CAPCE are almost the same under separability assumption (\ref{eq-sep}) and nonparametric setting (setting (D)). On the other hand, the MSE of PTSLS and P-CAPCE in setting (D) are worse than the others due to model misspecification.}
%Finally, 
Among the three proposed methods, the performance of P-CAPCE relies on correct parametric model assumption, and  RKHS CAPCE is  computationally expensive and requires  tuning many regularization parameters.

\section{Application in a Real-World Dataset}

In this section, we present an application of our CAPCE estimators to a real-world dataset in economics. 

%Additional information, Tables 10$\sim$15, are shown in Appendix.\\
{\bf Real-world Dataset.} 
We take up an open dataset ``the National Longitudinal Survey of Young Men'' in the R package ``wooldridge" (\url{https://cran.r-project.org/package=wooldridge}), which has been analyzed by many works, e.g., in \citep{Griliches1997,Blackburn1992}.
%The data source is the National Longitudinal Survey of Young Men. 
The sample size is 935 with 857 left after excluding missing values. %we exclude samples where one of the values is NA. 
We evaluate the heterogeneity of the effect of years of education on monthly wages, which is of great interest in economics \citep{Angrist1991,Card1999}. %, especially the heterogeneity of the effect. 
%Since researchers cannot force people to attend or drop out of school, they use the mother's years of education as an instrumental variable. 
%The mother’s education influences the subject's wage through their years of education.
We followed  \citet{Blackburn1992} to use mother's education as an instrument to uncover the
effect of education on wages. The use of mother's education as an instrument in this dataset has been subjected to debate in the literature (e.g., \citep{Card1999,Jeffrey2001,Wooldridge2010}). 
We take the subject's years of education as the treatment variable ($X$), their monthly wage as the outcome  ($Y$), their mother's years of education as the IV ($Z$), and their IQ as a covariate ($W$).
The domains of $X$ and $Z$ are $[9,18]$, ranging from the 1st year of high school to the 2nd year of a master's degree.
The domain of $W$ is $[50,145]$.
%We exclude samples where one of the four variables is NA. 
%\yuta{We exclude samples where one of the values is NA, and 857 samples are left.}

{\bf Settings.}
We applied  P-CAPCE and PTSLS. Other estimators are not used due to the small sample size.  %We learn the expected values of basis functions by the nonlinear model,
%\begin{eqnarray}
%    $\beta_0+\beta_1Z+\beta_2Z^2$.
%\end{eqnarray}
We use  terms $\{1,W,W^2,X,XW,XW^2\}$ for P-CAPCE and $\{1,W,W^2,X,XW,XW^2,X^2,X^2W,X^2W^2\}$ for PTSLS. {Detailed settings are in Appendix \ref{appG}.}
%We regularize the matrix $\displaystyle \hat{\bf G}^T \hat{\bf G}$ by adding $10^{-3}{\bf I}$.
%Regularize value is determined by test MSE from $\{1,10^{-1},10^{-2},10^{-3},\ldots\}$.\\
%Results of the test errors are shown in Table 7.\\
\begin{comment}
{\bf Setting of PTSLS estimator.} We learn $\mathbb{E}[Y|Z=z], \mathbb{E}[X^pW^q|Z=z]$ for $p=0,1,2$ and $q=0,1,2$ by the nonlinear model,
%\begin{eqnarray}
    $\beta_0+\beta_1Z+\beta_2Z^2$.
%\end{eqnarray}
We consider the following terms, $\{1,W,W^2,X,XW,XW^2,X^2,X^2W,X^2W^2\}$ to build model of $\mathbb{E}[Y_{x}|{W}={w}]$.
%We regularize the matrix $\displaystyle \hat{\bf G}^T \hat{\bf G}$ by adding $100{\bf I}$.
Regularize value is determined by test error from $\{1000,100,10,1,10^{-1}\}$.
We estimate CAPCE via differentiating estimated $\mathbb{E}[Y_{x}|{W}={w}]$.\\
%Results of the test errors are shown in Table 10.\\
\end{comment}

{\bf Results.} 
%We show the basic bootstrapping statistical properties of the CAPCE and PTSLS estimators are shown in Tables 8 and 11.%\ref{tab:NUM_APMP1} and \ref{tab:NUM_APMP2}.
%The prediction values of CAPCE are shown in Tables 9 and 12.
%The results of P-CAPCE and PTSLS estimators are similar; however, P-CAPCE estimator is lower than PTSLS estimators.
%From the results of PTSLS and P-CAPCE estimator, 
The estimated CAPCE values are shown in Appendix \ref{appG}.  For subjects with IQ 100, the estimated CAPCE $\mathbb{E}[\partial_x Y_x|W=100]$ of years of education ($X$) on  wages ($Y$) is given by $94.905-5.618 x$  by P-CAPCE and $108.491-5.882 x$ by PTSLS. Both predict that years of education increase wages, which is consistent with previous works \citep{Blackburn1992,Wooldridge2010}. 
The results also show that education significantly affects wages at the compulsory school level, but the effect  gets weaker with more years of education, consistent with the results in \citep{Angrist1991,Caplan2018}. %\citep{Angrist1991,Spence1973,Caplan2018}. 
On the other hand, for subjects with IQ 80, the estimated CAPCE $\mathbb{E}[\partial_x Y_x|W=80]$ is $60.740-3.598 x$  by P-CAPCE and $69.465-3.057 x$ by PTSLS. For subjects with IQ 120, the estimated CAPCE $\mathbb{E}[\partial_x Y_x|W=120]$ is $136.662-8.086 x$ by P-CAPCE and $156.181-9.531 x$ by PTSLS. 

While we estimate the heterogeneity of causal effects of education on wages across subjects with different IQs, existing works \citep{Blackburn1992,Card1999,Jeffrey2001,Wooldridge2010,Kawakami2023} using this dataset have %used PTSLS and 
focused on the effects of education on wages over the whole population. \citet{Card1999} and \citet{Wooldridge2010} provided a summary of the early works on IV estimates and showed that the estimates of all studies were positive implying education increases wages. On the other hand, our results give two new insights into the effects of education on wages. First, our results suggest that for each sub-population $\text{IQ}=80, 100, 120$, education significantly affects wages at the compulsory school level; but has little effect at the college level. This result is consistent with the result of APCE estimates for the whole population given in \citep{Kawakami2023}. Second, we reveal that the effect of education on wages is more significant for high IQ students, especially at the compulsory school level. To the best of our knowledge, this result has not been revealed in previous studies of this dataset, but it is consistent with the panel data analysis result in \citep{Altonji1996}.

\section{Conclusion}

We study conditional average partial causal effect (CAPCE)  to represent the heterogeneous causal effects of a continuous treatment. We present a method for identifying CAPCE in the IV model. 
Notably, CAPCE $\mathbb{E}[\partial_x Y_{x}|{\boldsymbol w}]$ is identifiable under a weaker assumption than required by $\mathbb{E}[Y_x|{\boldsymbol w}]$, showing the  merit of studying CAPCE instead of $\mathbb{E}[Y_x|{\boldsymbol w}]$, which has been the focus of existing work. We develop three families of  CAPCE estimators: sieve, parametric, and RKHS, and
analyze their statistical properties. 
We empirically demonstrate that the proposed CAPCE estimators are superior to the existing widely used IV methods PTSLS \citep{Angrist2009,Wooldridge2010}, sieve NTSLS \citep{Whitney2003}, and Kernel IV \citep{Singh2019} in settings where the standard separability assumption (\ref{eq-sep}) is violated. 
%We note that CAPCE is sufficient to compare the outcomes of the two treatments.
%In the case of the researcher's interest is focused on only CAPCE, our estimation methods are superior to PTSLS \citep{Wright1928,Angrist2009,Wooldridge2010}, sieve NTSLS \citep{Whitney2003}, and Kernel IV \citep{Singh2019}, which is the most widely used IV methods.
The work provides scientists with a new tool for analyzing the heterogeneous causal effects of a continuous treatment. 
{The results can be extended to an IV model with an additional edge ${\boldsymbol W} \rightarrow Z$. An identification theorem similar to Theorem~\ref{TEO2} can be derived, which uses $\mathbb{P}(Y|Z,{\boldsymbol W})$ as input instead of $\mathbb{P}(Y|Z)$. We present this result in Appendix \ref{appA2}.}

\section*{Acknowledgements}
The authors  thank the anonymous reviewers for their time and thoughtful comments. 
Yuta Kawakami was supported by JSPS KAKENHI Grant Number 22J21928. 
Manabu Kuroki was supported by JSPS KAKENHI Grant Number 21H03504 and 24K14851.
Jin Tian was partially supported by NSF grant CNS-2321786.

\bibliography{citation}

%\maketitle

\newpage

\onecolumn

\title{Appendix to ``Identification and Estimation of Conditional Average Partial Causal Effects via Instrumental Variable”}
\maketitle
\appendix 

%\tableofcontents
%\clearpage
%\section{Appendix}

\section{Proofs of Theorem 3.1 and RKHS CAPCE estimator}
%In this section, we give proofs of Theorems in the body of our paper.
\label{appA}
\begin{comment}
{\bf Proof of Proposition 1.}% \ref{TEO2}} 
We give proof of Proposition 1.

{\it
%\begin{theorem}
{\bf Proposition 1.}
    Under Assumption \ref{AS2}, conditional APCE is equal to the potential APCE, $\mathbb{E}[\partial_x Y_{x}|{\boldsymbol W}={\boldsymbol w}]=\mathbb{E}[\partial_x Y_{x}]$.
%\end{theorem}
}

\begin{proof}
Since $f_Y(X,{\boldsymbol W},{\boldsymbol H},{\boldsymbol u}_Y)=f_Y^1(X,{\boldsymbol W},{\boldsymbol u}_Y)+f_Y^2({\boldsymbol W},{\boldsymbol H},{\boldsymbol u}_Y)$, $\partial_x Y_{x} \indep {\boldsymbol W}$ holds. Then,
\begin{eqnarray}
    \mathbb{E}[\partial_x Y_{x}]=\mathbb{E}[\partial_x Y_{x}|{\boldsymbol W}={\boldsymbol w}]
\end{eqnarray}
holds, and 
\begin{eqnarray}
    \mathbb{E}[\partial_x Y_{x}|{\boldsymbol W}={\boldsymbol w}]=\mathbb{E}[\partial_x Y_{x}|{\boldsymbol W}={\boldsymbol w}]
\end{eqnarray}
holds from the counterfactual consistency.
\end{proof}

\end{comment}

\subsection{Proof of Theorem \ref{TEO2}}% \ref{TEO2}} 
We give proof of Theorem \ref{TEO2}.

%\begin{theorem}
{\it
{\bf Theorem \ref{TEO2}.} (Identification of CAPCE).
%\label{TEO2}
Under SCM ${\cal M}_{IV}$ and Assumptions \ref{AS1} and \ref{AS2}, CAPCE $\mathbb{E}[\partial_x Y_{x}|{\boldsymbol w}]$ is identifiable from distributions $\mathbb{P}(X, {\boldsymbol W}|Z)$ and $\mathbb{P}( Y |Z)$ via the  integral equation:
%\footnotesize
\begin{eqnarray}
\mu(z)=\int_{\Omega_{\boldsymbol W}}\int_{\Omega_X} k(z,x,{\boldsymbol w})\mathbb{E}[\partial_x Y_{x}|{\boldsymbol w}] dxd{\boldsymbol w},
\end{eqnarray}
%\normalsize
where  $\mu(z)=\mathbb{E}[Y|Z=z_0]-\mathbb{E}[Y|Z=z], k(z,x,{\boldsymbol w})=\mathfrak{p}(X\leq x,{\boldsymbol W}={\boldsymbol w}|Z=z)-\mathfrak{p}(X\leq x,{\boldsymbol W}={\boldsymbol w}|Z=z_0)$, and $z_0$ is a fixed value.
%\end{theorem}
}

\begin{proof}
%From the result of \citep{Wong2022}, the following integral equation holds under Assumptions 3.1 and 3.2:
%\jin
{First, we show the following integral equation holds under Assumptions \ref{AS1} and \ref{AS2} following the idea in \citep{Wong2022}:}
\begin{eqnarray}
&&\mathbb{E}[Y|Z=z,{\boldsymbol W}={\boldsymbol w}]-\mathbb{E}[Y|Z=z_0,{\boldsymbol W}={\boldsymbol w}]\\
&&=- \int_{\Omega_X}\{\mathbb{P}(X\leq x|Z=z,{\boldsymbol W}={\boldsymbol w})-\mathbb{P}(X\leq x|Z=z_0,{\boldsymbol W}={\boldsymbol w})\}
%\\&&\hspace{6cm}\times 
\mathbb{E}[\partial_xY_{x}|{\boldsymbol w}]dx.\nonumber
\end{eqnarray} 
%Here, we give the sketch of the proof as below.
From the setting of the IV, the following integral equation holds:
\begin{eqnarray}
Y_{X_{z}}= \int_{\Omega_X} \mathbbm{1}_{X_{z}=x}Y_{x}dx,
\end{eqnarray}
given ${\boldsymbol W}={\boldsymbol w}$ for each subject, where $\mathbbm{1}_{\cdot}$ is a delta function or indicator function. 
This equation means $X_{z}=x \Rightarrow Y_{X_{z}}=Y_{x}$ from the definition of delta function.
By substituting the integral equations $\displaystyle Y_{X_{z}}=\int_{\Omega_X} \mathbbm{1}_{X_{z}=x}Y_{x}dx$ and $\displaystyle Y_{X_{z_0}}= \int_{\Omega_X} \mathbbm{1}_{X_{z_0}=x}Y_{x}dx$, then
\begin{eqnarray}
Y_{X_{z}}-Y_{X_{z_0}}= \int_{\Omega_X} \{\mathbbm{1}_{X_{z}=x}-\mathbbm{1}_{X_{z_0}=x}\}Y_{x}dx
\end{eqnarray}
holds. Since the Heaviside step function is the integration of the delta function,
\begin{eqnarray}
&&Y_{X_{z}}-Y_{X_{z_0}}=\left[\{\mathbbm{1}_{X_{z}=x}-\mathbbm{1}_{X_{z_0}=x}\}\partial_xY_{x}\right]_{-\infty}^{\infty}- \int_{\Omega_X} \{\mathbbm{I}_{X_{z}\leq x}-\mathbbm{I}_{X_{z_0}\leq x}\}\partial_xY_{x}dx.
\end{eqnarray}
Because $\partial_xY_{x}<\infty$ for all $x \in \Omega_X$, $\displaystyle \left[ \{\mathbbm{1}_{X_{z}=x}-\mathbbm{1}_{X_{z_0}=x}\}\partial_xY_{x}\right]_{-\infty}^{\infty}=0$ holds.
Then, the integral equation becomes
\begin{eqnarray}
Y_{X_{z}}-Y_{X_{z_0}}=-\int_{\Omega_X} \{\mathbbm{I}_{X_{z}\leq x}-\mathbbm{I}_{X_{z_0}\leq x}\}\partial_xY_{x}dx.
\end{eqnarray}
From the separability with covariate $f_Y(X,{\boldsymbol W},{\boldsymbol H},{\boldsymbol u}_Y)=f_Y^1(X,{\boldsymbol W},{\boldsymbol u}_Y)+f_Y^2({\boldsymbol W},{\boldsymbol H},{\boldsymbol u}_Y)$, random variables $\mathbbm{I}_{X_{z}\leq x}-\mathbbm{I}_{X_{z_0}\leq x}$ and $\partial_xY_{x}$ are independent given ${\boldsymbol W}={\boldsymbol w}$.
Thus, we take expectations on both sides:
\begin{eqnarray}
&&\mathbb{E}[Y_{X_{z}}|{\boldsymbol W}={\boldsymbol w}]-\mathbb{E}[Y_{X_{z_0}}|{\boldsymbol W}={\boldsymbol w}]\\
&&=- \int_{\Omega_X} \mathbb{E}[\{\mathbbm{I}_{X_{z}\leq x}-\mathbbm{I}_{X_{z_0}\leq x}\}\partial_xY_{x}|{\boldsymbol W}={\boldsymbol w}]dx\\
&&=- \int_{\Omega_X}\{\mathbb{E}[\mathbbm{I}_{X_{z}\leq x}|{\boldsymbol W}={\boldsymbol w}]-\mathbb{E}[\mathbbm{I}_{X_{z_0}\leq x}|{\boldsymbol W}={\boldsymbol w}]\}\mathbb{E}[\partial_xY_{x}|{\boldsymbol w}]dx.
\end{eqnarray}
Then, the integral equation becomes
\begin{eqnarray}
\label{c1}
&&\mathbb{E}[Y|Z=z,{\boldsymbol W}={\boldsymbol w}]-\mathbb{E}[Y|Z=z_0,{\boldsymbol W}={\boldsymbol w}]\\
&&=-\int_{\Omega_X}\{\mathbb{P}(X\leq x|Z=z,{\boldsymbol W}={\boldsymbol w})-\mathbb{P}(X\leq x|Z=z_0,{\boldsymbol W}={\boldsymbol w})\}\mathbb{E}[\partial_xY_{x}|{\boldsymbol w}]dx.\nonumber
\end{eqnarray}

Next, %The uniqueness holds from the completeness of random variables $X_{z}$.
the integral equation can be given by multiplying $\mathfrak{p}({\boldsymbol W}={\boldsymbol w}|Z=z)$ and marginalizing for ${\boldsymbol W}$, then
\begin{eqnarray}
%&&\mathbb{E}[Y|Z=z,{\boldsymbol W}={\boldsymbol w}]=\int_{\Omega_X}\mathbb{P}(X \leq x|Z=z,{\boldsymbol W}={\boldsymbol w}) \mathbb{E}[\partial_x Y_{x}]dx\\
%&\Leftrightarrow&
&&\mathbb{E}_{\boldsymbol W}[\mathbb{E}[Y|Z=z,{\boldsymbol W}={\boldsymbol w}]]=\\
&&\int_{\Omega_X}\int_{\Omega_{\boldsymbol W}}\mathbb{P}(X \leq x|Z=z,{\boldsymbol W}={\boldsymbol w})\mathfrak{p}({\boldsymbol W}={\boldsymbol w}|Z=z) \mathbb{E}[\partial_x Y_{x}|{\boldsymbol w}]d{\boldsymbol w}dx\\
&\Leftrightarrow&\mathbb{E}[Y|Z=z]=\int_{\Omega_X}\int_{\Omega_{\boldsymbol W}}\mathfrak{p}(X \leq x,{\boldsymbol W}={\boldsymbol w}|Z=z)\mathbb{E}[\partial_x Y_{x}|{\boldsymbol w}]d{\boldsymbol w}dx\
\end{eqnarray}

Finally, we show the uniqueness of the solution. 
Since $X_z$ is a nontrivial function, there does not exist a function which satisfies $\mathbb{E}[\delta(X)|Z=z,{\boldsymbol W}={\boldsymbol w}]=0$ for any $z \in \Omega_Z$ and ${\boldsymbol w} \in \Omega_{\boldsymbol W}$.
Since $\mathbb{E}[\delta(X)|Z=z,{\boldsymbol W}={\boldsymbol w}]=\mathbb{E}[\delta(X),{\boldsymbol W}={\boldsymbol w}|Z=z]\mathbb{P}({\boldsymbol W}={\boldsymbol w})$, there exists a function which satisfies $\mathbb{E}[\delta(X),{\boldsymbol W}={\boldsymbol w}|Z=z]=0$ for any $z \in \Omega_Z$ and ${\boldsymbol w} \in \Omega_{\boldsymbol W}$ if there exists a function which satisfies $\mathbb{E}[\delta(X)|Z=z]=0$ for any $z \in \Omega_Z$ and ${\boldsymbol w} \in \Omega_{\boldsymbol W}$.
Taking a contraposition, there does not exist a function which satisfies $\mathbb{E}[\delta(X),{\boldsymbol W}={\boldsymbol w}|Z=z]=0$ for any $z \in \Omega_Z$ and ${\boldsymbol w} \in \Omega_{\boldsymbol W}$.
\end{proof}

%\begin{wrapfigure}{r}[1pt]{0.47\textwidth}
\begin{figure}
    \centering
    %\vspace{-1cm}
    \scalebox{1}{
\begin{tikzpicture}
    % x node set with absolute coordinates
    \node[mynode] (x) at (0,0) {$X$};
    \node[mynode] (y) at (3,0) {$Y$};
    \node[mynode] (z) at (-3,0) {$Z$};
    \node[myfillnode] (u) at (3,2) {${\boldsymbol H}$};
    \node[mynode] (w) at (0,2) {${\boldsymbol W}$};
    %\node[mynode] (d) at (-1.5,2) {${\boldsymbol u}_X$};
    %\node[mynode] (e) at (4.5,2) {${\boldsymbol u}_Y$};

    % Directed edge
    \path (x) edge[->] (y);
    \path (z) edge[->]  (x);

    \path (u) edge[->] (y);
    \path (u) edge[->]  (x);
    \path (u) edge[->]  (w);

    \path (w) edge[->] (y);
    \path (w) edge[->]  (x);

    \path (w) edge[->]  (z);
    
    %\path (e) edge[->] (y);
    %\path (d) edge[->]  (x);

\end{tikzpicture}
}
\vspace{-0cm}
    \caption{A causal graph representing the IV setting with covariates when there is an edge ${\boldsymbol W} \rightarrow Z$.}% Causal graph and two types of non-separability in the IV setting, ${\cal M}_{IV}$.}
    \label{DAG1d}
    
\vspace{-0cm}
\end{figure}
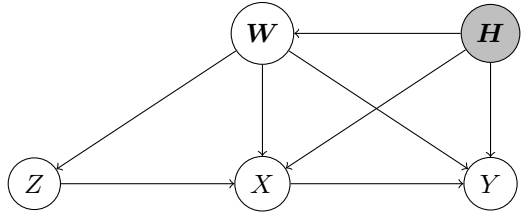
%\end{wrapfigure} 

\subsection{Identification theorem under IV model in Fig \ref{DAG1d}}
\label{appA2}
We consider the IV model with covariates represented by the causal graph in Fig \ref{DAG1d}, with the following SCM ${\cal M}_{IV}'$ over ${\boldsymbol V}=\{Z,X,Y,{\boldsymbol W}\}$ and ${\boldsymbol U}=\{{\boldsymbol H},{\boldsymbol u}_X,{\boldsymbol u}_Y,{\boldsymbol u}_Z,{\boldsymbol u}_{\boldsymbol W}\}$: %\jin{Do the results hold with Z = f(W, ) or an edge from W to Z in Figure 1?} \yuta{[If there is an edge from W to Z in Figure 1, the distribution $P(X,W|Z)$, say $P(X|Z)$, is biased.]}
%The SCM of the IV model, ${\cal M}_{IV}$, is defined as
%\begin{eqnarray}
%\vspace{-0.9cm}
\begin{equation}
%\small
\begin{gathered}
Y:=f_Y(X,{\boldsymbol W},{\boldsymbol H},{\boldsymbol u}_Y),\  X:=f_X(Z,{\boldsymbol W},{\boldsymbol H},{\boldsymbol u}_X),
{\boldsymbol W}:=f_{\boldsymbol W}({\boldsymbol H},{\boldsymbol u}_{\boldsymbol W}),\  
Z:=f_Z({\boldsymbol W},{\boldsymbol u}_Z),
%\left\{
%\begin{array}{l}
%    Y:=f_Y(X,{\boldsymbol W},{\boldsymbol H},{\boldsymbol u}_Y)\\
%    X:=f_X(Z,{\boldsymbol W},{\boldsymbol H},{\boldsymbol u}_X)\\
%    {\boldsymbol W}:=f_X({\boldsymbol H},{\boldsymbol u}_{\boldsymbol W})
%\end{array}
%\right.
%\end{eqnarray}
\end{gathered}
\end{equation}
%\normalsize
%with the conditional joint distribution $\mathbb{P}_{\{X,Y\}|Z}$.
where %$f_Y$, $f_X$, and $f_Z$ are scalar functions, and 
$f_{\boldsymbol W}$  is a vector function.  % ${\boldsymbol U}=\{{\boldsymbol H},{\boldsymbol u}_X,{\boldsymbol u}_Y,{\boldsymbol u}_Z,{\boldsymbol u}_{\boldsymbol W}\}$, and .
We assume all variables are continuous, %$Z$ are $M$-dimensional instrumental variables, 
${\boldsymbol W}$ are $d$-dimensional pre-treatment covariates, and 
 ${\boldsymbol H}$ stands for unmeasured confounders. 
 We show a similar identification result to Theorem \ref{TEO2}. 

{\bf Theorem 3.1'.}
{\it Under SCM ${\cal M}_{IV}'$ and Assumptions \ref{AS1} and \ref{AS2}, CAPCE $\mathbb{E}[\partial_x Y_{x}|{\boldsymbol w}]$ is identifiable from distributions $\mathbb{P}(X|Z,{\boldsymbol W})$ and $\mathbb{P}(Y|Z,{\boldsymbol W})$ via the integral equation:
\begin{eqnarray}
\label{d1}
\mu(z,{\boldsymbol w})=\int_{\Omega_X} k(z,x,{\boldsymbol w})\mathbb{E}[\partial_x Y_{x}|{\boldsymbol w}] dx,
\end{eqnarray}
where $\mu(z,{\boldsymbol w})=\mathbb{E}[Y|Z=z_0,{\boldsymbol W}={\boldsymbol w}]-\mathbb{E}[Y|Z=z,{\boldsymbol W}={\boldsymbol w}], k(z,x,{\boldsymbol w})=\mathfrak{p}(X\leq x|Z=z,{\boldsymbol W}={\boldsymbol w})-\mathfrak{p}(X\leq x|Z=z_0,{\boldsymbol W}={\boldsymbol w})$, and $z_0$ is a fixed value.}

\begin{proof}
    Eq. (\ref{d1}) is guaranteed by Eq. (\ref{c1}), which appears in the proof of Theorem \ref{TEO2}.
\end{proof}
Based on Theorem 3.1', we have to learn $\mathbb{E}[\partial_x Y_{x}|{\boldsymbol w}]$ as a function of $x$ for each ${\boldsymbol w} \in \Omega_{\boldsymbol W}$ respectively. In contrast, based on Theorem \ref{TEO2},  we can learn $\mathbb{E}[\partial_x Y_{x}|{\boldsymbol w}]$ directly as a function of $x$ and ${\boldsymbol w}$. 
%{Theorem 3.1' assumes the separabilty as Assumption 3.1 and 3.2. The disadvantage of Theorem 3.1' is that we have to learn $\mathbb{E}[\partial_x Y_{x}|{\boldsymbol w}]$ by functions on $x$ for each ${\boldsymbol w} \in \Omega_{\boldsymbol W}$ respectively. The advantage of Theorem 3.1 is that we can learn $\mathbb{E}[\partial_x Y_{x}|{\boldsymbol w}]$ by one function on $x$ and ${\boldsymbol w}$.}

{We perform experiments about estimating CAPCE based on Theorem 3.1' in Appendix~\ref{sec-prime}.}

\subsection{Derivation of RKHS CAPCE estimator}
%We give proof RKHS estimator following \citep{Singh2019}.
We show the detailed steps of deriving the RKHS CAPCE estimator.

{\bf RKHS estimator.} 
RKHS CAPCE estimator is given as $\hat{\mathbb{E}}[\partial_x{Y}_{x}|{\boldsymbol w}]=\hat{\boldsymbol \alpha}^T{\bf K}_{(X,{\boldsymbol W})^{(1)}(x,{\boldsymbol w})}$ where
\begin{eqnarray}
&&\hat{\boldsymbol \alpha}=(\hat{\bf O}\hat{\bf O}^T+N_2\xi {\bf K}_{(X,{\boldsymbol W})^{(1)}(X,{\boldsymbol W})^{(1)}}+N_2\lambda_3 {\bf I}_{N_2})^{-1}\hat{\bf O}\nonumber\\
&&\hspace{2cm}\times\{{\boldsymbol y}^{(2)T}({\bf K}_{Z^{(2)}Z^{(2)}}+N_2\lambda_2 {\bf I}_{N_2})^{-1}({\bf K}_{Z^{(2)}Z^{(2)}}-{\bf K}_{Z^{(2)}z_0})\}\nonumber\\
&&\hat{\bf O}={\bf K}_{(X,{\boldsymbol W})^{(1)}(X,{\boldsymbol W})^{(1)}}({\bf K}_{Z^{(1)}Z^{(1)}}+N_1\lambda_1 {\bf I}_{N_1})^{-1}({\bf K}_{Z^{(1)}Z^{(2)}}-{\bf K}_{Z^{(1)}z_0}),
\end{eqnarray}
$(\lambda_1,\lambda_2,\lambda_3,\xi)$ are regularization parameters, and ${\bf I}_N$ is a $N \times N$ identity matrix.

\begin{proof}
There are three optimization problems in RKHS estimator, {\bf Stage 1 (A)} {learning linear operator $G_1$}, {\bf Stage 1 (B)} {learning linear operator $G_2$}, and {\bf Stage 2} {learning linear operator $H$}.  We explain them respectively.

{\bf Stage 1 (A).}
We denote the feature map be $\psi(z)$ and $\pi(x,{\boldsymbol w})$, where $\pi(x,{\boldsymbol w})=-\int_{-\infty}^x \eta(x',{\boldsymbol w}) dx'$ for some feature function $\eta(x',{\boldsymbol w})$.
The optimization problem in {\bf Stage 1 (A)}  becomes
\begin{eqnarray}
    \min_{G_1 \in {\cal L}_2({\cal H}_Z,{\cal H}_{X,{\boldsymbol W}})} {N_1}^{-1}\sum\nolimits_{i=1}^{N_1}\left\|\pi(x_i^{(1)},{\boldsymbol w}_i^{(1)})-G_1(\psi(z_i^{(1)}))\right\|^2_{{\cal H}_{X,{\boldsymbol W}}}+\lambda_1\|G_1\|^2_{{\cal L}_2({\cal H}_Z,{\cal H}_{X,{\boldsymbol W}})}.
\end{eqnarray}
using ${\cal D}^{(1)}$. Then, the estimator $\hat{G_1}$ becomes
\begin{eqnarray}
\hat{G_1}(\cdot)=\left<{\pi_{X^{(1)},{\boldsymbol W}^{(1)}}}({\bf K}_{Z^{(1)}Z^{(1)}}+N_1 \lambda_1 {\bf I})^{-1}\psi_Z^{(1)T},\cdot\right>
\end{eqnarray}
where ${\bf K}_{Z^{(1)}Z^{(1)}}$ and ${\bf K}_{X^{(1)}X^{(1)}}$ are the empirical kernel matrices, the $i$-th column of ${\pi_{X^{(1)},{\boldsymbol W}^{(1)}}}$ is $\displaystyle -\int_{-\infty}^{x_i^{(1)}}\eta(x,{\boldsymbol w})dx$, and the $i$-th column of ${\psi_X}^{(1)}$ is $\psi(z_i^{(1)})$. 
The prediction values are
\begin{eqnarray}
    d_{0}(z)={\pi_{X^{(1)},{\boldsymbol W}^{(1)}}}({\bf K}_{ZZ}+N_1 \lambda_1 I)^{-1}\psi_Z^{(1)T}\psi(z)=-\sum_{i=1}^{N_1}\gamma_i(z)\int_{-\infty}^{x_i^{(1)}}\eta(x,{\boldsymbol w})dx
\end{eqnarray}
where $\gamma(z)=({\bf K}_{Z^{(1)}Z^{(1)}}+N_1 \lambda_1 {\bf I})^{-1}\psi_Z^{(1)T}\psi(z)=({\bf K}_{Z^{(1)}Z^{(1)}}+N_1 \lambda_1 {\bf I})^{-1}{\bf K}_{Z^{(1)}z}$.
Furthermore, the difference in the predication values are
\begin{eqnarray}
    d(z)=d_{0}(z)-d_{0}(z_0)=-\sum_{i=1}^{N_1}\{\gamma_i(z)-\gamma_i(z_0)\}\int_{-\infty}^{x_i^{(1)}}\eta(x,{\boldsymbol w})dx
\end{eqnarray}
 and $\gamma(z)-\gamma_i(z_0)=({\bf K}_{Z^{(1)}Z^{(1)}}+N_1 \lambda_1 {\bf I})^{-1}\psi_Z^{(1)T}\{\psi(z)-\psi(z_0)\}=({\bf K}_{Z^{(1)}Z^{(1)}}+N_1 \lambda_1 {\bf I})^{-1}({\bf K}_{Z^{(1)}z}-{\bf K}_{Z^{(1)}z_0})$ holds.
Letting $\hat{G_1}=\sum_{j=1}^{N_1}\alpha_j \eta(x_j^{(1)},{\boldsymbol w}_j^{(1)})$ since the optimal $\hat{G_1}$ exists in ${\text{span}}(\{\eta(x_j^{(1)},{\boldsymbol w}_j^{(1)})\}_{j=1}^{N_1})$ from the representer theorem \citep{Schlkopf2001}. 
Then the functional form of $d(z)$ is restricted by
\begin{eqnarray}
d(z)
    &=&-\left<\sum_{j=1}^{N_1}\alpha_j \int_{-\infty}^{x_i^{(1)}}\eta(x,{\boldsymbol w}_i^{(1)})dx,-\sum_{i=1}^{N_1}\{\gamma_i(z)-\gamma_i(z_0)\}\int_{-\infty}^{x_j^{(1)}}\eta(x,{\boldsymbol w}_j^{(1)})dx\right>\\
    &=&\sum_{i=1}^{N_1}\sum_{j=1}^{N_1}\alpha_j\{\gamma_i(z)-\gamma_i(z_0)\}\left< -\int_{-\infty}^{x_i^{(1)}}\eta(x,{\boldsymbol w}_i^{(1)})dx,-\int_{-\infty}^{x_j^{(1)}}\eta(x,{\boldsymbol w}_j^{(1)})dx\right>\\
    &=&\sum_{i=1}^{N_1}\sum_{j=1}^{N_1}\alpha_j\{\gamma_i(z)-\gamma_i(z_0)\}\left< \pi(x_i^{(1)},{\boldsymbol w}_i^{(1)}),\pi(x_j^{(1)},{\boldsymbol w}_j^{(1)})\right>.
    \end{eqnarray}
From the kernel trick, it becomes
    \begin{eqnarray}
    &=&\sum_{i=1}^{N_1}\sum_{j=1}^{N_1}\alpha_j\{\gamma_i(z)-\gamma_i(z_0)\}k((x_i^{(1)},{\boldsymbol w}_i^{(1)}),(x_j^{(1)},{\boldsymbol w}_j^{(1)}))\\
    &=& {\boldsymbol \alpha}^T w(z)
\end{eqnarray}
where $w(z)={\bf K}_{(X,{\boldsymbol W})^{(1)}(X,{\boldsymbol W})^{(1)}}({\bf K}_{ZZ}+N_1\lambda_1 {\bf I})^{-1}({\bf K}_{Z^{(1)}z}-{\bf K}_{Z^{(1)}z_0})$.
Note that the ${\boldsymbol \alpha}$ will be estimated in {\bf Stage 2.}

{\bf Stage 1 (B).}
The optimization problem in {\bf Stage 1 (B)} is %\jin{should $G_2 \in H_Z$ in the following?}
\begin{eqnarray}
\min_{G_2 \in {\cal L}_2({\cal H}_{{Z}},\Omega_Y)} {N_2}^{-1}\sum\nolimits_{i=1}^{N_2}\left\|y_i^{(2)}-G_2(\psi(z_i^{(2)}))\right\|^2+\lambda_2\|G_2\|^2_{{\cal L}_2({\cal H}_{{Z}},\Omega_Y)}
\end{eqnarray}
using ${\cal D}^{(2)}$. As {\bf Stage 1 (A)}, the estimator of $G_2$, $\hat{G_2}$, become
\begin{eqnarray}
\hat{G_2}(\cdot)=\left<{\boldsymbol y}^{(2)}({\bf K}_{Z^{(2)}Z^{(2)}}+N_2 \lambda_2 {\bf I})^{-1}\psi_Z^{(2)T},\cdot\right>
\end{eqnarray}
where ${\bf K}_{Z^{(2)}Z^{(2)}}$ are the gram matrices, the $i$-th column of ${\boldsymbol y}^{(2)}$ is $y_i^{(2)}$.
\begin{eqnarray}
    u_{0}(z)={\boldsymbol y}^{(2)T}({\bf K}_{Z^{(2)}Z^{(2)}}+N_2 \lambda_2 {\bf I})^{-1}\psi_Z^{(2)T}\psi(z)=\sum_{i=1}^{N_2}\gamma_i(z)\psi(z)
\end{eqnarray}
where $\gamma(z)={\boldsymbol y}^{(2)T}({\bf K}_{Z^{(2)}Z^{(2)}}+N_2 \lambda_2 I)^{-1}\psi_Z^{(2)T}$. Then,
\begin{eqnarray}
    u_{0}(z_0)={\boldsymbol y}^{(2)T}({\bf K}_{Z^{(2)}Z^{(2)}}+N_2\lambda_2 {\bf I})^{-1}{\bf K}_{Z^{(2)}z_0}
\end{eqnarray}
and, the difference of the predication values are 
\begin{eqnarray}
    u(z)=u_0(z)-u_0(z_0)={\boldsymbol y}^{(2)T}({\bf K}_{Z^{(2)}Z^{(2)}}+N_2\lambda_2 {\bf I})^{-1}({\bf K}_{Z^{(2)}z}-{\bf K}_{Z^{(2)}z_0}).
\end{eqnarray}
This is the estimator of $\mathbb{E}[Y|Z=z]-\mathbb{E}[Y|Z=z_0]$.

{\bf Stage 2.} 
The optimization problem in {\bf Stage 2} using ${\cal D}^{(2)}$ is 
%\jin{But you are using ${\cal D}^{(2)}$ in the following?}
\begin{eqnarray}
  &&\min_{H \in {\cal L}_2({\cal H}_{X,{\boldsymbol W}},\Omega_Y)} {N_2}^{-1}\sum\nolimits_{i=1}^{N_2}\left\|\hat{G}_2(\psi(z_i^{(2)})-\psi(z_0))-H(\hat{G}_1( \psi(z_i^{(2)})-\psi(z_0)))\right\|^2\nonumber\\
  &&\hspace{2cm}+\xi\|H\|^2_{{\cal L}_2({\cal H}_{X,{\boldsymbol W}},\Omega_Y)}+\lambda_3\|H\circ \hat{G}_1\|^2_{{\cal L}_2({\cal H}_Z,\Omega_Y)}.
\end{eqnarray}
Then, the estimation problem reduces to
\begin{eqnarray}
    &&\frac{1}{N_2}\sum_{i=1}^{N_2}(y_i^{(2)}-u_{0}(z_0)-{\boldsymbol \alpha}^T w(z))^2+\xi {\boldsymbol \alpha}^T{\bf K}_{XX}{\boldsymbol \alpha}+\lambda_3  {\boldsymbol \alpha}^T {\boldsymbol \alpha}\\
    &=&\frac{1}{N_2}\|{\boldsymbol y}^{(2)}-{\boldsymbol y}^{(2)T}({\bf K}_{Z^{(2)}Z^{(2)}}+N_2\lambda_2 {\bf I})^{-1}{\bf K}_{Z^{(2)}z_0}\\
    &&\hspace{0cm}-({\bf K}_{X^{(1)}X^{(1)}}({\bf K}_{Z^{(1)}Z^{(1)}}+N_1\lambda_1 {\bf I})^{-1}({\bf K}_{Z^{(1)}Z^{(2)}}-{\bf K}_{Z^{(1)}z_0}))^T{\boldsymbol \alpha}\|^2\\
    &&+\xi {\boldsymbol \alpha}^T{\bf K}_{(X,{\boldsymbol W})^{(1)}(X,{\boldsymbol W})^{(1)}}{\boldsymbol \alpha}+\lambda_3  {\boldsymbol \alpha}^T {\boldsymbol \alpha},
\end{eqnarray}
and the solution to this optimization problem can be represented as
\begin{eqnarray}
&&\hat{\boldsymbol \alpha}=(\hat{\bf O}\hat{\bf O}^T+N_2\xi {\bf K}_{(X,{\boldsymbol W})^{(1)}(X,{\boldsymbol W})^{(1)}}+N_2\lambda_3{\bf I})^{-1}\hat{\bf O}\\
&&\hspace{3cm}\times ({\boldsymbol y}^{(2)}-{\boldsymbol y}^{(2)T}({\bf K}_{Z^{(2)}Z^{(2)}}+N_2\lambda_2 {\bf I})^{-1}{\bf K}_{Z^{(2)}z_0})\\
&&\hat{\bf O}={\bf K}_{(X,{\boldsymbol W})^{(1)}(X,{\boldsymbol W})^{(1)}}({\bf K}_{Z^{(1)}Z^{(1)}}+N_1\lambda_1 {\bf I})^{-1}({\bf K}_{Z^{(1)}Z^{(2)}}-{\bf K}_{Z^{(1)}z_0}).
\end{eqnarray}
Finally, RKHS CAPCE estimator of $(x,{\boldsymbol w})$ becomes
$\hat{\mathbb{E}}[\partial_x{Y}_{x}|{\boldsymbol w}]=\hat{\boldsymbol \alpha}^T{\bf K}_{(X,{\boldsymbol W})(x,{\boldsymbol w})}$.
\end{proof}

\section{Consistency of Sieve CAPCE Estimator}
\label{appB}

In this section, we  show that sieve CAPCE estimator is consistent under
 assumptions similar to those guaranteeing the consistency of sieve NTSLS \citep{Whitney2003}.

\subsection*{Notations}
We introduce the notations for the assumptions. % of Theorem 4.1.
%First, we note that the combination of sieve CAPCE estimation and parametric CAPCE estimation is possible, and give the consistency theorem simultaneously.
%We denote the parametric and sieve paramator $(g,{\boldsymbol \gamma})=g$, and parameter space be ${\cal G}$. 
%$g$ is a functional parameter shown in sieve estimator, and ${\boldsymbol \gamma}$ is a vector parameter shown in parametric estimator.
%\begin{align}
  %\mathbb{E}[\partial_x Y_{x,{\boldsymbol w}}] =  \sum\nolimits_{j=1}^J\beta_{j} \psi_j(x,{\boldsymbol w})+g_0(x,{\boldsymbol w}) =  \sum\nolimits_{j=1}^J\beta_{j} \psi_j(x,{\boldsymbol w})+\sum\nolimits_{k=1}^{\infty}\gamma_{k} \sigma_k(x,{\boldsymbol w}),
  %\mathbb{E}[\partial_x Y_{x}|{\boldsymbol W}={\boldsymbol w}] = g_0 (x,{\boldsymbol w})+\sum_{k=1}^K\gamma_k\theta_k(x,{\boldsymbol w})=\sum_{j=1}^{\infty}\beta_{j} \phi_j(x,{\boldsymbol w})+\sum_{k=1}^K\gamma_k\theta_k(x,{\boldsymbol w}),
%\end{align}
%$g_0 \in {\cal G}$ is the true value of parameters.

{\bf Conditional Moment Restrictions.}
The estimation problem reduces to the problem called conditional moment restrictions, %\jin{Confusing writing. Do you mean "We will use the conditional moment restrictions method to solve the estimation problem"? }
and properties of the estimator are well studied \citep{Whitney2003,Ai2003}, and it is widely used in machine learning fields \citep{Kato2022}.
Since $\mathbb{E}[Y|Z=z_0]-\mathbb{E}[Y|Z]=\mathbb{E}[\mathbb{E}[Y|Z=z_0]-Y|Z]$ and 
$\mathbb{E}[\mathbbm{1}_{X\leq x,{\boldsymbol W}={\boldsymbol w}}|Z=z]-\mathbb{E}[\mathbbm{1}_{X\leq x,{\boldsymbol W}={\boldsymbol w}}|Z=z_0]=\mathbb{E}[\mathbbm{1}_{X\leq x,{\boldsymbol W}={\boldsymbol w}}-\mathbb{E}[\mathbbm{1}_{X\leq x,{\boldsymbol W}={\boldsymbol w}}|Z=z_0]|Z=z]$
, Theorem 3.1 reduces to
\begin{eqnarray}
\mathbb{E}\Big[(Y_{X_{z_0}}-Y)-\mathfrak{g}(X,X_{z_0},{\boldsymbol W},g)\Big|Z=z\Big]=0
\end{eqnarray}
where $\displaystyle \mathfrak{g}(X,X_{z_0},{\boldsymbol W},g)=\int_{\Omega_{\boldsymbol W}}\int_{\Omega_X}\{\mathbbm{1}_{X\leq x,{\boldsymbol W}={\boldsymbol w}}-\mathbbm{1}_{X_{z_0}\leq x,{\boldsymbol W}={\boldsymbol w}}\}g(X,{\boldsymbol W})dxd{\boldsymbol w}$.
%\jin{It's $g(... X_{z0},...)$ in (37)?} 
%where $\mathfrak{g}(X,X_{z_0},{\boldsymbol W},g)=\int_{\Omega_{\boldsymbol W}}\int_{\Omega_{X,{\boldsymbol W}}}\{\mathbbm{1}_{X\leq x,{\boldsymbol W}={\boldsymbol w}}-\mathbbm{1}_{X_{z_0}\leq x,{\boldsymbol W}={\boldsymbol w}}\}g(x,{\boldsymbol w})dxd{\boldsymbol w}$ for $z \in \Omega_{Z}$ and $m=1,\ldots,M$. 
We denote residual function $\rho(Y,Y_{X_{z_0}},X,X_{z_0},{\boldsymbol W},g)=(Y_{X_{z_0}}-Y)-\mathfrak{g}(X,{\boldsymbol W},g))$.
%${\rho}(Y,Y_{X_{z_0}},X,X_{z_0},{\boldsymbol W},g)=(\rho^1(Y,Y_{X_{z_0}},X,X_{z_0},{\boldsymbol W},g),\ldots,\rho^M(Y,Y_{X_{z_0}},X,X_{z_0},{\boldsymbol W},g))^T$ and $\mathfrak{g}(X,X_{z_0},{\boldsymbol W},g)=(\mathfrak{g}^1(X,X_{z_0},{\boldsymbol W},g),\ldots,\mathfrak{g}^M(X,X_{z_0},{\boldsymbol W},g))^T$.
Then, the integral equation can be represented by $\mathbb{E}[{\rho}(Y,Y_{X_{z_0}},X,X_{z_0},{\boldsymbol W},g)|Z]=0$. 
%In sieve NTSLS, the residuals is defined as $\rho(Y,X,{\boldsymbol W},g)=Y-g(X,{\boldsymbol W})$.

{\bf Consistency of Sieve CAPCE Estimator.} %\jin{Delete Parametric here?} 
First, we show consistency without compactness restriction. {The Sieve} CAPCE estimator reduces to the general form of the conditional moment restrictions method, which is well-studied in \citep{Whitney2003}, as below:
\begin{eqnarray}
    \hat{g}=\arg\min_{g \in {\cal G}}\sum_{i=1}^N\frac{1}{N}\hat{\rho}(z_i,g)^2,
\end{eqnarray}
where $\hat{\rho}(z_i,g)=\hat{c}_i-\hat{\boldsymbol d}_i{\boldsymbol \beta}$, and $\hat{\rho}(z_i,g)$ can be considered as the estimators of ${\mathbb{E}}[{\rho}(Y,Y_{X_{z_0}},X,X_{z_0},{\boldsymbol W},g)|Z=z_i]$.

\subsection*{Assumptions}
We make the following consistency assumptions introduced in \citep{Whitney2003}. % (All proof is in this paper).
We denote ${\cal G}_S=\{g \in{\cal G}: \|\mathfrak{g}_0(x,{\boldsymbol w})\|_{\tilde{W}^{l,2}}^2 \leq B_{S}\}$, and $\overline{{\cal G}_S}$ is a closure of ${\cal G}_S$.
\begin{assumption}[Uniqueness of $g$]
\label{A1}
    $g_0 \in {{\cal G}_S}$ is the only $g \in {{\cal G}_S}$ satisfying ${\mathbb{E}}[{\rho}(Y,Y_{X_{z_0}},X,X_{z_0},{\boldsymbol W},g)|Z=z]={0}$.
\end{assumption}
\begin{assumption}[Completeness of {\bf Stage 1.}]
\label{A2}
Taking limits $P\rightarrow \infty$, $N \rightarrow \infty$ with $P/N\rightarrow 0$,
     there {exists} ${\boldsymbol \pi}_P$ with $\mathbb{E}[\{b(z)-{\boldsymbol q}(z)^T{\boldsymbol \pi}_P\}^2]\rightarrow 0$
     for any $b(z)$ with $\mathbb{E}[b(z)^2]<\infty$. 
    %Also $\hat{\boldsymbol A}\xrightarrow{p} {\boldsymbol A}$, and ${\boldsymbol A}$ is positive definite and constant.
\end{assumption}
The above assumption
%\jin{Unclear what "this" refers to, use "The above assumption" or Assumption B.2"} 
is for the completeness of parameter space used in Stage 1.
\begin{assumption}[Boundedness of ${\rho}$]
\label{A3}
    $\mathbb{E}[\|{\rho}(Y,Y_{X_{z_0}},X,X_{z_0},{\boldsymbol W},g)\|^2|Z]$ is bounded and there exists $M(Y,Y_{X_{z_0}},X,X_{z_0},{\boldsymbol W})$, $\nu>0$ such that for all $\tilde{g},g \in \overline{{\cal G}_S}$, $\|{\rho}(Y,Y_{X_{z_0}},X,X_{z_0},{\boldsymbol W},\tilde{g})-{\rho}(Y,Y_{X_{z_0}},X,X_{z_0},{\boldsymbol W},g)\|\leq M(Y,Y_{X_{z_0}},X,X_{z_0},{\boldsymbol W})\|\tilde{g}- g\|_{W^{l,2}}^{\nu}$ and $\mathbb{E}[M(Y,Y_{X_{z_0}},X,X_{z_0},{\boldsymbol W})^2|Z]$ is bounded.
\end{assumption}
The above assumption is for the boundness of the parameters used in stage 2.
%\begin{assumption}[Compactness of $g$]
%\label{A4}
%    $g \in {\cal G}$ is compact for the norm $\|g\|_{W^{l,2}}$.
%\end{assumption}
%We denote a sub-space ${\cal G}_J=\{{\sum_{j=1}^J\beta_i\phi_j(x,{\boldsymbol w})},{\boldsymbol \gamma}\}$.
%\begin{assumption}[Completeness of ${\cal G}_J$]
%\label{A5}
%    For any $g \in {\cal G}$ there exists $g_J \in {\cal G}_J$ such that $\lim_{J \rightarrow \infty} \|g_J-g\|_{W^{l,2}}=0$.
%\end{assumption}

%\begin{theorem}
%    If Assumptions \ref{AS2}, \ref{AS1}, \ref{AS3}, \ref{A1}, \ref{A2}, \ref{A3}, \ref{A4} and \ref{A5} and $J \rightarrow \infty$, then $\|\hat{g}-g\|_{W^{l,2}}\xrightarrow{p} 0$.
%\end{theorem}

%We show that our Assumptions are weaker than their assumptions.
%\citep{Whitney2003} assume the separability $f_Y(X,{\boldsymbol W},{\boldsymbol U})=f_{Y_1}(X,{\boldsymbol W},{\boldsymbol \epsilon_Y})+f_{Y_2}({\boldsymbol U},{\boldsymbol \epsilon_Y})$, and let $\mathbb{E}[Y_{x,{\boldsymbol w}}]=f_{Y_1}(X,{\boldsymbol W};g)$. 
%We only assume the model $\mathbb{E}[\partial_x Y_{x,{\boldsymbol w}}]=g(X,{\boldsymbol W};g)$.

%Next, we show the consistency under compactness restriction. 
Let ${\cal W}$ denote the domain of $\mathfrak{g}(x,{\boldsymbol w},g)$.
\begin{assumption}[Openness and Convexness of Restricted Parameter Space]
\label{A6}
    %$g_B$ such that ${\boldsymbol \beta} \in \{{\boldsymbol \beta}^T{\boldsymbol \beta}\leq B_{\beta}\}$ and $g({\boldsymbol \gamma})\in \{ \sum_{m=1}^M\|\mathfrak{g}_0(x,{\boldsymbol w})\|_{\tilde{W}^{l,2}}^2 \leq B_{\gamma}\}$, 
    ${\cal W}$ is open and convex.
\end{assumption}

%\begin{theorem}[Consistency]
%    If Assumption \ref{AS2}, \ref{AS1}, \ref{B1}, \ref{AS3}, \ref{A1}, \ref{A2}, \ref{A3} are satisfied for $g \in \overline{{\cal G}_S}$, Assumption \ref{A6} is satisfied, and $J\rightarrow \infty$, then $\|\hat{g}-g\|_{W^{l,\infty}}\xrightarrow{p} 0$.
%\end{theorem}
%This theorem implies the uniform convergence of $\hat{g}_0$.
%$g=(B,\overline{\cal G})$ is satisfies if $B_{\beta}$ and $B_{\gamma}$ is large enough, or $\alpha$ is small enough.
%As mentioned in \citep{Newey2013}, ``{\it the bigger $\alpha$ is a the more weight the penalty has and so the less the variance and the larger the bias, with a $\alpha$ shrinking to zero as the sample size grows to ensure consistency."}
%\subsection*{Proof of Theorem 4.1.}

The following lemma is shown in \citep{Whitney2003}: 
\begin{lemma}
\label{COMLEM1}
If (i) $\Theta$ is a compact subset of a space with norm $\|\theta\|$: (ii) $\hat{Q}(\theta) \rightarrow_p Q(\theta)$ for all $\theta \in \Theta$: (iii) there is $v >0$ and $B_n O_p(1)$ such that for all $\tilde{\theta}, \theta \in \Theta$, $|\hat{Q}(\theta)-\hat{Q}(\tilde(\theta))| \leq B_n \Delta^v= B_n \epsilon/2M \leq \epsilon/2$ with a positive probability, then $Q(\theta)$ is continuous and $\sup_{\theta \in \Theta} |\hat{Q}(\theta)-Q(\theta)| \rightarrow_p 0$.

\end{lemma}

{\it
{\bf Theorem \ref{STHEO1}.}
Under SCM ${\cal M}_{IV}$ and Assumptions %3.1, 3.2, 4.1, 4.2, 4.3,
\ref{AS1}, \ref{AS2},  \ref{B1}, \ref{AS3}, \ref{COM}, \ref{A1}, \ref{A2}, \ref{A3}, and \ref{A6}, 
%    B.1, B.2, B.3, and B.4,
    letting $P \rightarrow \infty$ and $J\rightarrow \infty$, then $\|\hat{g}-g_0\|_{W^{l,\infty}}\xrightarrow{p} 0$.
}

\begin{proof}

From the Assumption \ref{A2} and \ref{A6}, the parameter space is compact subset.
From the Assumption \ref{A3}, the following relation is satisfied:
\begin{eqnarray}
    &&|{\rho}(Y,Y_{X_{z_0}},X,X_{z_0},{\boldsymbol W},\tilde{g})-{\rho}(Y,Y_{X_{z_0}},X,X_{z_0},{\boldsymbol W},g)|\\
    &&\leq M(Y,Y_{X_{z_0}},X,X_{z_0},{\boldsymbol W})\|\tilde{g}- g\|_{W^{l,2}}^{\nu}
\end{eqnarray}
From the Lemma \ref{COMLEM1},
\begin{eqnarray}
\|\tilde{g}- g\|_{W^{l,\infty}}  \rightarrow_p 0.
\end{eqnarray}
From Assumption \ref{A1}, the limits of $\tilde{g}$ is $g_0$.
\end{proof}
From the definition of ${W^{l,\infty}}$, this theorem means uniform convergence.

\section{Rate of Convergence of Sieve CAPCE Estimator}
%\citep{Ai2003} introduce theorem for the rate of convergence.
\label{appC}
\subsection*{Notations}

In this section, we explain the notations used in the assumptions for  Theorem 4.2 and Theorem 4.4.
%We introduce more detailed notations for rate of convergence.
Denote the estimation problem 
\begin{eqnarray}
    \inf_{g \in {\cal G}}\mathbb{E}\left[\mathfrak{g}(X,X_{z_0},{\boldsymbol W},g)^2\right]
\end{eqnarray}
and  introduce norm $\|\cdot\|_A$ as below:
\begin{eqnarray}
    \|g_1-g_0\|_A=\sqrt{\mathbb{E}\left[\left(\frac{d\mathfrak{g}(X,X_{z_0},{\boldsymbol W},g_0)}{dg}\right)^2\right]}
\end{eqnarray}
where
\begin{eqnarray}
    \frac{d{\rho}(Z,g_0)}{dg}[g-g_0]=\frac{d{\rho}(Z,(1-\tau)g_0+\tau g)}{d\tau} \text{ a.s. } Z
\end{eqnarray}
\begin{eqnarray}
    \frac{d{\rho}(Z,g_0)}{dg}[g_1-g_2]=\frac{d{\rho}(Z,g_0)}{dg}[g_1-g_0]-\frac{d{\rho}(Z,g_0)}{dg}[g_2-g_0]
\end{eqnarray}
\begin{eqnarray}
    \frac{d\mathfrak{g}(X,X_{z_0},{\boldsymbol W},g_0)}{dg}=\mathbb{E}\left[\frac{d{\rho}(Z,g_0)}{dg}[g_1-g_2] \Big|\{X,X_{z_0},{\boldsymbol W}\} \right].
\end{eqnarray}
These derivatives are called ``pathwise derivatives." See \citep{Ai2003} for details.

To evaluate the rate of convergence, we denote the number of the basis functions depending on sample size be $J_N$ and $P_N$. 
Note that $N \rightarrow \infty$ implies $J_N \rightarrow \infty$ and $P_N\rightarrow \infty$.
We use more basis functions, ${\boldsymbol q}^{P_
N}=(q^1,\ldots,q^{P_N})$, as the sample size grows for the stage 1.

\subsection*{Assumptions}
We make the following assumptions. 
\begin{assumption}[Compactness of Domain]
\label{RA1}
$\Omega_{(X,X_{z_0},{\boldsymbol W})}$ is compact with non empty interior.
\end{assumption}
\begin{assumption}[Order of Convergence of Stage 1]
\label{RA2}
For  any $h \in {\cal G}_S$ with $\kappa>(1+d)/2$, there exists ${\boldsymbol q}^{P_
N}(X,X_{z_0},{\boldsymbol W})^T{\boldsymbol \pi}_{P_N} \in {\cal G}_S$, where ${\boldsymbol \pi}_{P_N}$ is $P_N$ vector, such that $\sup_{(X,X_{z_0},{\boldsymbol W}) \in \Omega_{(X,X_{z_0},{\boldsymbol W})}}|h(X,X_{z_0},{\boldsymbol W})-{\boldsymbol q}^{P_
N}(X,X_{z_0},{\boldsymbol W})^T{\boldsymbol \pi}_{P_N}|={\cal O}(P_N^{-\kappa/(1+d)})$ and $P_N^{-\kappa/(1+d)}={o}(N^{-1/4})$.
\end{assumption}
The above assumption guarantees the order of convergence of regression (basis functions) used in Stage 1.
%\begin{assumption}[Oder of Weights]
%\label{RA3}
%$\hat{\Sigma}(X,X_{z_0},{\boldsymbol W})={\Sigma}(X,X_{z_0},{\boldsymbol W})+o_p({N^{-1/4}})$ uniformly over $(X,X_{z_0},{\boldsymbol W})\in \Omega_{(X,X_{z_0},{\boldsymbol W})}$.
%\end{assumption}
\begin{assumption}[Order of Convergence of Stage 2]
\label{RA4}
There is a constant $\mu_1>0$ such  that for any $g \in {\cal G}$, there is $\Pi g \in {\cal G}$ satisfying $\|\Pi g-g\|={\cal O}(J_N^{-\kappa/(1+d)})$ and $J_N^{-\kappa/(1+d)}={o}(N^{-1/4})$. $\Pi$ is  the projections to ${\cal G}$.
\end{assumption}
The above assumption guarantees the order of convergence of regression (basis functions) used in Stage 2.
\begin{assumption}[Envelope condition]
\label{RA5}
Each element of ${\rho}(Z,g)$ satisfies the envelope condition in $g \in {\cal G}$;  and, each element of ${\rho}(Z,g) \in {\cal G}_S$ with $\kappa>(1+d)/2$.
\end{assumption}
The envelope condition is shown in \citep{Milgrom2002}. 
%\jin{"Envelope" or "Envelop"?}

Denote $\xi_N=\sup_{(X,X_{z_0},{\boldsymbol W})}\|{\boldsymbol q}^{P_
N}(X,X_{z_0},{\boldsymbol W})\|$.
\begin{assumption}[Condition of $J_N$]
\label{RA6}
$J_N\times ln(N)\times \xi_N\times N^{-1/2}=o(1)$
\end{assumption}
We denote $N(\epsilon^{1/k},{\cal G},\|\cdot\|_{W^{l,2}})$ as the minimal number of radius $\delta$ covering ball of ${\cal G}$. 
\begin{assumption}[Condition of $J_N$]
\label{RA7}
$ln[N(\epsilon^{1/k},{\cal G},\|\cdot\|_{W^{l,2}})] \leq const. \times J_N \times ln(J_N/\epsilon)$
\end{assumption}
These assumptions show how to make the models complex depending on sample size.

\begin{assumption}[Convexness of Parameter Space]
\label{RA8}
${\cal G}$ is convex in $g$, and ${\rho}(Z,g)$ is pathwise differentiable at $g$; and, for some $c_1,c_2>0$,
\begin{eqnarray}
    c_1\mathbb{E}[\hat{\rho}(Z,g)^2]\leq \|\hat{g}-g\|^2 \leq c_2\mathbb{E}[\hat{\rho}(Z,g)^2]
\end{eqnarray}
holds for all $\hat{g} \in {\cal G}$ with $\|\hat{g}-g\|_{W^{l,2}}^2=o(1)$
\end{assumption}

The following lemma holds \citep{Ai2003}: %\jin{the lemma needs a proof or citation?}
\begin{lemma}
\label{LEM2}
Under Assumptions \ref{RA1}, \ref{RA2}, \ref{RA4}, \ref{RA5}, \ref{RA6}, \ref{RA7}, and \ref{RA8},
(i) $\hat{L}_N(g)-L_N(g)=o_p(N^{-1/4})$ uniformly over $g \in {\cal G}$; and (ii) $\hat{L}_N(g)-\hat{L}_N(g_0)-\{L_N(g)-L_N(g_0)\}=o_p(\tau_N N^{-1/4})$ uniformly over $g \in {\cal G}$ with $\|g-g_0\|\leq o(\tau_N)$, where $\tau_N=N^{-\tau}$ with $\tau\leq 1/4$.
\end{lemma}

{\it
{\bf Theorem \ref{STHEO2}.}
Under SCM ${\cal M}_{IV}$ and Assumptions %3.1, 3.2, 4.1, 4.2, 4.3,
\ref{AS1}, \ref{AS2},  \ref{B1}, \ref{AS3}, \ref{COM}, 
\ref{RA1}, \ref{RA2}, \ref{RA4}, \ref{RA5}, \ref{RA6}, \ref{RA7}, and \ref{RA8},
    %C.1, C.2, C.3, C.4, C.5, C.6, and C.7, 
    setting $N=N_1=N_2$, then $\|\hat{g}-{g}_0\|_{A}={o}_p(N^{-1/4})$.
}
\begin{proof}
Let 
\begin{eqnarray}
    \hat{L}_{N}(g)=-\frac{1}{2N}\hat{\mathfrak{g}}(X,X_{z_0},{\boldsymbol W}.g)^2,\ \ \  {L}_{N}(g)=-\frac{1}{2N}{\mathfrak{g}}(X,X_{z_0},{\boldsymbol W},g)^2.
\end{eqnarray}
%then the problem reduces to the problem shown in \citep{Ai2003}. \jin{ don't refer to some problem in another place, }

Then, Lemma \ref{LEM2} implies 
\begin{eqnarray}
    \hat{L}_{N}(g)-\hat{L}_{N}(g_0)-\{{L}_{N}(g)-{L}_{N}(g_0)\}=o_p(N^{-1/4})
\end{eqnarray}
and this proves
\begin{eqnarray}
    \|\hat{g}-g_0\|=o_p(N^{-1/4}).
\end{eqnarray}
\end{proof}

\section{Consistency of Parametric CAPCE Estimator}
\label{appD}

In this section, we show the consistency property of parametric CAPCE estimator.
We denote the functional space ${\cal G}$ be $\{g \in {\cal G}: g(x,{\boldsymbol w})=\sum_{k=1}^K \gamma_k \theta_k(x,{\boldsymbol w})\}$.

{\bf Consistency of Parametric CAPCE Estimator.} %\jin{Delete Sieve} 
First, we show consistency without compactness restriction. {The Parametric} CAPCE estimator reduces to the general form of the conditional moment restrictions method, which is well-studied in \citep{Whitney2003}, as below:
\begin{eqnarray}
    \hat{{\boldsymbol \gamma}}=\arg\min_{{\boldsymbol \gamma}}\sum_{i=1}^N\frac{1}{N}\hat{\rho}(z_i,{\boldsymbol \gamma})^2,
\end{eqnarray}
where $\hat{\rho}(z_i,{\boldsymbol \gamma})=\hat{c}_i-\hat{\boldsymbol e}_i{\boldsymbol \gamma}$.
$\hat{\rho}(z_i,{\boldsymbol \gamma})$ can be considered as the estimators of ${\mathbb{E}}[{\rho}(Y,Y_{X_{z_0}},X,X_{z_0},{\boldsymbol W},{\boldsymbol \gamma})|Z=z_i]$.

\subsection*{Assumptions}
We make the following  assumptions introduced in \citep{Whitney2003}. % (All proof is in this paper).
We denote ${\cal G}_P=\{{\boldsymbol \gamma}^T{\boldsymbol \gamma} \leq B_{P}\}$, and $\overline{{\cal G}_P}$ is the closure of ${\cal G}_P$.
\begin{assumption}[Uniqueness of $g$]
\label{PA1}
    ${\boldsymbol \gamma} \in {{\cal G}_P}$ is the only ${\boldsymbol \gamma} \in {{\cal G}_P}$ satisfying ${\mathbb{E}}[{\rho}(Y,Y_{X_{z_0}},X,X_{z_0},{\boldsymbol W},{\boldsymbol \gamma})|Z=z]={\boldsymbol 0}$.
\end{assumption}
\begin{assumption}[Completeness of ${\boldsymbol q}$]
\label{PA2}
Taking limits $P\rightarrow \infty$, $N \rightarrow \infty$ with $P/N\rightarrow 0$,
     there exists ${\boldsymbol \pi}_P$ with $\mathbb{E}[\{b(z)-{\boldsymbol q}(z)^T{\boldsymbol \pi}_P\}^2]\rightarrow 0$
     for any $b(z)$ with $\mathbb{E}[b(z)^2]<\infty$. %Also $\hat{\boldsymbol A}\xrightarrow{p} {\boldsymbol A}$, and ${\boldsymbol A}$ is positive definite and constant.
\end{assumption}
\begin{assumption}[Boundedness of ${\rho}$]
\label{PA3}
    $\mathbb{E}[\|{\rho}(Y,Y_{X_{z_0}},X,X_{z_0},{\boldsymbol W},{\boldsymbol \gamma})\|^2|Z]$ is bounded and there exists $M(Y,Y_{X_{z_0}},X,X_{z_0},{\boldsymbol W})$, $\nu>0$ such that for all $\tilde{{\boldsymbol \gamma}},{\boldsymbol \gamma} \in \overline{{\cal G}_P}$, $\|{\rho}(Y,Y_{X_{z_0}},X,X_{z_0},{\boldsymbol W},\tilde{{\boldsymbol \gamma}})-{\rho}(Y,Y_{X_{z_0}},X,X_{z_0},{\boldsymbol W},g)\|\leq M(Y,Y_{X_{z_0}},X,X_{z_0},{\boldsymbol W})\|\tilde{{\boldsymbol \gamma}}- {\boldsymbol \gamma}\|^{\nu}$ and $\mathbb{E}[M(Y,Y_{X_{z_0}},X,X_{z_0},{\boldsymbol W})^2|Z]$ is bounded.
\end{assumption}
%\begin{assumption}[Compactness of $g$]
%\label{A4}
%    $g \in {\cal G}$ is compact for the norm $\|g\|_{W^{l,2}}$.
%\end{assumption}
%We denote a sub-space ${\cal G}_J=\{{\sum_{j=1}^J\beta_i\phi_j(x,{\boldsymbol w})},{\boldsymbol \gamma}\}$.
%\begin{assumption}[Completeness of ${\cal G}_J$]
%\label{A5}
%    For any $g \in {\cal G}$ there exists $g_J \in {\cal G}_J$ such that $\lim_{J \rightarrow \infty} \|g_J-g\|_{W^{l,2}}=0$.
%\end{assumption}

%\begin{theorem}
%    If Assumptions \ref{AS2}, \ref{AS1}, \ref{AS3}, \ref{A1}, \ref{A2}, \ref{A3}, \ref{A4} and \ref{A5} and $J \rightarrow \infty$, then $\|\hat{g}-g\|_{W^{l,2}}\xrightarrow{p} 0$.
%\end{theorem}

%We show that our Assumptions are weaker than their assumptions.
%\citep{Whitney2003} assume the separability $f_Y(X,{\boldsymbol W},{\boldsymbol U})=f_{Y_1}(X,{\boldsymbol W},{\boldsymbol \epsilon_Y})+f_{Y_2}({\boldsymbol U},{\boldsymbol \epsilon_Y})$, and let $\mathbb{E}[Y_{x,{\boldsymbol w}}]=f_{Y_1}(X,{\boldsymbol W};g)$. 
%We only assume the model $\mathbb{E}[\partial_x Y_{x,{\boldsymbol w}}]=g(X,{\boldsymbol W};g)$.

%Next, we show the consistency under compactness restriction. 
Let ${\cal W}$ denote the domain of $\mathfrak{g}(x,{\boldsymbol w},{\boldsymbol \gamma})$.
\begin{assumption}[Openness and Convexness of Restricted Parameter Space]
\label{PA6}
    %$g_B$ such that ${\boldsymbol \beta} \in \{{\boldsymbol \beta}^T{\boldsymbol \beta}\leq B_{\beta}\}$ and $g({\boldsymbol \gamma})\in \{ \sum_{m=1}^M\|\mathfrak{g}_0(x,{\boldsymbol w})\|_{\tilde{W}^{l,2}}^2 \leq B_{\gamma}\}$, 
    ${\cal W}$ is open and convex.
\end{assumption}

%\begin{theorem}[Consistency]
%    If Assumption \ref{AS2}, \ref{AS1}, \ref{B1}, \ref{AS3}, \ref{A1}, \ref{A2}, \ref{A3} are satisfied for $g \in \overline{{\cal G}_S}$, Assumption \ref{A6} is satisfied, and $J\rightarrow \infty$, then $\|\hat{g}-g\|_{W^{l,\infty}}\xrightarrow{p} 0$.
%\end{theorem}
%This theorem implies the uniform convergence of $\hat{g}_0$.
%$g=(B,\overline{\cal G})$ is satisfies if $B_{\beta}$ and $B_{\gamma}$ is large enough, or $\alpha$ is small enough.
%As mentioned in \citep{Newey2013}, ``{\it the bigger $\alpha$ is a the more weight the penalty has and so the less the variance and the larger the bias, with a $\alpha$ shrinking to zero as the sample size grows to ensure consistency."}

%\subsection*{Proof of Theorem 4.3.}

{\it
{\bf Theorem \ref{PTHEO1}.}
    Under SCM ${\cal M}_{IV}$ and Assumptions %3.1, 3.2, 4.4,
    \ref{AS1}, \ref{AS2},  \ref{COM2},
    \ref{PA1}, \ref{PA2}, \ref{PA3}, and \ref{PA6},
    %D.1, D.2, D.3, and D.4,
    letting $P \rightarrow \infty$, then $\|\hat{\boldsymbol \gamma}-{\boldsymbol \gamma}\|\xrightarrow{p} 0$.
}

\begin{proof}

From the Assumption \ref{PA2} and \ref{PA6}, the parameter space is compact subset.
From the Assumption \ref{PA3}, the following relation is satisfied:
\begin{eqnarray}
    &&|{\rho}(Y,Y_{X_{z_0}},X,X_{z_0},{\boldsymbol W},\tilde{{\boldsymbol \gamma}})-{\rho}(Y,Y_{X_{z_0}},X,X_{z_0},{\boldsymbol W},{\boldsymbol \gamma})|\\
    &&\leq M(Y,Y_{X_{z_0}},X,X_{z_0},{\boldsymbol W})\|\tilde{{\boldsymbol \gamma}}- {\boldsymbol \gamma}\|^{\nu}
\end{eqnarray}
From Lemma \ref{COMLEM1},
\begin{eqnarray}
\|\tilde{{\boldsymbol \gamma}}- {\boldsymbol \gamma}\|  \rightarrow_p 0.
\end{eqnarray}
From Assumption \ref{PA1}, the limits is ${\boldsymbol \gamma}_0$.
\end{proof}

\section{Rate of Convergence of Parametric CAPCE Estimator}
%\citep{Ai2003} introduce theorem for the rate of convergence.
\label{appE}

\subsection*{Assumptions}
We make the following assumptions. 
\begin{assumption}[Compactness of Domain]
\label{PRA1}
$\Omega_{(X,X_{z_0},{\boldsymbol W})}$ is compact with non empty interior.
\end{assumption}
\begin{assumption}[Order of Convergence of Stage 1]
\label{PRA2}
For  any $h \in {\cal G}_P$ with $\kappa>(1+d)/2$, there exists ${\boldsymbol q}^{P_
N}(X,X_{z_0},{\boldsymbol W})^T{\boldsymbol \pi}_{P_N} \in {\cal G}_P $, where ${\boldsymbol \pi_{P_N}}$ is $P_N$ vector, such that $\sup_{(X,X_{z_0},{\boldsymbol W}) \in \Omega_{(X,X_{z_0},{\boldsymbol W})}}|h(X,X_{z_0},{\boldsymbol W})-{\boldsymbol q}^{P_
N}(X,X_{z_0},{\boldsymbol W})^T{\boldsymbol \pi}_{P_N}|={\cal O}(P_N^{-\kappa/(1+d)})$ and $P_N^{-\kappa/(1+d)}={o}(N^{-1/4})$.
\end{assumption}
\begin{assumption}[Order of Convergence of Stage 2]
\label{PRA4}
There is a constant $\mu_1>0$ such  that for any ${\boldsymbol \gamma} \in {\cal G}_P$, there is $\Pi {\boldsymbol \gamma} \in {\cal G}_P$ 
%\jin{What is ${\cal G}$?}
satisfying $\|\Pi {\boldsymbol \gamma}-{\boldsymbol \gamma}\|={\cal O}(1)$.% and $J_N^{-B/(1+d)}={o}(N^{-1/4})$.
\end{assumption}
\begin{assumption}[Envelope condition]
\label{PRA5}
Each element of ${\rho}(Z,{\boldsymbol \gamma})$ satisfies the envelope condition in ${\boldsymbol \gamma} \in {\cal G}_P$; and, each element of ${\rho}(Z,{\boldsymbol \gamma}) \in {\cal G}_P$ with $\kappa>(1+d)/2$, for all ${\boldsymbol \gamma} \in {\cal G}_P$.
\end{assumption}
The envelope condition is shown in \citep{Milgrom2002}. %\jin{${\cal G}_P$ in the above and following?} 
%Denote $\xi_N=\sup_{(X,X_{z_0},{\boldsymbol W})}\|{\boldsymbol q}^{P_N}(X,X_{z_0},{\boldsymbol W})\|$.
%\begin{assumption}[Condition of $J_N$]
%\label{PRA6}
%$J_N\times ln(N)\times \xi\times N^{-1/2}=o(1)$
%\end{assumption}
%We denote $N(\epsilon^{1/k},{\cal G},\|\cdot\|_{W^{l,2}})$ as the minimal number of radius $\delta$ covering ball of ${\cal G}$. 
%\begin{assumption}[Condition of $J_N$]
%\label{PRA7}
%$ln[N(\epsilon^{1/k},{\cal G},\|\cdot\|_{W^{l,2}})] \leq const. \times J_N \times ln(J_N/\epsilon)$
%\end{assumption}
\begin{assumption}[Convexness of Parameter Space]
\label{PRA8}
${\cal G}_P$ is convex in ${\boldsymbol \gamma}$, and ${\rho}(Z,{\boldsymbol \gamma})$ is pathwise differentiable at ${\boldsymbol \gamma}$; and, for some $c_1,c_2>0$,
\begin{eqnarray}
    c_1\mathbb{E}[\hat{\rho}(Z,{\boldsymbol \gamma})^2]\leq \|\hat{{\boldsymbol \gamma}}-{\boldsymbol \gamma}\|^2 \leq c_2\mathbb{E}[\hat{\rho}(Z,{\boldsymbol \gamma})^2]
\end{eqnarray}
holds for all $\hat{{\boldsymbol \gamma}} \in {\cal G}_P$ with $\|\hat{{\boldsymbol \gamma}}-{\boldsymbol \gamma}\|^2=o(1)$
\end{assumption}

%\begin{theorem}[Rate of Convergence]
%    Give SCM ${\cal M}_{IV}$ and Assumption \ref{AS2}, \ref{AS1}, \ref{B1}, \ref{AS3}, \ref{RA1}, \ref{RA2}, \ref{RA3}, \ref{RA4}, \ref{RA5}, \ref{RA6}, \ref{RA7}, and \ref{RA8}, then $\|\hat{g}-{g}_0\|_{W^{l,2}}={o}_p(N^{-1/4})$.
%\end{theorem}

%\subsection*{Proof of Theorem 4.4.}

The following lemma holds \citep{Ai2003}: %\jin{Proof or citation}
\begin{lemma}
\label{LEM3}
Under Assumptions \ref{PRA1}, \ref{PRA2}, \ref{PRA4}, \ref{PRA5}, and \ref{PRA8},
(i) $\hat{L}_N({\boldsymbol \gamma})-L_N({\boldsymbol \gamma})=o_p(N^{-1/4})$ uniformly over ${\boldsymbol \gamma}$; and (ii) $\hat{L}_N({\boldsymbol \gamma})-\hat{L}_N({\boldsymbol \gamma}_0)-\{L_N({\boldsymbol \gamma})-L_N({\boldsymbol \gamma}_0)\}=o_p(\tau_N N^{-1/4})$ uniformly over ${\boldsymbol \gamma}$ with $\|{\boldsymbol \gamma}-{\boldsymbol \gamma}_0\|\leq o(\tau_N)$, where $\tau_N=N^{-\tau}$ with $\tau\leq 1/4$.
\end{lemma}

{\it
{\bf Theorem \ref{PTHEO2}.}
Under SCM ${\cal M}_{IV}$ and Assumptions %3.1, 3.2, 4.4,
\ref{AS1}, \ref{AS2}, \ref{COM2},
 \ref{PRA1}, \ref{PRA2}, \ref{PRA4}, \ref{PRA5}, and \ref{PRA8},
% E.1, E.2, E.3, E.4, and E.5,
 setting $N=N_1=N_2$, then $\|\hat{\boldsymbol \gamma}-{\boldsymbol \gamma}\|={o}_p(N^{-1/4})$.
}

\begin{proof}

Let 
\begin{eqnarray}
    \hat{L}_{N}({\boldsymbol \gamma})=-\frac{1}{2N}\hat{\mathfrak{g}}(X,X_{z_0},{\boldsymbol W},{\boldsymbol \gamma})^2, \ \ \ {L}_{N}(g)=-\frac{1}{2N}{\mathfrak{g}}(X,X_{z_0},{\boldsymbol W},{\boldsymbol \gamma})^2.
\end{eqnarray}
%then the problem reduces to the problem in \citep{Ai2003}. \jin{Revise this proof.}
Then, Lemma \ref{LEM3} implies %\jin{Lemma 3?}
\begin{eqnarray}
    \hat{L}_{N}({\boldsymbol \gamma})-\hat{L}_{N}({\boldsymbol \gamma}_0)-\{{L}_{N}({\boldsymbol \gamma})-{L}_{N}({\boldsymbol \gamma}_0)\}=o_p(N^{-1/4})
\end{eqnarray}
and this proves
\begin{eqnarray}
    \|\hat{{\boldsymbol \gamma}}-{\boldsymbol \gamma}_0\|=o_p(N^{-1/4}).
\end{eqnarray}

\end{proof}

\section{Properties of RKHS CAPCE Estimator}
\label{appF}
We show the consistency and rate of convergence of RKHS CAPCE estimator following \citep{Singh2019} when $\lambda_3$ is $0$.

\subsection*{Notations}
We use the integral operator notations from the kernel methods literature. ${\cal L}_2(\Omega_Z,\mathfrak{p}_{Z})$ denotes a  ${\cal L}_2$ integrable function from $\Omega_{Z}$ to $\Omega_Y$ with respect to measure $\mathfrak{p}_{Z}$.
\begin{definition}
The stage 1 operators are
\begin{eqnarray}
    &&S^*_1: {\cal H}_{Z} \rightarrow {\cal L}_2(\Omega_Z,\mathfrak{p}_{Z}), l \mapsto \left<l,\psi(\cdot)\right>_{{\cal H}_{\boldsymbol Z}}\\ 
    &&S_1: {\cal L}_2(\Omega_Z,\mathfrak{p}_{Z}) \rightarrow {\cal H}_{Z}, \tilde{l} \mapsto \int \psi(z) \tilde{l}(z) \mathfrak{p}_{Z}(z) dz
\end{eqnarray}
and $T_1=S^*_1 \circ S_1$ is the uncentered covariance operator.
The details of the theory of vector-valued RKHS are shown in \citep{Singh2019}.
\end{definition}

In addition, we denote
\begin{definition}
\begin{eqnarray}
{G_1}_{\rho}=\arg \min {\cal E}_1 (E), {\cal E}_1=\mathbb{E}[\pi(X,{\boldsymbol W})-{G_1}(\psi(Z))]_{{\cal H}_{X,{\boldsymbol W}}}^2,
\end{eqnarray}
\begin{eqnarray}
{G_1}_{\lambda}=\arg \min {\cal E}_1 ({G_1}), {\cal E}_1=\mathbb{E}[\pi(X,{\boldsymbol W})-{G_1}(\psi(Z))]_{{\cal H}_{X,{\boldsymbol W}}}^2+\lambda \|G_1\|_{{\cal L}_2({\cal H}_Z,{\cal H}_{X,{\boldsymbol W}})}^2,
\end{eqnarray}
\begin{eqnarray}
\hat{{G_1}}_{\lambda}=\arg \min {\cal E}_1 ({G_1}), {\cal E}_1=\hat{\mathbb{E}}[\pi(X,{\boldsymbol W})-{G_1}(\psi(Z))]_{{\cal H}_{X,{\boldsymbol W}}}^2+\lambda \|G_1\|_{{\cal L}_2({\cal H}_Z,{\cal H}_{X,{\boldsymbol W}})}^2,
\end{eqnarray}
\begin{eqnarray}
{G_2}_{\rho}=\arg \min {\cal E}_1 (E), {\cal E}_1=\mathbb{E}[Y-{G_2}(\psi(Z))]^2,
\end{eqnarray}
\begin{eqnarray}
{G_2}_{\lambda}=\arg \min {\cal E}_1 ({G_2}), {\cal E}_1=\mathbb{E}[Y-{G_2}(\psi(Z))]^2+\lambda \|G_2\|_{{\cal L}_2({\cal H}_Z,\Omega_Y)}^2,
\end{eqnarray}
\begin{eqnarray}
\hat{{G_2}}_{\lambda}=\arg \min {\cal E}_1 ({G_2}), {\cal E}_1=\hat{\mathbb{E}}[Y-{G_2}(\psi(Z))]^2+\lambda \|G_2\|_{{\cal L}_2({\cal H}_Z,\Omega_Y)}^2.
\end{eqnarray}

\end{definition}

\begin{definition}
The stage 2 operators are
\begin{eqnarray}
    &&S_2^*: {\cal L}_2({\cal H}_Z,{\cal H}_{X,{\boldsymbol W}}) \rightarrow {\cal L}_2({\cal H}_{X,{\boldsymbol W}},\mathfrak{p}_{{\cal H}_{X,{\boldsymbol W}}}),H \mapsto \Omega^*_{(\cdot)}H \\
    &&S_2: {\cal L}_2({\cal H}_{X,{\boldsymbol W}},\mathfrak{p}_{{\cal H}_{X,{\boldsymbol W}}}) \rightarrow {\cal L}_2({\cal H}_Z,{\cal H}_{X,{\boldsymbol W}}),\\
    &&\tilde{H} \mapsto \int \Omega_{\mu(z)-\mu(z_0)}\circ \tilde{H}\{\mu(z)-\mu(z_0)\} \mathfrak{p}_{{\cal H}_{X,{\boldsymbol W}}}(\mu(z))
\end{eqnarray}
 and $T_2=S^*_2 \circ S_2$ is the uncentered covariance operator.
\end{definition}

\begin{definition}
    
We denote
\begin{eqnarray}
H_{\rho}=\arg \min {\cal E} (H), {\cal E}(H)=\mathbb{E}[Y-\mu_2(z_0)-H(\mu(Z)-\mu(z_0))]_{{\cal H}_{X,{\boldsymbol W}}}^2,
\end{eqnarray}
\begin{eqnarray}
&&H_{\xi}=\arg \min {\cal E}_{\xi} (H),\\
&&{\cal E}(H)=\mathbb{E}[Y-\mu_2(z_0)-H(\mu(Z)-\mu(z_0))]_{{\cal H}_{X,{\boldsymbol W}}}^2+\xi \|H\|_{{\cal L}_2({\cal H}_{X,{\boldsymbol W}},\Omega_Y)}^2,
\end{eqnarray}
\begin{eqnarray}
&&\hat{H}_{\xi}=\arg \min \hat{\cal E}_{\xi} (H),\\
&&\hat{\cal E}(H)=\hat{\mathbb{E}}[Y-\mu_2(z_0)-H(\mu(Z)-\mu(z_0))]_{{\cal H}_{X,{\boldsymbol W}}}^2+\xi \|H\|_{{\cal L}_2({\cal H}_{X,{\boldsymbol W}},\Omega_Y)}^2.
\end{eqnarray}
\end{definition}

\subsection*{Assumptions}

Next, we show assumptions for Theorem \ref{RTEO1}.
\begin{assumption}[Restriction for the domains]
\label{RAS1}
Suppose that $\Omega_{X,{\boldsymbol W}}$ and $\Omega_Z$ are Polish spaces, i.e., separable and completely metrizable topological spaces.
\end{assumption}

\begin{assumption}[Restriction for the feature functions]
\label{RAS2}
Suppose that 
\begin{enumerate}
    \item $k_{X,{\boldsymbol W}}$ and $k_{\boldsymbol Z}$ are continuous and bounded: $\sup_{x \in \Omega_{X,{\boldsymbol W}}} \|\pi(x,{\boldsymbol w})\|_{{\cal H}_{X,{\boldsymbol W}}} \leq Q$ and $\sup_{z \in \Omega_Z} \|\psi(z)\|_{{\cal H}_{\boldsymbol Z}} \leq \kappa$.
    \item $\pi$ and $\psi$ are measurable.
    \item $k_{X,{\boldsymbol W}}$ is characteristic.    
\end{enumerate}
\end{assumption}

\begin{assumption}[Uniqueness]
\label{RAS3}
Suppose that ${G_1}_{\rho} \in {\cal L}_2({\cal H}_{Z},{\cal H}_{\boldsymbol Z})$, then ${\cal E}_1({G_1}_{\rho})=\inf_{{G_1} \in {\cal H}_Z}{\cal E}_1({G_1})$.
Furthermore, suppose that ${G_2}_{\rho} \in {\cal L}_2({\cal H}_{Z},{\cal H}_{\boldsymbol Z})$, then ${\cal E}_1({G_2}_{\rho})=\inf_{{G_2} \in {\cal H}_{Z}}{\cal E}_1({G_2})$.
\end{assumption}

\begin{assumption}[Boundness of stage 1]
\label{RAS4}
    Fix $\zeta_1, \zeta_2 \leq \infty$. For given $c_1,c_2 \in (1,2]$, define the prior ${\cal P}(\zeta_1,c_1)$ and ${\cal P}(\zeta_2,c_2)$ as the set of the probability distributions on $\Omega_{X,{\boldsymbol W}} \times \Omega_Z$ such that a range space assumption is satisfied: $\exists C_1 \in {\cal L}_2({\cal H}_Z,{\cal H}_{X,{\boldsymbol W}})$ such that $G_{1\rho}=T_1^{\frac{c_1-1}{2}} \circ C_1$ and $\|C_1\|_{{\cal L}_2({\cal H}_Z,{\cal H}_{X,{\boldsymbol W}})}^2 \leq \zeta_1$, and 
    $\exists C_2 \in {\cal L}_2({\cal H}_Z,\Omega_Y)$ such that $G_{2\rho}=T_1^{\frac{c_2-1}{2}} \circ C_2$ and $\|C_2\|_{{\cal L}_2({\cal H}_Z,\Omega_Y)}^2 \leq \zeta_2$.
\end{assumption}

\begin{lemma}[Rate of convergence of stage 1 (A)]
Make Assumptions \ref{RAS1}, \ref{RAS2}, \ref{RAS3} and \ref{RAS4}. For all $\delta \in (0,1)$, the following holds w.p. $1-\delta$:
\begin{eqnarray}
&&    \|\hat{{G_1}}_{\lambda}-{G_1}_{\rho}\|_{{\cal L}_2({\cal H}_Z,{\cal H}_{X,{\boldsymbol W}})} \nonumber\\
&&\leq \frac{\sqrt{\zeta_1}(c_1+1)}{4^{\frac{1}{c_1+1}}}\left( \frac{4\kappa(Q+\kappa\|{G_1}_{\rho}\|_{{\cal L}_2({\cal H}_Z,{\cal H}_{X,{\boldsymbol W}})})ln(2/\delta)}{\sqrt{n\zeta_1}(c_1-1)}\right)
\end{eqnarray}
\end{lemma}

\begin{lemma}[Rate of convergence of stage 1 (B)]
Make Assumptions \ref{RAS1}, \ref{RAS2}, \ref{RAS3} and \ref{RAS4}. For all $\delta \in (0,1)$, the following holds w.p. $1-\delta$:
\begin{eqnarray}
&&    \|\hat{{G_2}}_{\lambda}-{G_2}_{\rho}\|_{{\cal L}_2({\cal H}_Z,\Omega_Z)} \nonumber\\
&&\leq \frac{\sqrt{\zeta_2}(c_2+1)}{4^{\frac{1}{c_2+1}}}\left( \frac{4\kappa(Q+\kappa\|{G_2}_{\rho}\|_{{\cal L}_2({\cal H}_Z,\Omega_Z)})ln(2/\delta)}{\sqrt{n\zeta_2}(c_2-1)}\right)
\end{eqnarray}
\end{lemma}
The proof is shown in \citep{Singh2019}.
The above lemma implies consistency of Stage 1 (A). 
%\jin{What happens to Stage 1(B)?}

\begin{assumption}[Restriction of domain]
\label{RAS5}
    Suppose that $\Omega_Y$ is a Polish space, i.e., separable and completely metrizable topological spaces.
\end{assumption}

\begin{assumption}[Boundness of stage 2]
\label{RAS6}
    Suppose that
\begin{enumerate}
    \item The $\{\Psi_{\mu(z)-\mu(z_0)}\}$ operator family is uniformly bounded in Hilbert-Schmidt norm: $\exists B$ such that $\forall \mu(z)$, $\|\Psi_{\mu(z)-\mu(z_0)}\|^2_{{\cal L}_2(\Omega_Z,{\cal L}_2({\cal H}_Z,{\cal H}_{X,{\boldsymbol W}}))}=Tr(\Psi^*_{\mu(z)-\mu(z_0)} \circ \Psi_{\mu(z)-\mu(z_0)}) \leq B$.
    \item The $\{\Psi_{\mu(z)-\mu(z_0)}\}$ operator family is H\"{o}lder continuous in operator norm: $\exists L > 0$, $\iota \in (0,1]$ such that $\forall \mu(z), \mu(z')$, $\|\Psi_{\mu(z)-\mu(z_0)}-\Psi_{\mu(z')-\mu(z_0)}\|_{L(\Omega_Z,{\cal L}_2({\cal H}_Z,{\cal H}_{X,{\boldsymbol W}}))} \leq L \|\mu(z)-\mu(z')\|^{\iota}_{{\cal H}_{X,{\boldsymbol W}}}$.
\end{enumerate}
\end{assumption}

\begin{assumption}[Boundness of stage 2]
\label{RAS7}
Suppose that
\begin{enumerate}
    \item $\left<H_{\rho},\cdot\ \right> \in {\cal L}_2({\cal H}_{X,{\boldsymbol W}},\Omega_Y)$. Then, ${\cal E}(H_{\rho})=\inf_{H \in {\cal H}_{X,{\boldsymbol W}}}{\cal E}(H)$.
    \item $Y$ is bounded, i.e. $\exists C < \infty$ such that $\|Y\| \leq C$ almost surely.
\end{enumerate}
\end{assumption}

\begin{assumption}[Boundness of stage 2]
\label{RAS8}
    Fix $\zeta < \infty$. For given $b \in (1,\infty]$ and $c \in (1,2]$, define the prior ${\cal P}(\zeta,b,c)$ as the set of probability distributions $\mathfrak{p}$ on ${\cal H}_{X,{\boldsymbol W}} \times \Omega_Y$ such that
    \begin{enumerate}
        \item A range space assumption is satisfied: $\exists C \in {\cal L}_2({\cal H}_{X,{\boldsymbol W}},\Omega_Y)$ such that $H_{\rho}=T_2^{\frac{c-1}{2}}\circ C$ and $\|C\|^2_{{\cal L}_2({\cal H}_{X,{\boldsymbol W}},\Omega_Y)} \leq \zeta$.
        \item In the spectral decomposition $T=\sum_{k=1}^{\infty}\lambda_ke_k\left< \cdot, e_k \right>_{{\cal H}_{X,{\boldsymbol W}}}$, where $\{e_k\}_{k=1}^\infty$ is a basis of $Ker(T)^{\perp}$, the eigenvalues satisfies $\alpha \leq k^b\lambda_k\leq \beta$ for some $\alpha, \beta>0$.
    \end{enumerate}
\end{assumption}
These assumptions are for the boundness of {\bf Stage 2}.

\begin{lemma}
\label{LEMKER}
    Make Assumptions \ref{RAS1}, \ref{RAS2}, \ref{RAS3}, \ref{RAS4}, \ref{RAS5}, \ref{RAS6}, \ref{RAS7} and \ref{RAS8}. Let $\lambda=N_1^{-\frac{1}{c_1+1}}$, $N_1=N_2^{\frac{a(c_1+1)}{\iota(c_1-1)}}$, $a>0$, and $\lambda_3=0$. We have 
    \begin{enumerate}
        \item if $a \leq \frac{b(c+1)}{bc+1}$ then ${\cal E}(\hat{H}_{\xi})-{\cal E}({H}_{\rho})={\cal O}_p(N_2^{-\frac{ac}{c+1}})$ with $\xi=N_2^{-\frac{a}{c+1}}$.
        \item if $a \geq \frac{b(c+1)}{bc+1}$ then ${\cal E}(\hat{H}_{\xi})-{\cal E}({H}_{\rho})={\cal O}_p(N_2^{-\frac{bc}{bc+1}})$ with $\xi=N_2^{-\frac{b}{bc+1}}$.
    \end{enumerate}
\end{lemma}
Lemma~\ref{LEMKER} can be proved  from the proof of Theorem 4 
in \citep{Singh2019} by subsituting $\mu(z)$ with $\mu(z)-\mu(z_0)$.

%\begin{theorem}
%    Make Assumptions \ref{RAS1}, \ref{RAS2}, \ref{RAS3}, \ref{RAS4}, \ref{RAS5}, \ref{RAS6}, \ref{RAS7} and \ref{RAS8}, then RKHS CAPCE estimator is consistent to CAPCE.
%\end{theorem}

%\subsection*{Proof of Theorem 4.5.}

{\it
{\bf Theorem \ref{RTEO1}.}
    Under SCM ${\cal M}_{IV}$ and Assumptions %3.1, 3.2,
    \ref{AS1}, \ref{AS2},  
    \ref{RAS1}, \ref{RAS2}, \ref{RAS3}, \ref{RAS4}, \ref{RAS5}, \ref{RAS6}, \ref{RAS7} and \ref{RAS8},
    %F.1, F.2, F.3, F.4, F.5, F.6, F.7, and F.8,
   the  RKHS CAPCE estimator in (25) %$\hat{\mathbb{E}}[\partial_x{Y}_{x}|{\boldsymbol W}={\boldsymbol w}]=\hat{\boldsymbol \alpha}^T{\bf K}_{(X,{\boldsymbol W})^{(1)}(x,{\boldsymbol w})}$ 
    converges pointwise to CAPCE %$\mathbb{E}[\partial_xY_{x}|{\boldsymbol W}={\boldsymbol w}]$ 
    when $\lambda_3=0$.
}
\begin{proof}
Lemma \ref{LEMKER} implies consistency of RKHS CAPCE estimator by taking limit $N_2 \rightarrow \infty$.
\end{proof}

%\newpage
%\section{Experiment: Additional Information of Parametric Settings}

%\newpage
\section{Additional Information on Experiments and the Application}

%\subsection{Additional Information of Experiments}

\label{appG}

In this section, we give detailed information about the settings of  the experiments and  additional experimental results.

{We note that the choice of the reference point $z_0$ does not affect the consistency results or rate of convergence, but it may affect the variance of the estimator. In our experiments, we take the minimum value of $Z$ as a standard reference point $z_0$. The choice of the reference point $z_0$ did not affect the standard deviation of the estimators much in our experiments.}

\subsection{Detailed settings of experiments}

We present detailed settings of numerical experiments in the following.
%to demonstrate the performance of the proposed P-CAPCE, sieve CAPCE,  and RKHS CAPCE estimators. \jin{Detailed settings are  in Appendix G.} 
 
%{\bf Baselines.} We compare with the most widely used methods PTSLS (parametric), NTSLS (sieve), and Kernel IV. These methods compute $\mathbb{E}[Y_{x}|{w}]$ which we differentiate to compute CAPCE  $\mathbb{E}[\partial_x Y_{x}|{w}]$.\\
%Additional information shown in Appendix.
%\subsection{Parametric Estimation}
%First, we compare the PTSLS and P-CAPCE estimator when $g_0$ is null in Eq. (\ref{EQ1}).\\
%{\bf SCM Settings.} We consider the following two SCMs:  $W:=U+E_1, X:=Z+W+U+E_2$, and 
%\begin{eqnarray}
%\label{eq-scm}
%\left\{
%\begin{array}{l}
%Y:=10X^2+WX+X+W+50(W^5+W^4+W^3+W^2)U+E_3\ \ \ \hspace*{\fill}\cdots\text{ (A)}\\Y:=\text{exp}(X)\text{exp}(W)+25(W^5+W^4+W^3+W^2)U+E_3\ \ \ \hspace*{\fill}\cdots\text{ (B)}\\
%\end{array}
%\right..
%\end{eqnarray}
%We use setting (A) as a parametric setting and setting (B) as a nonparametric setting. 
% Values of $Z$, $U$, $E_1$, $E_2$, and $E_3$ are   sampled i.i.d. from a uniform distribution $U[-1,1]$.
%True CAPCE  is $20x+w+1$ in setting (A) and $\text{exp}(x)\text{exp}(w)$ in setting (B). %The sample size are $N=1000$ and $N=10000$.

{\bf Setting of P-CAPCE  and PTSLS.} We learn the conditional expectations of basis functions
$\mathbb{E}[Y|Z=z]$, $\mathbb{E}[X|Z=z]$, $\mathbb{E}[WX|Z=z]$ and $\mathbb{E}[X^2|Z=z]$
by the nonlinear model, 
%\begin{eqnarray}
    $b_0+ b_1Z+b_2Z^2$.
%\end{eqnarray}
We used the basis terms $\{1,W,X\}$ for P-CAPCE and $\{1,W,X,WX,X^2\}$ for PTSLS, which match setting (A), and let $z_0=-1$.
%We regularize the matrix $\displaystyle \hat{\bf G}^T \hat{\bf G}$ by adding $0.001 {\bf I}$ for PTSLS estimator and $0.1 {\bf I}$ for P-CAPCE estimator, where ${\bf I}$ is an identity matrix of size $M$.
Regularize value is determined by test error from $\{1,10^{-1},10^{-2},10^{-3}\}$.
%The results of the test errors are shown in Table 1.

{\bf Setting of NTSLS and sieve CAPCE.} We learn the conditional expectations by the nonlinear model, 
%\begin{eqnarray}
    $b_0+ b_1Z+b_2Z^2+b_3Z^3$,
%\end{eqnarray}
We consider the  basis terms $h_p(X)h_q(W)$ for $p=0,1,2$ and $q=0,1,2$, where $h_p$ is Hermite polynomial functions ($h_0(t)=1$, $h_1(t)=t$, $h_2(t)=t^2-1$ and $h_3(t)=t^3-3t$), and let $z_0=-1$.
Let $\kappa=2$ and {$l=1$}, and we calculate $\hat{\Lambda}$ by Monte Carlo integration using uniform distribution $(x,w)=(U(-4,4),U(-2,2))$, where $\Omega_X \subseteq [-4,4]$ and $\Omega_X \subseteq [-2,2]$. Regularize value is determined by test error from $\{1,10^{-1},10^{-2},10^{-3}\}$.
%We regularize the matrix $\displaystyle \hat{\bf G}^T \hat{\bf G}$ by adding $10^{-2} \hat{\Lambda}$.
%In addition, we give an experiment using multivariate linear basis function $\{1,W,X\}$, which is a minimal basis function to build CAPCE.
%Results of the test errors are shown in Table 5.
We estimate CAPCE via differentiating estimated $\mathbb{E}[Y_{x}|{W}={w}]$.
%{\bf Setting of sieve NTSLS estimator.} We learn $\mathbb{E}[Y|Z=z], \mathbb{E}[h_p(X)h_q(W)|Z=z]$ for any $p=0,1,2,3$, $q=0,1,2,3$ and $q=0,1$ by the nonlinear model, 
%\begin{eqnarray}
%    $b_0+ b_1Z+b_2Z^2+b_3Z^3$.
%\end{eqnarray}
%where $h_0(t)=1$, $h_1(t)=t$, $h_2(t)=t^2-1$ and $h_3(t)=t^3-3t$.
%Multivariate linear basis function are $\{1,W,X,WX,X^2\}$.
%In this situation, the function $f_Y^2$ is mis-specified.
%Let $\kappa=2$, and we calculate $\hat{\Lambda}$ by Monte Carlo integration using uniform distridution $(x,w)=(U(-4,4),U(-2,2))$.
%Regularize value is determined by test error from $\{1,10^{-1},10^{-2},10^{-3},\ldots\}$.
%We regularize the matrix $\displaystyle \hat{\bf G}^T \hat{\bf G}$ by adding $10^{-3} \hat{\Lambda}$.
%We estimate CAPCE via differentiating estimated $\mathbb{E}[Y_{x}|{W}={w}]$. \jin{Explain why NTSLS uses different settings than S-CAPCE.}
%Results of the test errors are shown in Table 6.\\

{\bf Setting of kernel IV and RKHS CAPCE estimator.} We use polynomial kernel function $k_Z(z,z')=(z^Tz'+C_1)^{C_2}$ and $k_{X,{W}}((x,{w})(x,{w})^T+C_3)^{C_4}$.
We select the kernel parameters $(C_1,C_2)$ and $(C_3,C_4)$ from $\{1,2,3,4,5\} \times \{1,2,3,4,5\}$, respectively.
%, and determined $(\zeta_1,\zeta_2)=(4,5)$ by test error.
We select the regularize values $\lambda_1$ and $\lambda_2$ from $\{1,10^{-1},10^{-2},10^{-3}\}$, respectively, and $(\lambda_3,\xi)$ is from Cartesian product set $\{100,10,1\} \times \{100,10,1\}$. 

\subsection{Additional information on experimental results in the body of paper}
%We give additional  information on experiments. 
{\textbf{Results: Parametric setting (A).} The basic statistics of estimated coefficients by $100$ time simulations of 
PTSLS and P-CAPCE  are shown in Tables ~\ref{tab:PNUM_EXMP1} and \ref{tab:PNUM_EXMP2}. 
These tables supplement  Table 1 in the paper. The true and estimated CAPCE surfaces over $(X, W)$ are shown in Fig.~\ref{fig:FIG2}.}

{\textbf{Results: Nonparametric setting (B).} The true and estimated CAPCE surfaces over $(X, W)$ are shown in Fig.~\ref{fig:FIG22}.}

\begin{table}[H]
\centering
\caption{Basic statistics of the P-CAPCE estimator over 1000 runs when $N=1000$ and $N=10000$ in setting (A).
\vspace{0cm}}
\label{tab:PNUM_EXMP1}
\renewcommand{\arraystretch}{1.1}
\begin{minipage}[c]{0.45\textwidth}
\centering
\begin{tabular}{l|cccc}
\hline
   $N=1000$      & $1$ &$W$ &$X$ \\
   \hline
   True Coeff. & 1 & 1 & 20\\
         \hline \hline
Min.    & -51.515 & -18.715 & -116.480 \\
1st Qu. & -10.044 & 0.160   & -10.809  \\
Median  & -1.849  & 3.035   & 17.691   \\
3rd Qu. & 7.926   & 12.239  & 50.007   \\
Max.    & 40.458  & 109.791 & 213.306  \\
\hline
Mean    & -1.651  & 10.383  & 19.293   \\
SD      & 14.707  & 22.309  & 50.957   \\
\hline
\end{tabular}
\end{minipage}
\hspace{1cm}
\begin{minipage}[c]{0.45\textwidth}
\centering
\begin{tabular}{l|cccc}
\hline
   $N=10000$      & $1$ &$W$ &$X$ \\
   \hline
   True Coeff. & 1 & 1 & 20\\
         \hline \hline
Min.    & -13.466 & -2.302 & -23.863 \\
1st Qu. & -2.324  & -0.157 & 10.908  \\
Median  & 1.138   & 0.543  & 22.075  \\
3rd Qu. & 4.643   & 1.568  & 29.509  \\
Max.    & 14.988  & 11.017 & 59.559  \\
\hline
Mean    & 1.226   & 0.963  & 19.971  \\
SD      & 5.380   & 2.124  & 15.487 \\
\hline
\end{tabular}
\end{minipage}
\end{table}

%\subsection{NTSLS estimator}

\begin{comment}
\begin{table}[H]
\caption{Basic statistics of the test error of the NTSLS over 1000 runs for regularization value ($N=1000$); the bold number is the smallest.}
\centering
\label{tab:TEST2}

\begin{tabular}{l|ccc}
\hline
$\lambda$  & 1 & 0.1 & 0.01 \\
\hline \hline
Min.   & 15606279  & 13623674  & 14019980  \\
1st Qu.    & 20943014  & 20609263  & 21674532  \\
Median & 23919469  & 23994604  & 24633018  \\
3rd Qu.    & 27036646  & 27329930  & 28037611  \\
Max.   & 38088863  & 40235204  & 41456719  \\
\hline
Mean   & 24258777  & {\bf 24218756}  & 25204835  \\
SD     & 4837954.0 & 4749243.0 & 5104092.0 \\
\hline
\end{tabular}
\end{table}
\end{comment}

\begin{table}[H]
\centering
\caption{Basic statistics of the PTSLS over 1000 runs when $N=1000$ and $N=10000$ in setting (A).
\vspace{0cm}}
\label{tab:PNUM_EXMP2}
\renewcommand{\arraystretch}{1.1}
\begin{minipage}[c]{0.45\textwidth}
\centering
%
}
\vspace{-1.5cm}
\captionsetup{labelformat=empty}
\caption*{\footnotesize \hspace{0.1cm}(a) True CAPCE}
\end{minipage}
\hspace{0.1cm}
    \begin{minipage}[c]{0.31\textwidth}
    \scalebox{0.3}{
% Created by tikzDevice version 0.12.3.1 on 2023-04-23 17:06:01
% !TEX encoding = UTF-8 Unicode
%
}
\vspace{-1.18cm}
\captionsetup{labelformat=empty}
\caption*{\footnotesize \hspace{0.1cm}(b) PTSLS}
\end{minipage}
\hspace{0.05cm}
\begin{minipage}[c]{0.31\textwidth}
\scalebox{0.3}{% Created by tikzDevice version 0.12.3.1 on 2023-04-25 12:56:04
% !TEX encoding = UTF-8 Unicode
%

}
\vspace{-1.18cm}
\captionsetup{labelformat=empty}
\caption*{\footnotesize \hspace{0.1cm}(c) P-CAPCE}
\end{minipage}
    \caption{Parametric estimated surfaces in setting (A) (Mean, $N=10000$). X-axis is the value of  treatment variable ($X=x$), Y-axis is the value of covariate ($W=w$), and Z-axis is the value of CAPCE.
    %$\mathbb{E}[\partial_x Y_{x}|W=w]$.
    }
    \label{fig:FIG2}
\end{figure}
%\end{wrapfigure}

\begin{figure}
\Huge
    \centering
    \begin{minipage}[c]{0.475\textwidth}
    \scalebox{0.39}{
% Created by tikzDevice version 0.12.3.1 on 2023-04-23 17:12:20
% !TEX encoding = UTF-8 Unicode


}
\vspace{-0.5cm}
\captionsetup{labelformat=empty}
\caption*{\footnotesize \hspace{0.1cm}(a) True CAPCE}
\end{minipage}
    \begin{minipage}[c]{0.475\textwidth}
    \scalebox{0.35}{
% Created by tikzDevice version 0.12.3.1 on 2023-04-23 19:39:43
% !TEX encoding = UTF-8 Unicode
%

}
%\vspace{-0.45cm}
\captionsetup{labelformat=empty}
\caption*{\footnotesize \hspace{0.1cm}(b) Sieve NTSLS}
\end{minipage}

\begin{minipage}[c]{0.475\textwidth}
\scalebox{0.35}{
% Created by tikzDevice version 0.12.3.1 on 2023-04-23 19:37:41
% !TEX encoding = UTF-8 Unicode
%
}
%\vspace{-0.6cm}
\captionsetup{labelformat=empty}
\caption*{\footnotesize \hspace{0.1cm}(c) Sieve CAPCE}
\end{minipage}
\begin{minipage}[c]{0.475\textwidth}
\scalebox{0.35}{% Created by tikzDevice version 0.12.3.1 on 2023-04-26 21:52:12
% !TEX encoding = UTF-8 Unicode
%
}
%\vspace{-0.6cm}
\captionsetup{labelformat=empty}
\caption*{\footnotesize \hspace{0.1cm}(d) RKHS CAPCE}
\end{minipage}
    \caption{Nonparametric estimated surfaces in setting (B) (Mean, $N=10000$). X-axis is the value of  treatment variable ($X=x$), Y-axis is the value of  covariate ($W=w$), and Z-axis is the value of CAPCE.% $\mathbb{E}[\partial_x Y_{x}|W=w]$.
    }
    \label{fig:FIG22}
\end{figure}

\subsection{Additional experiments: No interaction  between covariates  and unobserved confounders}

In this section, we give additional experiments with no interaction  between covariates  and unobserved confounders.

{\bf SCM Settings.} We consider the following two SCMs:  $W:=H+E_1, X:=Z+W+H+E_2$, and 
\begin{eqnarray}
%\label{eq-scm}
\left\{
\begin{array}{l}
Y:=10X^2+WX+X+W+50\, H+E_3\ \ \ \hspace*{\fill}\cdots\text{ (C)}\\Y:=\text{exp}(X)\, \text{exp}(W)+50\, H+E_3\ \ \ \hspace*{\fill}\cdots\text{ (D)}\\
\end{array}
\right..
\end{eqnarray}
The other settings of each estimator are the same as in setting (A) and (B).

\textbf{Results.} 
The basic statistics of estimated coefficients by $100$ time simulations of 
PTSLS and P-CAPCE in setting (C) are shown in Tables ~\ref{tab:PNUM_EXMP1_A} and \ref{tab:PNUM_EXMP2_A}. 
The MSE of each estimator in settings (C) and (D) are shown in Table ~\ref{tab:TAB2_A}.
The results  show that the performance of the previous works PTSLS, NTSLS, Kernel IV is comparable with our proposed methods under the settings where the interaction between the covariates $W$ and unobserved confounders $H$ is absent.

%The MSE of all estimator are almost the same in setting (C). The MSE of NTSLS, Kernel IV, S-CAPCE and RKHS CAPCE are almost the same in setting (D). On the other hand, the MSE of PTSLS and P-CAPCE are worse than the others due to misspecification of the models.

\begin{table}[H]
\centering
\caption{Basic statistics of the P-CAPCE estimator over 1000 runs when $N=1000$ and $N=10000$ in setting (C).
\vspace{0.2cm}}
\label{tab:PNUM_EXMP1_A}
\renewcommand{\arraystretch}{1.1}
\begin{minipage}[c]{0.45\textwidth}
\centering
\begin{tabular}{l|cccc}
\hline
   $N=1000$      & $1$ &$W$ &$X$ \\
   \hline
   True Coeff. & 1 & 1 & 20\\
         \hline \hline
Min.    & -4.419 & -38.895 & -21.176 \\
1st Qu. & 0.896  & 0.351   & 19.614  \\
Median  & 0.983  & 1.361   & 19.919  \\
3rd Qu. & 1.065  & 2.620   & 20.246  \\
Max.    & 6.066  & 19.864  & 41.021  \\
\hline
Mean    & 0.944  & 1.151   & 19.642  \\
SD      & 0.811  & 5.535   & 4.884  \\
\hline
\end{tabular}
\end{minipage}
\hspace{1cm}
\begin{minipage}[c]{0.45\textwidth}
\centering
\begin{tabular}{l|cccc}
\hline
   $N=10000$      & $1$ &$W$ &$X$ \\
   \hline
   True Coeff. & 1 & 1 & 20\\
         \hline \hline
Min.    & 0.522 & -21.630 & 18.680 \\
1st Qu. & 0.976 & 0.158   & 19.901 \\
Median  & 1.000 & 0.830   & 20.006 \\
3rd Qu. & 1.021 & 1.989   & 20.101 \\
Max.    & 1.724 & 24.138  & 21.684 \\
\hline
Mean    & 0.999 & 0.966   & 19.998 \\
SD      & 0.106 & 4.851   & 0.305 \\
\hline
\end{tabular}
\end{minipage}
\end{table}

\begin{table}[H]
\centering
\caption{Basic statistics of the PTSLS over 1000 runs when $N=1000$ and $N=10000$ in setting (C).
\vspace{0.2cm}}
\label{tab:PNUM_EXMP2_A}
\renewcommand{\arraystretch}{1.1}
\begin{minipage}[c]{0.45\textwidth}
\centering
\begin{tabular}{l|ccccccc}
\hline
   $N=1000$      & $1$ &$W$ &$X$ \\
   \hline
   True Coeff. & 1 & 1 & 20\\
         \hline \hline
Min.    & 0.350 & -2.045 & 15.831 \\
1st Qu. & 0.957 & -0.199 & 19.303 \\
Median  & 1.021 & 1.009  & 19.731 \\
3rd Qu. & 1.101 & 1.921  & 19.953 \\
Max.    & 1.456 & 5.165  & 20.260 \\
\hline
Mean    & 1.029 & 0.997  & 19.474 \\
SD      & 0.155 & 1.609  & 0.782 \\
\hline
\end{tabular}
\end{minipage}
\hspace{1cm}
\begin{minipage}[c]{0.45\textwidth}
\centering
\begin{tabular}{l|ccccc}
\hline
   $N=10000$      & $1$ &$W$ &$X$\\
   \hline
   True Coeff. & 1 & 1 & 20\\
         \hline \hline
Min.    & 0.904 & -1.251 & 19.501 \\
1st Qu. & 0.986 & 0.515  & 19.905 \\
Median  & 1.003 & 0.962  & 19.963 \\
3rd Qu. & 1.021 & 1.407  & 20.019 \\
Max.    & 1.074 & 3.148  & 20.120 \\
\hline
Mean    & 1.003 & 0.939  & 19.939 \\
SD      & 0.028 & 0.814  & 0.118 \\
\hline
\end{tabular}
\end{minipage}
\end{table}

\begin{table}[H]
\centering
\caption{MSE of estimators in settings (C) and (D).}
\vspace{3pt}
\label{tab:TAB2_A}
\renewcommand{\arraystretch}{1.1}
\begin{tabular}{l|lll:lll}
\hline
   \multicolumn{1}{c|}{MSE}      & PTSLS &NTSLS & Kernel IV & P-CAPCE & S-CAPCE & RKHS CAPCE \\
        \hline \hline
(C) $N=1000$  & 3.004 & 4.124 & 8.165 &  8.316& 11.827 & 3.572 \\
(C) $N=10000$ & 0.254 & 0.367 & 0.614 & 0.774 & 1.324 & 0.567  \\ \hline
(D) $N=1000$  & 139.713 & 4.224  & 4.438 & 52.861 &  4.254  & 2.943 \\
(D) $N=10000$ & 57.006 &  0.334 & 1.026  & 40.062 &  0.477  &  0.696  \\ 
\hline
\end{tabular}
\end{table}

\subsection{Additional Experiments: Weaker interaction between covariates  and unobserved confounders}

In this section, we give additional experiments with weak interaction  between covariates  and unobserved confounders.

{\bf SCM Settings.} We consider the following two SCMs:  $W:=H+E_1, X:=Z+W+H+E_2$, and 
\begin{eqnarray}
%\label{eq-scmg4}
\left\{
\begin{array}{l}
Y:=10X^2+WX+X+W+10 (W^5+W^4+W^3+W^2)H+E_3\ \hfill\cdots\text{(E)}\\
Y:=\text{exp}(X)\text{exp}(W)+5 (W^5+W^4+W^3+W^2)H+E_3\hfill\cdots\text{(F)}
\end{array}
\right..
\end{eqnarray}
The other settings of each estimator are the same as in setting (A) and (B).

\textbf{Results.} 
The basic statistics of estimated coefficients by $100$ time simulations of 
PTSLS and P-CAPCE in setting (E) are shown in Tables ~\ref{tab:PNUM_EXMP1_B} and \ref{tab:PNUM_EXMP2_B}. 
The MSE of estimators in settings (E) and (F) are shown in Table ~\ref{tab:TAB2_B}.
The results show that our methods are superior to the previous works PTSLS, NTSLS, Kernel IV while the performance differences are less than that in the settings (A) and (B) where the interaction  between covariates  and unobserved confounders are stronger. 
%The difference between the performance of the previous works PTSLS, NTSLS, Kernel IV and our methods is weaker than the results of setting (A) and (B).

\begin{table}[H]
\centering
\caption{Basic statistics of the P-CAPCE estimator over 1000 runs when $N=1000$ and $N=10000$ in setting (E).
\vspace{0.2cm}}
\label{tab:PNUM_EXMP1_B}
\renewcommand{\arraystretch}{1.1}
\begin{minipage}[c]{0.45\textwidth}
\centering

\begin{tabular}{l|ccccccc}
\hline
   $N=1000$      & $1$ &$W$ &$X$ \\
   \hline
   True Coeff. & 1 & 1 & 20\\
         \hline \hline
Min.    & -2.056 & -9.914 & -6.174 \\
1st Qu. & -0.103 & -0.589 & 11.002 \\
Median  & 0.848  & 1.996  & 15.528 \\
3rd Qu. & 1.898  & 6.510  & 29.150 \\
Max.    & 2.047  & 3.502  & 25.181 \\
\hline
Mean    & 1.185  & 1.765  & 18.248 \\
SD      & 2.416  & 5.991  & 14.165 \\
\hline
\end{tabular}
\end{minipage}
\hspace{1cm}
\begin{minipage}[c]{0.45\textwidth}
\centering
\begin{tabular}{l|ccccc}
\hline
   $N=10000$      & $1$ &$W$ &$X$\\
   \hline
   True Coeff. & 1 & 1 & 20\\
         \hline \hline
Min.    & 0.057 & -1.633 & 10.648 \\
1st Qu. & 0.635 & -0.286 & 17.381 \\
Median  & 1.193 & 1.187  & 22.993 \\
3rd Qu. & 1.743 & 2.306  & 23.704 \\
Max.    & 2.047 & 3.502  & 25.181 \\
\hline
Mean    & 1.160 & 0.979  & 20.531 \\
SD      & 0.662 & 1.713  & 4.798  \\
\hline
\end{tabular}
\end{minipage}
\end{table}

\begin{table}[H]
\centering
\caption{Basic statistics of the PTSLS over 1000 runs when $N=1000$ and $N=10000$ in setting (E).
\vspace{0.2cm}}
\label{tab:PNUM_EXMP2_B}
\renewcommand{\arraystretch}{1.1}
\begin{minipage}[c]{0.45\textwidth}
\centering
\begin{tabular}{l|cccc}
\hline
   $N=1000$      & $1$ &$W$ &$X$ \\
   \hline
   True Coeff. & 1 & 1 & 20\\
         \hline \hline
Min.    & -5.345 & -1.064 & -3.732 \\
1st Qu. & -0.074 & 9.187  & 14.958 \\
Median  & 1.195  & 10.862 & 20.450 \\
3rd Qu. & 2.667  & 13.322 & 24.988 \\
Max.    & 5.766  & 34.620 & 37.757  \\
\hline
Mean    & 1.195  & 10.862 & 20.450 \\
SD      & 2.161  & 5.437  & 8.597  \\
\hline
\end{tabular}
\end{minipage}
\hspace{1cm}
\begin{minipage}[c]{0.45\textwidth}
\centering
\begin{tabular}{l|cccc}
\hline
   $N=10000$      & $1$ &$W$ &$X$ \\
   \hline
   True Coeff. & 1 & 1 & 20\\
         \hline \hline
Min.    & -1.072 & 7.402  & 11.514 \\
1st Qu. & 0.394  & 9.613  & 17.370 \\
Median  & 0.957  & 10.748 & 19.355 \\
3rd Qu. & 1.446  & 11.977 & 21.967 \\
Max.    & 2.684  & 15.641 & 26.645 \\
\hline
Mean    & 0.934  & 10.883 & 19.459 \\
SD      & 0.778 & 1.687  & 3.478  \\
\hline
\end{tabular}
\end{minipage}
\end{table}

\begin{table}[H]
\centering
\caption{MSE of estimators in settings (E) and (F).}
\vspace{3pt}
\label{tab:TAB2_B}
\renewcommand{\arraystretch}{1.1}
\begin{tabular}{l|lll:lll}
\hline
   \multicolumn{1}{c|}{MSE}      & PTSLS &NTSLS & Kernel IV & P-CAPCE & S-CAPCE & RKHS CAPCE \\
        \hline \hline
(E) $N=1000$  & 170.152 & 129.132 & 113.733 & 29.408  & 17.181 & 31.267 \\
(E) $N=10000$ & 64.645  & 55.104  & 65.055 & 3.592   & 4.527  & 3.171  \\\hline
(F) $N=1000$  & 141.678 & 28.73   & 24.125 & 101.902 & 13.84  & 14.806 \\
(F) $N=10000$ & 61.569  & 4.323   & 3.197 & 42.422  & 1.788  & 1.731  \\ 
\hline
\end{tabular}
\end{table}

\subsection{Additional Experiments: Estimation based on Theorem 3.1'.}
\label{sec-prime}

In this section, we give additional experiments about estimating CAPCE in the settings (A) and (B) in Eq.~(\ref{eq-scm}) based on Theorem 3.1' in Appendix~\ref{appA2}.
P-CAPCE', S-CAPCE', and RKHS CAPCE' estimate CAPCE based on Theorem 3.1'.
%PTSLS', NTSLS', and Kernel IV' estimate $\mathbb{E}[Y_{x}|{\boldsymbol w}]$ via an integral equation, $\mathbb{E}[Y|Z=z,W=1]=\int_{\Omega_X} \mathfrak{p}(X=x|Z=z,,W=1)\mathbb{E}[Y_{x}|,W=1]dx$.
We present detailed settings of numerical experiments in the following.

{\bf Setting of P-CAPCE'.} We learn the conditional expectations of basis functions
$\mathbb{E}[Y|Z=z,W=w]$, $\mathbb{E}[X|Z=z,W=w]$, $\mathbb{E}[WX|Z=z,W=w]$ and $\mathbb{E}[X^2|Z=z,W=w]$
by the nonlinear model, 
%\begin{eqnarray}
    $b_0+ b_1Z+b_2Z^2$.
%\end{eqnarray}
We used the basis terms $\{1,X,X^2\}$ for P-CAPCE' and $\{1,X\}$ for PTSLS, which match setting (A), and let $z_0=-1$ and $w=1$.\
Regularize value is determined by test error from $\{1,10^{-1},10^{-2},10^{-3}\}$.

{\bf Setting of S-CAPCE'.} We learn the conditional expectations by the nonlinear model, 
%\begin{eqnarray}
    $b_0+ b_1Z+b_2Z^2+b_3Z^3+ b_4W+b_5W^2+b_6W^3$,
%\end{eqnarray}
We consider the  basis terms $h_p(X)$ for $p=0,1,2$, where $h_p$ is Hermite polynomial functions ($h_0(t)=1$, $h_1(t)=t$, $h_2(t)=t^2-1$ and $h_3(t)=t^3-3t$), and let $z_0=-1$ and $w=1$.
Let $\kappa=2$ and $l=1$, and we calculate $\hat{\Lambda}$ by Monte Carlo integration using uniform distribution $x \sim U(-4,4)$, where $\Omega_X \subseteq [-4,4]$. Regularize value is determined by test error from $\{1,10^{-1},10^{-2},10^{-3}\}$.
We estimate CAPCE via differentiating estimated $\mathbb{E}[Y_{x}|{W}={w}]$.

{\bf Setting of RKHS CAPCE' estimator.} We use polynomial kernel function $k_{Z,{W}}((z,{w})(z,{w})^T+C_1)^{C_2}$ and  $k_X(x,x')=(x x'+C_3)^{C_4}$.
We select the kernel parameters $(C_1,C_2)$ and $(C_3,C_4)$ from $\{1,2,3,4,5\} \times \{1,2,3,4,5\}$, respectively.
%, and determined $(\zeta_1,\zeta_2)=(4,5)$ by test error.
We select the regularize values $\lambda_1$ and $\lambda_2$ from $\{1,10^{-1},10^{-2},10^{-3}\}$, respectively, and $(\lambda_3,\xi)$ is from Cartesian product set $\{100,10,1\} \times \{100,10,1\}$.

\textbf{Results.} 
The MSEs of P-CAPCE, S-CAPCE, RKHS CAPCE, P-CAPCE', S-CAPCE', and RKHS CAPCE' in settings (A) and (B) for $w=1$ are shown in Table ~\ref{tab:TABG5}.
%The MSEs of PTSLS', NTSLS', Kernel IV', P-CAPCE', S-CAPCE', RKHS CAPCE' in settings (A) and (B) for $w=1$ are shown in Table ~\ref{tab:TABG52}.
%The results show that our methods are superior to the previous methods PTSLS, NTSLS, and Kernel IV while the performance differences are less than that in the settings (A) and (B) based on both Theorem 3.1 and Theorem 3.1'. 
The results show that estimators  based on Theorem 3.1 and 3.1' have very similar performance.

\begin{table}[H]
\renewcommand{\arraystretch}{1.1}
\centering
\caption{MSE of each estimator based on Theorem 3.1 and 3.1' in settings (A) and (B) for $w=1$.}
\vspace{3pt}
\label{tab:TABG5}
\renewcommand{\arraystretch}{1.1}
\begin{tabular}{l|lll:lll}
\hline
   \multicolumn{1}{c|}{MSE}      & P-CAPCE & S-CAPCE & RKHS CAPCE & P-CAPCE' & S-CAPCE' & RKHS CAPCE' \\
        \hline \hline
(A) $N=1000$  & 453.233 & 225.301 & 339.091 &  132.167 & 399.446 & 193.306 \\
(A) $N=10000$  & 98.885 & 220.358 & 164.798  & 91.5647  & 275.907 & 153.689 \\\hline
(B) $N=1000$   & 284.598 & 14.398 & 30.562  &  129.721 & 11.780 & 28.266 \\
(B) $N=10000$  & 52.217 & 5.189 & 3.475  &  63.302 & 5.726 & 3.194 \\
\hline
\end{tabular}
\end{table}

%\begin{table}[H]
%\renewcommand{\arraystretch}{1.1}
%\centering
%\caption{MSE of each estimator by Theorem 3.1 in settings (A) and (B) for $w=1$.}
%\vspace{3pt}
%\label{tab:TABG51}
%\renewcommand{\arraystretch}{1.1}
%\begin{tabular}{l|lll:lll}
%\hline
%   \multicolumn{1}{c|}{MSE}      & PTSLS &NTSLS & Kernel IV & P-CAPCE & S-CAPCE & RKHS CAPCE \\
%        \hline \hline
%(A) $N=1000$  & 1037.884 & 730.138 & 453.233 & 225.301 & 339.091 &  \\
%(A) $N=10000$ & 510.511 & 381.156 &  & 98.885 & 220.358 &  \\\hline
%(B) $N=1000$   & 413.298  & 37.472 & 55.766 & 284.598 & 14.398 &  \\
%(B) $N=10000$  & 319.894 & 31.253 & 32.031 &  52.217 & 5.189 &  \\
%\hline
%\end{tabular}
%\end{table}

\begin{comment}
\begin{table}[H]
\renewcommand{\arraystretch}{1.1}
\centering
\caption{MSE of each estimator by Theorem 3.1' in settings (A) and (B) for $w=1$.}
\vspace{3pt}
\label{tab:TABG52}
\renewcommand{\arraystretch}{1.1}
\begin{tabular}{l|lll:lll}
\hline
   \multicolumn{1}{c|}{MSE}      & PTSLS' &NTSLS' & Kernel IV' & P-CAPCE' & S-CAPCE' & RKHS CAPCE' \\
        \hline \hline
(A) $N=1000$  & 1640.885 & 667.687 & 446.921 &  132.167 & 399.446 & 193.306 \\
(A) $N=10000$  & 635.293 & 467.709 & 432.868 & 91.5647  & 275.907 & 153.689 \\\hline
(B) $N=1000$   & 403.596 & 34.502 & 60.529 &  129.721 & 11.780 & 28.266 \\
(B) $N=10000$  & 365.361 & 25.841 & 50.644 &  63.302 & 5.726 & 3.194 \\
\hline
\end{tabular}
\end{table}
\end{comment}

\subsection{Additional information on the application}

We present detailed settings of the application in Section 6. 
We applied  P-CAPCE and PTSLS. %Other estimators are not used due to the small sample size.  
We learn the expected values of basis functions by the nonlinear model,
%\begin{eqnarray}
    $\beta_0+\beta_1Z+\beta_2Z^2$.
%\end{eqnarray}
We use terms $\{1,W,W^2,X,XW,XW^2\}$ for P-CAPCE and $\{1,W,W^2,X,XW,XW^2,X^2,X^2W,X^2W^2\}$ for PTSLS, {and let $z_0=8$}.
We estimate CAPCE via differentiating estimated $\mathbb{E}[Y_{x}|{w}]$ for PTSLS.
%We regularize the matrix $\displaystyle \hat{\bf G}^T \hat{\bf G}$ by adding $10^{-3}{\bf I}$.
Regularize parameter is determined by test error from $\{1,10^{-1},10^{-2},10^{-3},\ldots\}$. %\jin{Are there missing setting information?}
%Results of the test errors are shown in Table 7.\\
\begin{comment}
{\bf Setting of PTSLS estimator.} We learn $\mathbb{E}[Y|Z=z], \mathbb{E}[X^pW^q|Z=z]$ for $p=0,1,2$ and $q=0,1,2$ by the nonlinear model,
%\begin{eqnarray}
    $\beta_0+\beta_1Z+\beta_2Z^2$.
%\end{eqnarray}
We consider the following terms, $\{1,W,W^2,X,XW,XW^2,X^2,X^2W,X^2W^2\}$ to build model of $\mathbb{E}[Y_{x}|{W}={w}]$.
%We regularize the matrix $\displaystyle \hat{\bf G}^T \hat{\bf G}$ by adding $100{\bf I}$.
Regularize value is determined by test error from $\{1000,100,10,1,10^{-1}\}$.
We estimate CAPCE via differentiating estimated $\mathbb{E}[Y_{x}|{W}={w}]$.\\
%Results of the test errors are shown in Table 10.\\
\end{comment}

%\subsection*{Additional information on application}
{
{\bf Results.} 
The basic bootstrapping statistical properties of the P-CAPCE and PTSLS estimators are shown in Tables~\ref{tab:NUM_APMP1} and \ref{tab:NUM_APMP2}. The predicted CAPCE  values  are shown in Tables~\ref{tab:app-p1} and \ref{tab:app-p2}. The estimated CAPCE surfaces  are shown in Fig.~\ref{fig:FIG3}.
}

\begin{comment}
\begin{table}[H]
\caption{Basic statistics of the test error of the CAPCE estimator over 1000 runs for regularization value; the bold number is the smallest.}
\centering
\label{tab:ATEST1}

\begin{tabular}{l|ccc}
\hline
$\lambda$ & 1 & 0.1 & 0.01 \\
\hline \hline
Min.   & 8457869   & 6021324   & NA \\
1st Qu.    & 14552258  & 13155384  & NA \\
Median & 17036292  & 16429917  & NA \\
3rd Qu.    & 20301844  & 19727542  & NA \\
Max.   & 29788501  & 28455256  & NA \\
\hline
Mean   & 17657884  & {\bf 16589425}  & NA \\
SD     & 4662390.0 & 5035935.0 & NA \\
\hline
\end{tabular}
\end{table}
\end{comment}

\begin{table}[H]
\centering
\caption{Basic statistics of the P-CAPCE estimator over 1000 bootstrapping.
\vspace{0.2cm}}
\label{tab:NUM_APMP1}
\renewcommand{\arraystretch}{1.1}
\begin{minipage}[c]{1\textwidth}
\centering
\begin{tabular}{l|ccccccc}
\hline
   Terms    & $1$ &$W$ &$W^2$ &$X$&$WX$&$W^2X$ \\
         \hline \hline
Min.    & -0.00267 & -0.00003 & -0.00126 & -0.00006 & -0.00084 & -0.00237 \\
1st Qu. & -0.00061 & 0.00002  & 0.00510  & -0.00001 & -0.00029 & -0.00099 \\
Median  & -0.00006 & 0.00004  & 0.00904  & -0.00001 & -0.00016 & -0.00053 \\
3rd Qu. & 0.00058  & 0.00007  & 0.01331  & 0.00000  & -0.00008 & -0.00017 \\
Max.    & 0.00226  & 0.00018  & 0.02786  & 0.00004  & 0.00059  & 0.00068  \\
\hline
Mean    & -0.00003 & 0.00005  & 0.00949  & -0.00001 & -0.00018 & -0.00056 \\
SD      & 0.00090  & 0.00004  & 0.00602  & 0.00001  & 0.00018  & 0.00062 \\
\hline
\end{tabular}
\end{minipage}

\end{table}

\begin{table}[H]
\centering
\caption{Basic statistics of the PTSLS estimator over 1000 bootstrapping.
\vspace{0.2cm}}
\label{tab:NUM_APMP2}
\renewcommand{\arraystretch}{1.1}
\begin{minipage}[c]{1\textwidth}
\centering
%
}
%\vspace{-1.25cm}
\captionsetup{labelformat=empty}
\caption*{\footnotesize \hspace{0.1cm}(a) P-CAPCE}
\end{minipage}
    \begin{minipage}[c]{0.475\textwidth}
\centering
\scalebox{0.275}{
% Created by tikzDevice version 0.12.3.1 on 2023-04-25 12:51:13
% !TEX encoding = UTF-8 Unicode
%
}
%\vspace{-1.25cm}
\captionsetup{labelformat=empty}
\caption*{\footnotesize \hspace{0.1cm}(b) PTSLS}
\end{minipage}
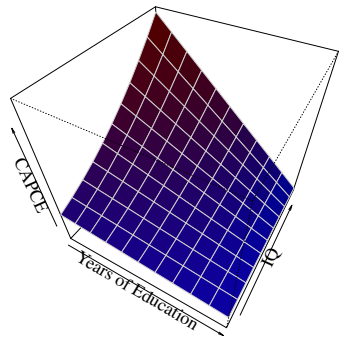
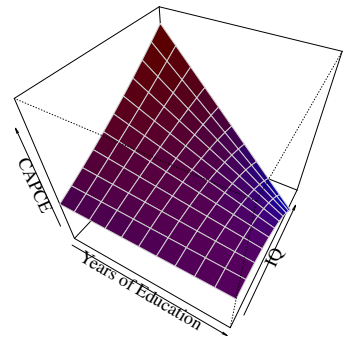
    \caption{Bootstrap mean surface of each estimator. X-axis is years of education, Y-axis is IQ, and Z-axis is CAPCE.% $\mathbb{E}[\partial_x Y_{x}|W=w]$ on wage.
    }
    \label{fig:FIG3}
\vspace{-0cm}
\end{figure}

%\end{wrapfigure}

%\subsection{PTSLS estimator}

\begin{comment}
\begin{table}[H]
\caption{Basic statistics of the test error of the PTSLS estimator over 1000 runs for regularization value; the bold number is the smallest.}
\centering
\label{tab:ATEST2}

\begin{tabular}{l|cccc}
\hline
$\lambda$& 1000 &100  & 10 & 1  \\
\hline \hline
Min.   & 708704433  & 718265667  & 707217598 & NA \\
1st Qu.    & 774156768  & 752119583  & 764372413 & NA \\
Median & 782873375  & 772095012  & 776719683 & NA \\
3rd Qu.    & 797060591  & 788883609  & 798262546 & NA \\
Max.   & 845302818  & 860740473  & 856730822 & NA \\
\hline
Mean   & 784901321  & {\bf 772438050}  & 779760448 & NA \\
SD     & 22690033.0 & 28528693.0 & 5035935.0 & NA\\
\hline
\end{tabular}
\end{table}
\end{comment}

\begin{landscape}

\label{PRE1}
\begin{table}[H]
\centering
\caption{Predicted CAPCE values  by P-CAPCE estimator
\vspace{0.2cm}\label{tab:app-p1}}
\renewcommand{\arraystretch}{1.1}
\begin{tabular}{l|lllllllllll}
\hline
 \diagbox{$X$}{$W$}  & 50       & 60       & 70       & 80       & 90       & 100      & 110       & 120        & 130        & 140        & 150        \\
   \hline\hline
8  & 12.375 & 17.838 & 24.296 & 31.750 & 40.199 & 49.645 & 60.086 & 71.523 & 83.955  & 97.384  & 111.808 \\
9  & 10.955 & 15.794 & 21.517 & 28.122 & 35.610 & 43.981 & 53.235 & 63.372 & 74.391  & 86.294  & 99.079  \\
10 & 9.534  & 13.751 & 18.738 & 24.495 & 31.021 & 38.318 & 46.384 & 55.221 & 64.827  & 75.204  & 86.350  \\
11 & 8.114  & 11.708 & 15.959 & 20.867 & 26.432 & 32.654 & 39.534 & 47.070 & 55.263  & 64.114  & 73.622  \\
12 & 6.693  & 9.665  & 13.180 & 17.239 & 21.843 & 26.991 & 32.683 & 38.919 & 45.700  & 53.024  & 60.893  \\
13 & 5.273  & 7.621  & 10.401 & 13.612 & 17.254 & 21.328 & 25.832 & 30.768 & 36.136  & 41.934  & 48.164  \\
14 & 3.852  & 5.578  & 7.622  & 9.984  & 12.665 & 15.664 & 18.982 & 22.618 & 26.572  & 30.844  & 35.435  \\
15 & 2.432  & 3.535  & 4.843  & 6.357  & 8.076  & 10.001 & 12.131 & 14.467 & 17.008  & 19.755  & 22.707  \\
16 & 1.011  & 1.492  & 2.064  & 2.729  & 3.487  & 4.338  & 5.280  & 6.316  & 7.444   & 8.665   & 9.978   \\
17 & -0.409 & -0.552 & -0.715 & -0.898 & -1.102 & -1.326 & -1.570 & -1.835 & -2.120  & -2.425  & -2.751  \\
18 & -1.829 & -2.595 & -3.494 & -4.526 & -5.691 & -6.989 & -8.421 & -9.986 & -11.684 & -13.515 & -15.480\\\hline
\end{tabular}
\end{table}

%\vspace{-0.7cm}
\label{PRE2}
\begin{table}[H]
\centering
\caption{Predicted CAPCE values by PTSLS estimator
\vspace{0.2cm} \label{tab:app-p2}}
\renewcommand{\arraystretch}{1.1}
\begin{tabular}{l|lllllllllll}
\hline
 \diagbox{$X$}{$W$}  & 50       & 60       & 70       & 80       & 90       & 100      & 110       & 120        & 130        & 140        & 150        \\
   \hline\hline
8  & 24.192 & 30.570 & 37.461 & 44.866 & 52.785 & 61.218 & 70.164 & 79.625  & 89.598  & 100.086 & 111.087 \\
9  & 23.820 & 29.504 & 35.495 & 41.793 & 48.399 & 55.312 & 62.532 & 70.059  & 77.894  & 86.035  & 94.484  \\
10 & 23.449 & 28.438 & 33.529 & 38.721 & 44.013 & 49.406 & 54.899 & 60.494  & 66.189  & 71.985  & 77.881  \\
11 & 23.077 & 27.373 & 31.563 & 35.648 & 39.626 & 43.499 & 47.267 & 50.928  & 54.484  & 57.934  & 61.279  \\
12 & 22.705 & 26.307 & 29.597 & 32.575 & 35.240 & 37.593 & 39.634 & 41.363  & 42.779  & 43.884  & 44.676  \\
13 & 22.334 & 25.242 & 27.631 & 29.502 & 30.854 & 31.687 & 32.001 & 31.797  & 31.075  & 29.833  & 28.073  \\
14 & 21.962 & 24.176 & 25.665 & 26.429 & 26.467 & 25.781 & 24.369 & 22.232  & 19.370  & 15.782  & 11.470  \\
15 & 21.590 & 23.110 & 23.699 & 23.356 & 22.081 & 19.874 & 16.736 & 12.666  & 7.665   & 1.732   & -5.133  \\
16 & 21.219 & 22.045 & 21.733 & 20.283 & 17.694 & 13.968 & 9.104  & 3.101   & -4.040  & -12.319 & -21.736 \\
17 & 20.847 & 20.979 & 19.767 & 17.210 & 13.308 & 8.062  & 1.471  & -6.465  & -15.745 & -26.369 & -38.339 \\
18 & 20.476 & 19.914 & 17.801 & 14.137 & 8.922  & 2.156  & -6.162 & -16.030 & -27.449 & -40.420 & -54.941 \\\hline
\end{tabular}
\end{table}

\end{landscape}

\end{document}